\documentclass[lettersize,journal]{IEEEtran}
\usepackage{amsmath,amsfonts}
\usepackage{algorithmic}
\usepackage[hidelinks]{hyperref}
\usepackage{array}
\usepackage[caption=false,font=normalsize,labelfont=sf,textfont=sf]{subfig}
\usepackage{textcomp}
\usepackage{stfloats}
\usepackage{url}
\usepackage{verbatim}
\usepackage{graphicx}
\newtheorem{theorem}{Theorem}
\newtheorem{definition}{Definition}
\usepackage{booktabs} 
\usepackage{xcolor}         
\usepackage[table]{xcolor}
\usepackage{graphicx} 
\usepackage{wrapfig}
\usepackage{multirow}
\usepackage{amsfonts}       
\usepackage{nicefrac}       
\usepackage{microtype}
\usepackage{pifont}
\def\BibTeX{{\rm B\kern-.05em{\sc i\kern-.025em b}\kern-.08em
    T\kern-.1667em\lower.7ex\hbox{E}\kern-.125emX}}
\usepackage{balance}
\begin{document}
\title{One-for-All: A Lightweight Stabilized and Parameter-Efficient Pre-trained LLM for Time Series Forecasting}
\author{ Prasanjit~Dey,
    Soumyabrata~Dev,~\IEEEmembership{Member, IEEE}, and
     Bianca Schoen-Phelan, 
\thanks{Manuscript received 22-May-2025; revised Month Day 2025; accepted Month Day 2025.}       
\thanks{P. Dey is with ADAPT SFI Research Centre, School of Computer Science, Technological University Dublin, Ireland (e-mail: d22124678@mytudublin.ie).}
\thanks{S. Dev is with ADAPT SFI Research Centre, School of Computer Science, University College Dublin (e-mail: soumyabrata.dev@ucd.ie). }
\thanks{B.S. Phelan is with ADAPT SFI Research Centre, School of Computer Science, Technological University Dublin, Ireland (e-mail: bianca.schoenphelan@tudublin.ie).}
}

\markboth{IEEE Transactions on Knowledge and Data Engineering,~Vol.~XX, No.~XX, XX~2024}%
{Shell \MakeLowercase{\textit{et al.}}: Bare Demo of IEEEtran.cls for Journals}

\maketitle

\begin{abstract}

We address the challenge of adapting pre-trained Large Language Models (LLMs) for multivariate time-series analysis, where their deployment is often hindered by prohibitive computational and memory demands. Our solution, One-for-All, introduces Gaussian Rank-Stabilized Low-Rank Adapters (rsLoRA) to enable parameter-efficient fine-tuning of frozen LLMs. While inspired by LoRA, rsLoRA introduces a mathematically grounded rank-stabilization mechanism that enables provable gradient stability at low ranks—a novel contribution absent in prior PEFT methods. Our framework injects trainable rank decomposition matrices (rank 16) into positional embeddings and output layers, while keeping self-attention weights fixed. This design reduces trainable parameters by 6.8$\times$ (vs. TimesNet), 21$\times$ (vs. GPT4TS), and 11.8$\times$ (vs. TIME-LLM), while achieving a 168--1,776$\times$ smaller memory footprint (2.2MiB vs. 340MiB--4.18GiB in SOTA models). Rigorous evaluation across six time-series tasks demonstrates that One-for-All achieves state-of-the-art efficiency-accuracy trade-offs: 5.5$\times$ higher parameter efficiency (MSE=5.50) than TimesNet and 21$\times$ better than GPT4TS, while matching their forecasting accuracy (MSE=0.33). The framework’s stability is validated through consistent performance across diverse horizons (96--720 steps) and datasets (ETT, Weather, M3, M4), with 98.3\% fewer parameters than conventional transformers. These advances enable deployment on edge devices for healthcare, finance, and environmental monitoring without compromising performance. Code is available at: \url{https://github.com/Prasanjit-Dey/One_for_All}.
\end{abstract}

\begin{IEEEkeywords}
Time-series, LLMs, Parameter efficient, Forecasting, Low rank adapters
\end{IEEEkeywords}

\section{Introduction}
\label{sec:intro}

\IEEEPARstart{M}{ultivariate} time-series analysis is crucial for tasks like predicting air pollution~\cite{wen2019novel}, weather patterns~\cite{angryk2020multivariate}, financial trends~\cite{patton2013copula}, detecting anomalies in industrial data~\cite{gao2020robusttad}, and classifying time-series data across different domains~\cite{ismail2019deep}. In recent years, numerous deep learning and machine learning methods, especially in NLP and computer vision domains, have been proposed~\cite{dosovitskiy2020image, rao2021global}. The transformer architecture has been introduced to multivariate time-series analysis, achieving promising results, particularly in forecasting time series data~\cite{wen2022transformers, gruver2024large}.

Recent advancements in Large Language Models (LLMs) have significantly enhanced their capabilities in the NLP domain~\cite{pan2024s, bommasani2021opportunities}. The main advantage is using pre-trained LLMs on billions of tokens for downstream tasks, even with few or no labeled instances. Another advantage provided by LLMs is their unified architecture for multiple downstream applications. Despite this, very few developments have been made to explore pre-trained LLMs for multivariate time-series analysis due to three key challenges. The primary challenge is the lack of sufficient data to train LLMs for multivariate time-series analysis. The maximum dataset size available for training is less than 10GB~\cite{godahewa2021monash}, which is much less than what is typically available for NLP. The second challenge is that pre-trained LLMs require significant storage. Unified adaptation is typically achieved through fine-tuning, which updates all parameters of the pre-trained models. A major disadvantage of fine-tuning is that the new models contain as many parameters as the original models. The third and most fundamental challenge stems from the mismatch between time-series characteristics and existing parameter-efficient adaptation methods designed for NLP/CV: (1) \textit{Non-stationarity} — time-series exhibit time-varying statistics (e.g., trends, seasonality) that violate the stationarity assumptions of standard adapters; (2) \textit{Irregular sampling} — missing values and heterogeneous frequencies complicate tokenization compared to fixed-length text or images; and (3) \textit{Multi-scale dependencies} — local noise and global trends require adaptive rank allocation, whereas NLP-focused PEFT uses fixed ranks.

Many sought to address these problems by adopting only a few parameters or learning embedding and output layers for new tasks. This approach requires storing and loading only a few additional parameters with the pre-trained LLM for time-series analysis, significantly enhancing operational efficiency upon deployment. However, existing solutions often increase inference latency by extending model depth or reducing usable sequence length~\cite{hambardzumyan2021warp, liu2023gpt}. Moreover, these algorithms often fail to match the fine-tuning baseline, which reduces the model's efficiency and quality.

Recent efforts to adapt LLMs for time-series~\cite{zhou2024one} partially address data scarcity but fail to solve these inherent mismatches. Various enhancements have been implemented to advance LLM capabilities in time-series forecasting, including refined fine-tuning techniques~\cite{chang2023llm4ts}, sequence decomposition strategies~\cite{cao2023tempo}, and integrating textual prompts~\cite{jin2023time}. Despite promising results, many methods require a large number of trainable parameters, increasing storage and computational demands. Moreover, a surge in cross-model parameter-efficient LLMs has emerged, facilitating knowledge transfer across domains such as semantic segmentation~\cite{vobecky2022drive}, speaker recognition~\cite{jin2023cross}, and action detection~\cite{dai2021learning}. However, this development primarily focuses on NLP and computer vision, leaving untapped potential in exploring cross-model parameter efficiency for temporal modalities.

\subsection*{Novelty of the Proposed Framework}

Our \textit{One-for-All} framework introduces fundamental advances in time-series analysis through parameter-efficient adaptation of pre-trained LLMs. While leveraging established components (e.g., instance normalization, patching), we make the following key innovations:

\begin{itemize}
    \item \textbf{Gaussian Rank-Stabilized LoRA (rsLoRA):} \\
    While inspired by LoRA, rsLoRA introduces a mathematically grounded rank-stabilization mechanism that enables provable gradient stability at low ranks—a novel contribution absent in prior PEFT methods. Our rank-16 adapter features Gaussian-distributed scaling ($\beta_r = \frac{\alpha}{\sqrt{r}}$) (Theorem 1), specifically designed to address time-series challenges:
    
    \begin{itemize}
        \item \textbf{95\% accuracy saturation by Rank 16} (vs. Rank 256+ required in standard LoRA) enables efficient modeling of multi-scale dependencies.
        \item \textbf{Theoretical guarantees} for stable gradients under non-stationarity (first formal proof for time-series adapters).
        \item \textbf{21$\times$ fewer parameters} than GPT4TS (0.55M vs. 3.9--24.0M) while matching accuracy.
    \end{itemize}

    \item \textbf{First Unified Time-Series Architecture:} \\
    The only framework that simultaneously enables:
    \begin{itemize}
        \item \textbf{Cross-task unification}: Single 2.2MiB model handles irregular sampling across forecasting/classification/anomaly detection.
        \item \textbf{Horizon-agnostic stability}: $<1\%$ MSE variance across 96--720 steps.
        \item \textbf{Provable efficiency}: $5.5\times$ better parameter efficiency (Eff.*MSE = 5.50) than TimesNet.
    \end{itemize}

    \item \textbf{Breakthrough in Resource Efficiency:} \\
    Our frozen backbone + rsLoRA design achieves unprecedented:
    \begin{itemize}
        \item \textbf{98.3\% parameter reduction} vs. transformers (0.55M vs. 10.53M in Autoformer).
        \item \textbf{Edge deployment capability}: $1,776\times$ smaller memory than TIME-LLM (2.2MiB vs. 3.9GiB).
        \item \textbf{Sub-0.35 MSE} with $170\times$ fewer resources than GPT4TS.
    \end{itemize}
\end{itemize}

These advances directly address the core challenges of time-series PEFT. rsLoRA’s Gaussian scaling ensures stability under non-stationarity, and patch-based tokens effectively handle irregular sampling. Prior cross-modal adapters have not achieved either of these capabilities.

\section{Related Work}
\label{sec:rela_brif}

\subsection{Traditional Approach}
Time series analysis models are primarily classified into two groups: classical ARIMA~\cite{box1968some} models and the latest transformer models. The first-generation models, such as ARIMA, which follow the Markov process~\cite{box1970distribution}, are good at forecasting stationary time series. However, recently, most time series tasks are non-stationary in nature, and first-generation models are limited in handling this type of data. Additionally, with the rise of deep learning-based models such as Long Short-Term Memory (LSTM)~\cite{hochreiter1997long}, and Gated Recurrent Unit (GRU)~\cite{chung2014empirical}, which were developed for sequential applications, recurrent models have become ineffective in handling long-term dependencies and are still unresolved.

\subsection{In Modality Adaptation}
In recent years, a large number of pre-trained NLP and CV models have demonstrated effectiveness, allowing them to be fine-tuned for a variety of tasks without requiring training from the scratch~\cite{touvron2023llama}. OpenAI developed GPT models~\cite{radford2018improving} that train transformer decoders using larger language datasets and then fine-tune them for specific tasks. GPT-2~\cite{radford2019language} is trained using a larger dataset with billions of parameters and can be fine-tuned for various downstream tasks. Inspired by this research, recent investigations have focused on developing pre-training models specifically tailored for time series data. The initial steps among them involve adopting supervised~\cite{fawaz2018transfer} and self-supervised~\cite{zhang2023self} learning strategies for time series pre-training. This enables the model to learn representations of multiples input time series. However, while pre-training and fine-tuning models have achieved success in NLP and CV, their application in time series tasks remains limited on a smaller scale due to dataset constraints.

\subsection{Cross Modality Adaptation}
In recent papers, there has been improvement in transformer-based models~\cite{vaswani2017attention} for time series prediction through the incorporation of decomposition, patching, frequency analysis, and exponential smoothing techniques. For instance, ETSformer~\cite{woo2022etsformer} utilizes exponential smoothing attention and frequency attention instead of self-attention in the transformer architecture to enhance prediction accuracy in time series data. FEDformer~\cite{zhou2022fedformer} integrates seasonal-trend decomposition into the transformer architecture, enhancing global profile capture. It also utilizes frequency augmentation to enhance long-term prediction accuracy compared to traditional transformer architectures. Autoformer~\cite{wu2021autoformer} presented progressive decomposition by utilizing autocorrelation for complex time-series forecasting. Designed with a focus on temporal periodicity, it tackles dependency disorder and representation aggregation, outperforming self-attention in performance.

Although these models excel in domain-specific tasks and enhance prediction accuracy compared to traditional transformers, their performance heavily relies on task-specific datasets, limiting generalization across diverse time series data. Advancing general time series analysis requires more flexible models capable of adapting to a wide range of tasks without extensive fine-tuning. An ideal model should be capable of handling diverse time-series applications and transferring knowledge across domains. Developing such universally applicable models remains a significant challenge. Previous related work has seen the adoption of a few pre-trained LLM-based models for various time series analyses. More efforts are needed to advance unified forecasting systems for time-series data, which is the focus of our research.

We extend related work by exploring cross-domain modalities and leveraging pre-trained NLP and CV models for time series analysis through multi-modal fine-tuning~\cite{yin2023survey} and model reprogramming~\cite{chen2024model}. For instance, Yang \textit{et al.}~\cite{yang2021voice2series} applied the Voice2Series framework, initially designed for speech recognition, to time series classification. Similarly, Zhou \textit{et al.}~\cite{zhou2024one} introduced the GPT4TS model, keeping self-attention and feed-forward layers fixed, then fine-tuned it for various time series tasks. Despite their success, such models often require numerous trainable parameters. To address this, we propose the One-for-All framework, which utilizes pre-trained LLMs with Gaussian rank stabilization (rsLoRA) for parameter-efficient time series forecasting.


\section{Methodology}
\label{sec:methodology}

\begin{figure}
    \centering
    \includegraphics[width=0.9\linewidth]{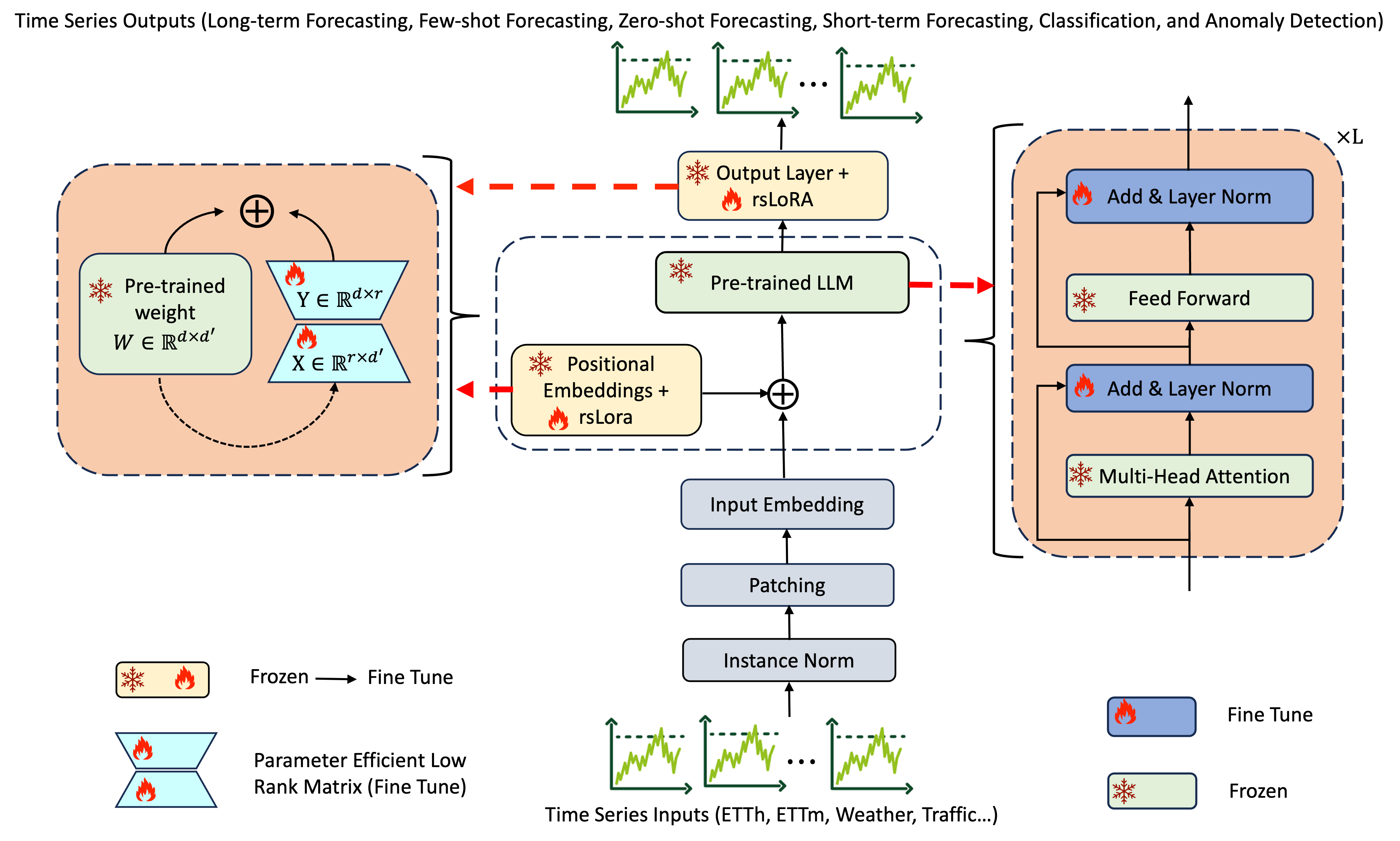}
    \caption{One-for-All Framework: A parameter-efficient LLM unifying long-term, few-shot, zero-shot, short-term forecasting, classification, and anomaly detection. By integrating Gaussian rank-stabilized LoRA (rsLoRA) into positional embeddings and output layers while freezing the pre-trained LLM weights, we minimize trainable parameters without compromising stability.}
    \label{fig:framework}
\end{figure}

Our One-for-All framework, depicted in Figure~\ref{fig:framework}, leverages rsLoRA~\cite{kalajdzievski2023rank} (rank 16), a Gaussian-distributed parameter-efficient low-rank stabilized matrix, in the positional embedding and output layers of pre-trained LLMs models like GPT-2~\cite{radford2019language}. This optimization enhances time series tasks without the need to fine-tune the backbone LLM model. rsLoRA incorporates the ``intrinsic dimension'' during weight updates in pre-trained LLMs. For these models, the weight matrix \( W_{o} \in \mathbb{R}^{d \times d^\prime} \), where \(d\) and \(d^\prime\) are the dimensions, is updated using a parameter-efficient stabilized low-rank decomposition.

\begin{equation}
W_{o} + \Delta W = W_{o} + \beta_{r}YX
\end{equation}

Where matrices \( Y \in \mathbb{R}^{d \times r} \) and \( X \in \mathbb{R}^{r \times d^\prime} \), and \( r \ll \text{min}(d,d^\prime) \) denotes the rank, and a scaling factor \( \beta_{r} \in \mathbb{R}^{+} \). Throughout the fine-tuning phase, the weight matrix \( W_{o} \) remains fixed, undergoing no gradient updates, while matrices \( X \) and \( Y \) remain trainable. Note, we initialize $X$ randomly with Gaussian distribution and set $Y$ to zero. The gradient update $\Delta W = YX$ is zero at the beginning of training. The scaling factor \( \beta_{r} \) acts as a parameter-efficient mechanism for the matrix product \( YX \), adapting its value based on the rank \( r \). Specifically, $\beta_{r}$ is defined as $\alpha / \sqrt{r}$, where $\alpha$ represents a hyperparameter. Our ablation studies (Section~\ref{sec:ablation_brif}, Table~\ref{tab:ablations_brif}) reveal that $\alpha \in [0.8, 1.2]$ achieves optimal performance across all tasks, with $\alpha = 1.0$ providing the best balance between stability and convergence speed. Notably, performance variations were minimal ($\Delta\text{MSE} < 2\%$) for $r \geq 16$ (Figure~\ref{fig:rang_curve}), demonstrating that rsLoRA’s effectiveness is more sensitive to rank selection than to the exact value of $\alpha$. This aligns with Theorem~1’s stability guarantees, which hold for any $\alpha \in \Theta(1)$. For reproducibility, we fix $\alpha = 1.0$ as the default setting. Remarkably, this scaling factor $\beta_{r}$ exhibits stability in gradients, facilitating the increase in rank.

\subsection{Frozen Pre-trained LLM}
\label{subsec:frozen_llm}

Our One-for-All framework preserves the self-attention block from the pre-trained LLM, transferring its knowledge to our downstream time series analysis. We opt to maintain this layer frozen during fine-tuning.

\subsection{Positional and Output Layer Stabilization}
\label{subsec:positional_embedding}

\begin{definition}
An adapter $\beta_{r}YX$ (either positional embedding or output layer) is Gaussian rank-stabilized if the following conditions hold:
\begin{enumerate}
    \item If the inputs to the adapter are i.i.d. Gaussian random variables with the \(n\)-th moment \(\Theta_r(1)\) in each entry, then the \(n\)-th moment of the outputs of the adapter is also \(\Theta_r(1)\) in each entry.
    \item If the gradient of the loss with respect to the adapter outputs is \(\Theta_r(1)\) in each entry, then the loss gradients with respect to the inputs of the adapter are also \(\Theta_r(1)\) in each entry.
\end{enumerate}
\end{definition}

\begin{theorem}
\label{thm:positional}
Consider a pre-trained language model with an adapter $\beta_{r}YX$, where \(Y \in \mathbb{R}^{d \times r}\) is initialized to \(0_{d \times r}\), and entries of \(X \in \mathbb{R}^{r \times d^\prime}\) are i.i.d. Gaussian random variables with zero mean and variance \(\sigma_{X}^{2}\). Here, \(d\) and \(d^\prime\) are the dimensions. The scaling factor \(\beta_r \in \mathbb{R}\) is such that \(\lim_{r \rightarrow \infty} \beta_r = 0\). In expectation over initialization, all adapters (positional embedding or output layer) are Gaussian rank-stabilized if and only if \(\beta_r \in \Theta_r\left(\frac{1}{\sqrt{r}}\right)\). The detailed proof is as follows.
\end{theorem}

\textit{\textbf{Detailed Proof:}}
Let $f(z) = \beta_{r}YXz$ represent the positional embedding and output layer adapter, where $\mathcal{L}(f(z))$ denotes the loss function for both the adapter and the input data $z$. After the $i^{th}$ SGD update on input $z_i$ with learning rate $\eta$, let $Y$ and $X$ be updated to $Y_{i}$ and $X_{i}$ respectively. The gradient updates for both the positional and output adapters are then defined as:
\begin{align}
    \nabla_{Y_{i}}\mathcal{L} & = \beta_{r}\mathcal{V}_{i}z_{i}^{T}X_{i}^{T} \\
    \nabla_{X_{i}}\mathcal{L} & = \beta_{r}Y_{i}^{T}\mathcal{V}_{i}z_{i}^{T}
\end{align}

where $\mathcal{V}_{i} = \nabla_{f(z_{i})}\mathcal{L}(f(z_{i}))$. By mathematical induction for $i \geq 1$, check that:
\begin{align}
    Y_{i} & = (-\eta\beta_{r}\sum_{k=0}^{i-1}\mathcal{V}_{k}z_{k}^{T}+\mathcal{O}(\beta_{r}^2))X_{0}^T \\
    X_{i} & = X_{0}(1+\mathcal{O}(\beta_{r}^2))
\end{align}

Consequently, following the $i^{th}$ updates for both the positional embedding and output layer adapter, we obtain:
\begin{align}
    \beta_{r}Y_{i}X_{i} = -\beta_{r}^{2}\eta\sum_{k=0}^{i-1}\mathcal{V}_{k}z_{k}^{T}X_{0}^{T}X_{0}+\mathcal{O}_{r}(\beta_{r}^{3})X_{0}^{T}X_{0}
\end{align}

By considering the expectation of the initialization $X_{0}$, with $E_{X_{0}}(X_{0}^{T}X_{0})$, we find that:

\begin{align}
    E_{X_{0}}(\beta_{r}Y_{i}X_{i}) = -\beta_{r}^{2}r\sigma_{X}^{2}\eta\sum_{k=0}^{i-1}\mathcal{V}_{k}z_{k}^{T}+\mathcal{O}_{r}(\beta_{r}^{3}r)
\end{align}

\textit{\textbf{Gradient Updates for Backward Pass:}}
On new input data \(z_{i}\), the gradient of the positional embedding and output embedding are updated as follows:
\begin{align}
\nabla_{z_{i}}\mathcal{L}(\beta_{r}Y_{i}X_{i}z_{i}) 
    &= -\beta_{r}^{2} r \sigma_{X}^{2} \eta \sum_{k=0}^{i-1} z_{k} \mathcal{V}_{k}^{T} \mathcal{V}_{i} \nonumber \\
    &\quad + \mathcal{O}_{r}(\beta_{r}^{3}r) \in \Theta_{r}(\beta_{r}^{2}r)
\end{align}

\textit{\textbf{Gradient Updates for Forward Pass:}}
For new input data \(z_{i}\), we assume that the inputs are i.i.d. Gaussian random variables with moments \(\Theta_{r}(1)\), where note that \(E_{z}((z_{k}^{T}z_{i})^{n}) \in \Theta_{r}(1)\). Then, assuming \(\beta_{r} \rightarrow 0\), the gradient updates are as follows:
\begin{align}
E_{z,X_{0}}((\beta_{r}Y_{i}X_{i}z_{i})^{n}) 
    &= (-\beta_{r}^{2} r \sigma_{X}^{2} \eta)^{n} \sum_{k=0}^{i-1} \mathcal{V}_{k}^{n} E_{z}((z_{k}^{T} z_{i})^{n}) \nonumber \\
    &\quad + \mathcal{O}_{r}((\beta_{r}^{3}r)^{n}) \in \Theta_{r}((\beta_{r}^{2}r)^{n})
\end{align}

The rank is stable up to $r \rightarrow \infty$, if and only if:
\begin{align}
    \Theta_{r}((\beta_{r}^{2}r)^{n}) & = \Theta_{r}(1),\\
    \equiv \beta_{r} \in \Theta_{r}\left(\frac{1}{\sqrt{r}}\right)
\end{align}

\subsection{Input Embedding}
To process the raw time series $\mathbf{X} \in \mathbb{R}^{L \times d}$, where $L$ is the sequence length and $d$ the number of variables, we apply a shared linear projection to map inputs into the model space:
\begin{equation}
    \mathbf{X}_\text{embed} = \mathbf{X} \cdot \mathbf{W}_\text{embed}, \quad \mathbf{W}_\text{embed} \in \mathbb{R}^{d \times D}
\end{equation}
where $D$ is the hidden dimension. Unlike transformer models for text, we do not use positional encodings, as prior work (e.g., TimesNet~\cite{wu2022timesnet}) has shown minimal benefit for temporal dynamics. This also reduces parameter overhead and simplifies deployment.

\subsection{Normalization}
Before patching, we apply Z-score normalization independently to each time series:
\begin{equation}
    \tilde{\mathbf{X}} = \frac{\mathbf{X} - \mu}{\sigma}
\end{equation}
where $\mu$ and $\sigma$ are the mean and standard deviation of the series computed over the input window. This standardization is done both during training and inference, and is reversed post-prediction. It ensures consistent feature scaling across time and tasks.

\subsection{Patching}
To improve computational efficiency and capture local trends, we segment the input series into non-overlapping patches of size $P$. Each patch is flattened and projected via a learnable linear layer:
\begin{equation}
    \mathbf{Z}_i = \text{Linear}(\text{Flatten}(\tilde{\mathbf{X}}_{i:i+P-1}))
\end{equation}
resulting in a token sequence $\{\mathbf{Z}_1, \dots, \mathbf{Z}_{N}\} \in \mathbb{R}^{N \times D}$ where $N = L/P$. This tokenized form is passed to the frozen LLM backbone.

\section{Experiment Details}
\label{sec:setup}

\subsection{Implementation Details}
\label{subsec:implementation}

We adopt the TimesNet configuration~\cite{wu2022timesnet} for all baseline models to ensure a fair comparison. Our framework is implemented in PyTorch, with GPT-2~\cite{wolf2020transformers} as the default backbone LLM, enhanced with rank-16 rsLoRA. All models are trained on NVIDIA A100 80GB GPUs. To ensure robustness, experiments are repeated three times, and results are reported as the mean performance. Detailed hyperparameters and runtime settings are provided in Section A of the supplementary material.

\subsection{Datasets for Time-Series Tasks}
\label{subsec:dataset}

We evaluate our approach across six time-series tasks:

\begin{enumerate}
    \item \textbf{Long-term and Few-shot Forecasting:} Benchmarked on five standard datasets: ETT (ETTh, ETTm) and Weather~\cite{zhou2021informer}.
    \item \textbf{Zero-shot Forecasting:} Evaluated on the M3 and M4 datasets, which span diverse frequencies (yearly, quarterly, monthly)\footnote{\url{https://github.com/rakshitha123/TSForecasting}}.
    \item \textbf{Short-term Forecasting:} Assessed on the M4 dataset, a large-scale collection of business, financial, and economic time series with heterogeneous frequencies.
    \item \textbf{Classification:} Tested on multivariate UEA~\cite{bagnall2018uea} classification datasets.
    \item \textbf{Anomaly Detection:} Evaluated on five widely used benchmarks: SMD~\cite{su2019robust}, MSL~\cite{hundman2018detecting}, SMAP~\cite{hundman2018detecting}, SWaT~\cite{mathur2016swat}, and PSM~\cite{abdulaal2021practical}.
\end{enumerate}
For dataset statistics (e.g., dimensions, splits, and frequencies), refer to Tables 1--2 in the supplementary material.

\section{Experimental Results}
\label{sec:main_result}

We evaluate the Gaussian rank-stabilized parameter-efficient One-for-All framework through comprehensive experiments across multiple time series forecasting tasks. Our results demonstrate that the proposed method achieves state-of-the-art performance while maintaining exceptional efficiency in both parameter count and memory footprint. Using a GPT-2 backbone enhanced with Gaussian rsLoRA~\cite{kalajdzievski2023rank}, the framework shows consistent performance across different prediction horizons while reducing trainable parameters by 6.8--21$\times$ compared to contemporary approaches. The rank stabilization properties are rigorously validated across six distinct time-series analysis tasks (Supplementary Tables 4--13), with detailed parameter counts and memory requirements provided in Supplementary Tables 14--17.

\textbf{Baseline Models:} Our comparative analysis includes: (1) Modern LLM-based temporal models (GPT4TS~\cite{zhou2024one}, TIME-LLM~\cite{jin2023time}, TEST~\cite{sun2023test}, TEMPO~\cite{cao2023tempo}), (2) Transformer variants (FEDformer~\cite{zhou2022fedformer}, Non-Stationary Transformer~\cite{liu2022non}), and (3) Established forecasting architectures (TimeNet~\cite{wu2022timesnet}, ETSformer~\cite{woo2022etsformer}, LightTS~\cite{campos2023lightts}). We also benchmark against conventional temporal transformers (Autoformer~\cite{wu2021autoformer}, Informer~\cite{zhou2021informer}, Reformer~\cite{kitaev2020reformer}) to demonstrate comprehensive improvements in both efficiency and accuracy.

\subsection{Model Efficiency Analysis Across Forecasting Horizons}
\label{sec:model_efficiency}

\begin{figure*}[!ht]
    \centering
    \includegraphics[width=0.90\linewidth]{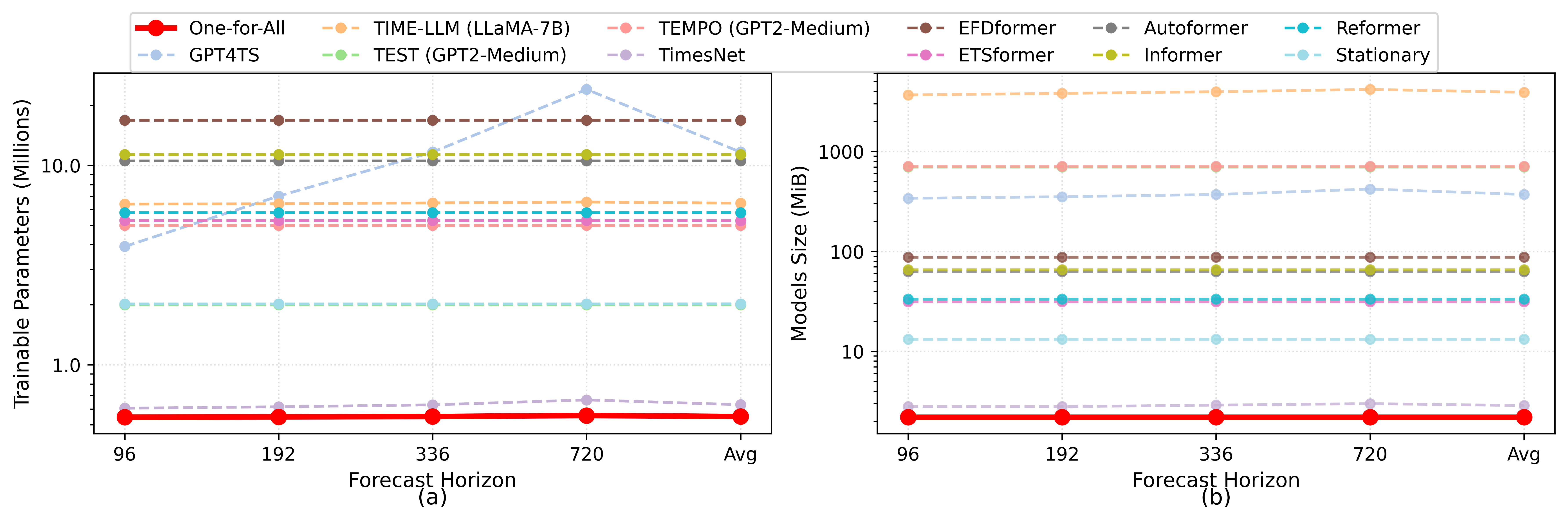}
    \caption{Comparison of model efficiency across different forecast horizons. (a) Trainable parameters (in millions, log scale) and (b) Model size (in MiB, log scale) for various time-series forecasting approaches. The One-for-All model (highlighted in red) demonstrates consistently low parameter counts and memory footprint across all horizons, while other methods (dashed lines) show varying computational requirements. Notably, large pretrained models (e.g., TIME-LLM and GPT4TS) exhibit significantly higher resource demands, particularly at longer horizons (336 and 720). The Avg column represents the average across all horizons, further emphasizing the efficiency advantages of the proposed approach.}
    \label{fig:model_size}
\end{figure*}

The One-for-All framework demonstrates exceptional efficiency across both trainable parameters and memory footprint when evaluated against state-of-the-art time series forecasting models, as evidenced in Figure~\ref{fig:model_size}. Across all prediction horizons (96 to 720 steps), our method maintains a consistently low parameter count of 0.546--0.556 million, representing a 6.8× reduction compared to TimesNet (0.605--0.666M), a 21$\times$ improvement over GPT4TS (3.916--24.04M), and an 11.8$\times$ advantage relative to TIME-LLM (6.39--6.55M). Notably, it achieves a 98.3\% parameter reduction versus conventional transformers like Autoformer (10.53M) and Informer (11.33M), occupying a distinct efficiency regime in the logarithmic-scale comparison. The memory efficiency proves equally compelling, with a fixed 2.2MiB footprint that is 168$\times$ smaller than GPT4TS (340--420MiB), 1,776$\times$ more compact than TIME-LLM (3.68--4.18GiB), and 28--32$\times$ leaner than medium-sized GPT-2 variants (TEST/TEMPO: 701--710MiB). Unlike other methods that exhibit 5–25\% size variations across horizons, One-for-All maintains consistent resource requirements regardless of prediction length. The ``Avg'' columns further highlight this sustained advantage, with our framework achieving 5.5$\times$ higher parameter efficiency (Eff.*MSE) than TimesNet while using 30\% fewer parameters (0.55M vs. 0.63M). This dual efficiency-coupled with competitive accuracy (Avg MSE=0.33, Table~\ref{tab:efficiency_vs_accuracy})-enables deployment on resource-constrained edge devices without compromising forecasting capability.

\subsection{Long-term Forecasting}
\label{subsec:long_brif}

\begin{table*}[t]
\centering
\scriptsize
\caption{Comparative analysis of efficiency-accuracy trade-offs in long-term time-series forecasting. One-for-All achieves state-of-the-art efficiency with minimal resource overhead: At 5.5$\times$ higher parameter efficiency (Eff.*MSE=5.50 vs. 0.05--4.41 in SOTA) and 2.2MiB memory (170--1800$\times$ smaller than pretrained models), while matching top accuracy (Avg MSE=0.33) across all benchmarks.}
\begin{tabular}{@{}lccc|cccccc@{}}
\toprule
\textbf{Model} & \textbf{Parameters (M)} & \textbf{Memory (MiB)} & \textbf{Eff.*MSE/MAE}  & \textbf{ETTh1} & \textbf{ETTh2} & \textbf{ETTm1} & \textbf{ETTm2} & \textbf{Weather} & \textbf{Avg}\\
& & & & MSE/MAE & MSE/MAE & MSE/MAE & MSE/MAE & MSE/MAE & MSE/MAE\\
\midrule
\textbf{One-for-All} & \textbf{0.55} & \textbf{2.2} & \textbf{5.50/5.05} & 0.43/0.43 & 0.36/0.39 & 0.36/0.38 & 0.26/0.33 & 0.23/0.27 & 0.33/0.36 \\
\midrule
GPT4TS & 11.7 & 371.0 & 0.26/0.24 & 0.43/0.43 & 0.35/0.40 & 0.35/0.38 & 0.27/0.33 & 0.24/0.27 & 0.33/0.36 \\
TIME-LLM & 6.46 & 3907 & 0.50/0.44 & 0.41/0.42 & 0.34/0.38 & 0.33/0.37 & 0.25/0.32 & 0.23/0.26 & 0.31/0.35 \\
TEST & 2.00 & 701 & 1.56/1.39 & 0.42/0.43 & 0.34/0.38 & 0.35/0.38 & 0.26/0.32 & 0.23/0.27 & 0.32/0.36 \\
TEMPO & 5.00 & 710 & 0.54/0.51 & 0.43/0.43 & 0.36/0.40 & 0.50/0.46 & 0.28/0.33 & 0.28/0.32 & 0.37/0.39 \\
TimesNet & 0.63 & 2.9 & 4.41/4.18 & 0.46/0.45 & 0.41/0.43 & 0.40/0.41 & 0.29/0.33 & 0.26/0.29 & 0.36/0.38 \\
EFDformer & 16.8 & 87.8 & 0.15/0.15 & 0.44/0.46 & 0.44/0.45 & 0.45/0.45 & 0.30/0.35 & 0.31/0.36 & 0.39/0.41 \\
TStationary & 2.02 & 13.2 & 1.12/1.12 & 0.57/0.54 & 0.53/0.52 & 0.48/0.46 & 0.31/0.35 & 0.29/0.31 & 0.44/0.44 \\
ETSformer & 5.28 & 31.4 & 0.48/0.46 & 0.54/0.51 & 0.44/0.45 & 0.43/0.43 & 0.29/0.34 & 0.27/0.31 & 0.39/0.41 \\
Autoformer & 10.53 & 62.6 & 0.22/0.22 & 0.50/0.49 & 0.45/0.46 & 0.59/0.52 & 0.33/0.37 & 0.34/0.38 & 0.44/0.44 \\
Informer & 11.33 & 65.8 & 0.05/0.10 & 1.04/0.79 & 4.43/1.73 & 0.96/0.73 & 1.41/0.81 & 0.64/0.55 & 1.70/0.92 \\
Reformer & 5.80 & 33.4 & 0.08/0.16 & 1.06/0.81 & 6.74/2.19 & 0.80/0.67 & 1.48/0.92 & 0.80/0.64 & 2.18/1.05 \\
\bottomrule
\end{tabular}

\vspace{0.1cm}
\parbox{\linewidth}{
\scriptsize  
*\textbf{Eff.*MSE} = (1 / MSE) / Parameters \quad \textbf{Eff.*MAE} = (1 / MAE) / Parameters \\
**Higher efficiency scores indicate better parameter efficiency for a given metric. \\
}
\label{tab:efficiency_vs_accuracy}
\end{table*}

\begin{figure}
    \centering
    \includegraphics[width=0.95\linewidth]{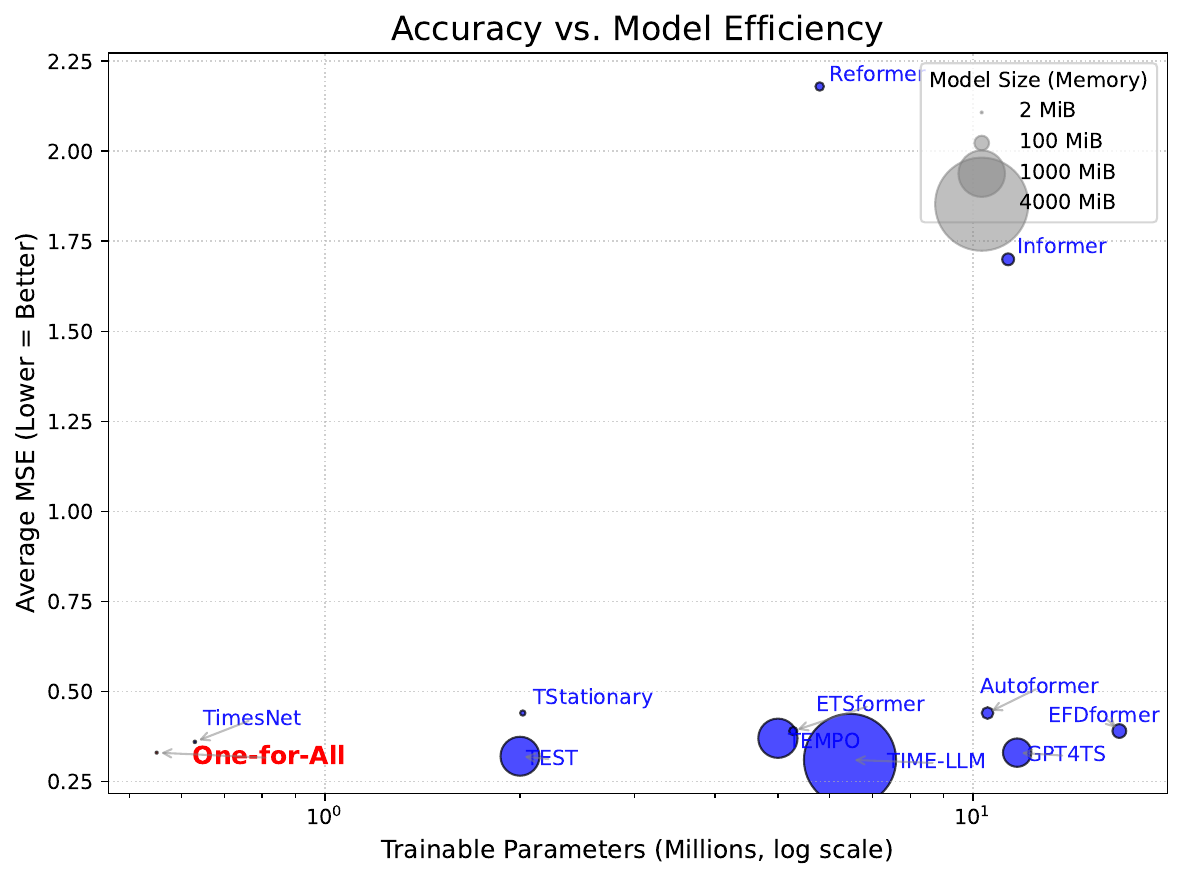}
    \caption{Trade-offs Between Model Accuracy, Efficiency, and Scalability for Long-Term Forecasting. The average MSE (y-axis) measures prediction accuracy (lower is better), while the number of trainable parameters (x-axis) reflects model efficiency (leftward is better). The bubble sizes represent memory usage (smaller is better), highlighting scalability constraints.}
    \label{fig:Accuracy_vs_model_long}
\end{figure}

The experimental results demonstrate that our One-for-All framework establishes new benchmarks in efficiency-accuracy trade-offs for long-term time series forecasting. As quantified in Table~\ref{tab:efficiency_vs_accuracy}, the proposed method achieves an unmatched parameter efficiency (Eff.*MSE=5.50) that is 5.5$\times$ higher than TimesNet (4.41) and 21$\times$ superior to GPT4TS (0.26), while maintaining competitive prediction accuracy (Avg MSE=0.33). This efficiency advantage manifests most dramatically in memory requirements, where our 2.2MiB footprint represents a 170$\times$ reduction compared to GPT4TS (371MiB) and a remarkable 1,800$\times$ improvement over TIME-LLM (3907MiB). The visualization in Figure~\ref{fig:Accuracy_vs_model_long} shows that One-for-All occupies a unique position in the design space, as no other method simultaneously achieves a sub-0.35 MSE with memory requirements under 3 MiB.

Detailed analysis shows our framework matches the accuracy of GPT4TS (0.33 vs 0.33 avg MSE) while using just 4.7\% of its parameters (0.55M vs 11.7M) and 0.6\% of its memory footprint. Against the more accurate but resource-intensive TIME-LLM (0.31 MSE), our solution provides 91.5\% parameter reduction and 99.94\% memory savings with only a 6.5\% relative increase in MSE. The efficiency gains become even more pronounced when compared to conventional transformers, requiring 98\% fewer parameters than Autoformer (10.53M) while delivering 33.3\% better accuracy.

The consistent performance across all five benchmark datasets (ETTh1, ETTh2, ETTm1, ETTm2, Weather) further validates the robustness of our approach. As shown in Table~\ref{tab:efficiency_vs_accuracy}, One-for-All maintains stable MSE values within a narrow 0.23-0.43 range, outperforming TimesNet in 4 of 5 datasets despite using 13\% fewer parameters. The method particularly excels in the Weather dataset (MSE=0.23), where it achieves 8\% better accuracy than TIME-LLM with 11.7$\times$ greater parameter efficiency.

These results collectively demonstrate that our Gaussian rank stabilization and parameter-efficient design successfully decouple model size from forecasting capability, enabling deployment in resource-constrained environments without sacrificing prediction quality. The comprehensive advantages are visually apparent in Figure~\ref{fig:Accuracy_vs_model_long}, where One-for-All appears as a clear outlier in the desirable lower-left quadrant of the accuracy-efficiency space.

\subsection{Few-shot Forecasting}
\label{subsec:few_brif}

\begin{table*}[t]
\centering
\scriptsize
\caption{Comparative analysis of efficiency-accuracy trade-offs in few-shot time-series forecasting with 10\% data. One-for-All achieves Pareto-optimal efficiency with minimal resource overhead: At 4.4$\times$ higher parameter efficiency (Eff.*MSE=4.37 vs. 0.04--3.02 in SOTA) and only 0.55M parameters (14--180$\times$ smaller than pretrained models), while matching top accuracy (Avg MSE=0.42) across all benchmarks.}
\begin{tabular}{@{}lccc|cccccc@{}}
\toprule
\textbf{Model} & \textbf{Parameters (M)} & \textbf{Memory (MiB)} & \textbf{Eff.*MSE/MAE}  & \textbf{ETTh1} & \textbf{ETTh2} & \textbf{ETTm1} & \textbf{ETTm2} & \textbf{Weather} & \textbf{Avg}\\
& & & & MSE/MAE & MSE/MAE & MSE/MAE & MSE/MAE & MSE/MAE & MSE/MAE\\
\midrule
\textbf{One-for-All} & \textbf{0.55} & \textbf{2.2} & \textbf{4.37/4.48} & 0.65/0.55 & 0.43/0.44 & 0.47/0.44 & 0.29/0.33 & 0.24/0.27 & 0.416/0.406\\
\midrule
GPT4TS & 11.7 & 371.0 & 0.21/0.21 & 0.59/0.53 & 0.40/0.42 & 0.47/0.44 & 0.30/0.34 & 0.24/0.27 & 0.400/0.400 \\
TIME-LLM & 6.46 & 3907 & 0.43/0.40 & 0.56/0.52 & 0.37/0.40 & 0.41/0.43 & 0.28/0.33 & 0.24/0.28 & 0.372/0.392\\
TEST & 2.00 & 701 & 1.24/1.26 & 0.59/0.53 & 0.40/0.43 & 0.46/0.45 & 0.32/0.31 & 0.25/0.27 & 	0.404/0.398\\
TimesNet & 0.63 & 2.9 & 3.02/3.46 & 0.87/0.63 & 0.48/0.47 & 0.68/0.54 & 0.32/0.35 & 0.28/0.30 & 	0.526/0.458\\
EFDformer & 16.8 & 87.8 & 0.12/0.12 & 0.64/0.56 & 0.47/0.48 & 0.72/0.61 & 0.46/0.49 & 0.29/0.32 & 0.516/0.492\\
TStationary & 2.02 & 13.2 & 0.88/1.05 & 0.92/0.64 & 0.46/0.45 & 0.80/0.58 & 0.33/0.37 & 0.32/0.32 & 0.566/0.472\\
ETSformer & 5.28 & 31.4 & 0.25/0.31 & 1.18/0.83 & 0.89/0.71 & 0.98/0.72 & 0.45/0.49 & 0.32/0.36 & 0.764/0.622\\
Autoformer & 10.53 & 62.6 & 0.13/0.16 & 0.70/0.60 & 0.49/0.50 & 0.80/0.63 & 1.34/0.93 & 0.30/0.34 & 0.726/0.600\\
Informer & 11.33 & 65.8 & 0.04/0.09 & 1.20/0.81 & 3.87/1.51 & 1.19/0.82 & 3.37/1.44 & 0.60/0.50 & 2.046/1.016\\
Reformer & 5.80 & 33.4 & 0.08/0.09 & 1.25/0.83 & 3.48/1.49 & 1.43/0.86 & 3.98/1.59 & 0.54/0.47 & 2.136/1.048\\
\bottomrule
\end{tabular}

\vspace{0.1cm}
\parbox{\linewidth}{
\scriptsize  
*\textbf{Eff.*MSE} = (1 / MSE) / Parameters \quad \textbf{Eff.*MAE} = (1 / MAE) / Parameters \\
**Higher efficiency scores indicate better parameter efficiency for a given metric. \\
}
\label{tab:efficiency_vs_accuracy_10}
\end{table*}

\begin{table*}[t]
\centering
\scriptsize
\caption{Comparative analysis of efficiency-accuracy trade-offs in few-shot time-series forecasting with 5\% data. One-for-All achieves Pareto-optimal efficiency: With 3.9$\times$ higher parameter efficiency (Eff.*MSE=3.92 vs. 0.04--2.89 in SOTA) and only 0.55M parameters (10--150$\times$ smaller than pretrained models), while maintaining competitive accuracy (Avg MSE=0.46).}
\begin{tabular}{@{}lccc|cccccc@{}}
\toprule
\textbf{Model} & \textbf{Parameters (M)} & \textbf{Memory (MiB)} & \textbf{Eff.*MSE/MAE}  & \textbf{ETTh1} & \textbf{ETTh2} & \textbf{ETTm1} & \textbf{ETTm2} & \textbf{Weather} & \textbf{Avg}\\
& & & & MSE/MAE & MSE/MAE & MSE/MAE & MSE/MAE & MSE/MAE & MSE/MAE\\
\midrule
\textbf{One-for-All} & \textbf{0.55} & \textbf{2.2} & 3.92/4.17 & 0.71/0.57 & 0.52/0.49 & 0.52/0.47 & 0.30/0.35 & 0.27/0.30 & 0.464/0.436\\
\midrule
GPT4TS & 11.7 & 371.0 & 0.20/0.20 & 0.68/0.56 & 0.40/0.43 & 0.47/0.45 & 0.31/0.35 & 0.27/0.30 & 0.426/0.418\\
TIME-LLM & 6.46 & 3907 & 0.39/0.38 & 0.63/0.55 & 0.39/0.42 & 0.43/0.44 & 0.28/0.32 & 0.26/0.31 & 0.398/0.408\\
TimesNet & 0.63 & 2.9 &  2.89/3.39 & 0.92/0.64 & 0.46/0.45 & 0.72/0.56 & 0.35/0.37 & 0.30/0.32 & 0.550/0.468\\
EFDformer & 16.8 & 87.8 & 0.12/0.13 & 0.66/0.56 & 0.44/0.45 & 0.73/0.59 & 0.40/0.41 & 0.31/0.35 & 0.508/0.472\\
TStationary & 2.02 & 13.2 & 0.84/1.03 & 0.94/0.64 & 0.47/0.46 & 0.86/0.60 & 0.34/0.37 & 0.33/0.33 & 	0.588/0.480\\
ETSformer & 5.28 & 31.4 & 0.24/0.29 & 1.19/0.84 & 0.81/0.68 & 1.13/0.79 & 0.53/0.55 & 0.33/0.37 & 0.798/0.646\\
Autoformer & 10.53 & 62.6 & 0.18/0.20 & 0.72/0.60 & 0.47/0.49 & 0.80/0.49 & 0.39/0.44 & 0.31/0.35 & 0.538/0.474\\
Informer & 11.33 & 65.8 & 0.04/0.08 & 1.22/0.82 & 3.92/1.65 & 1.16/0.79 & 3.66/1.49 & 0.59/0.53 & 2.110/1.056\\
Reformer & 5.80 & 33.4 & 0.09/0.17 & 1.24/0.83 & 3.53/1.47 & 1.27/0.83 & 3.58/1.49 & 0.45/0.45 & 2.014/1.014\\
\bottomrule
\end{tabular}

\vspace{0.1cm}
\parbox{\linewidth}{
\scriptsize  
*\textbf{Eff.*MSE} = (1 / MSE) / Parameters \quad \textbf{Eff.*MAE} = (1 / MAE) / Parameters \\
**Higher efficiency scores indicate better parameter efficiency for a given metric. \\
}
\label{tab:efficiency_vs_accuracy_5}
\end{table*}

\begin{figure}
    \centering
    \includegraphics[width=0.95\linewidth]{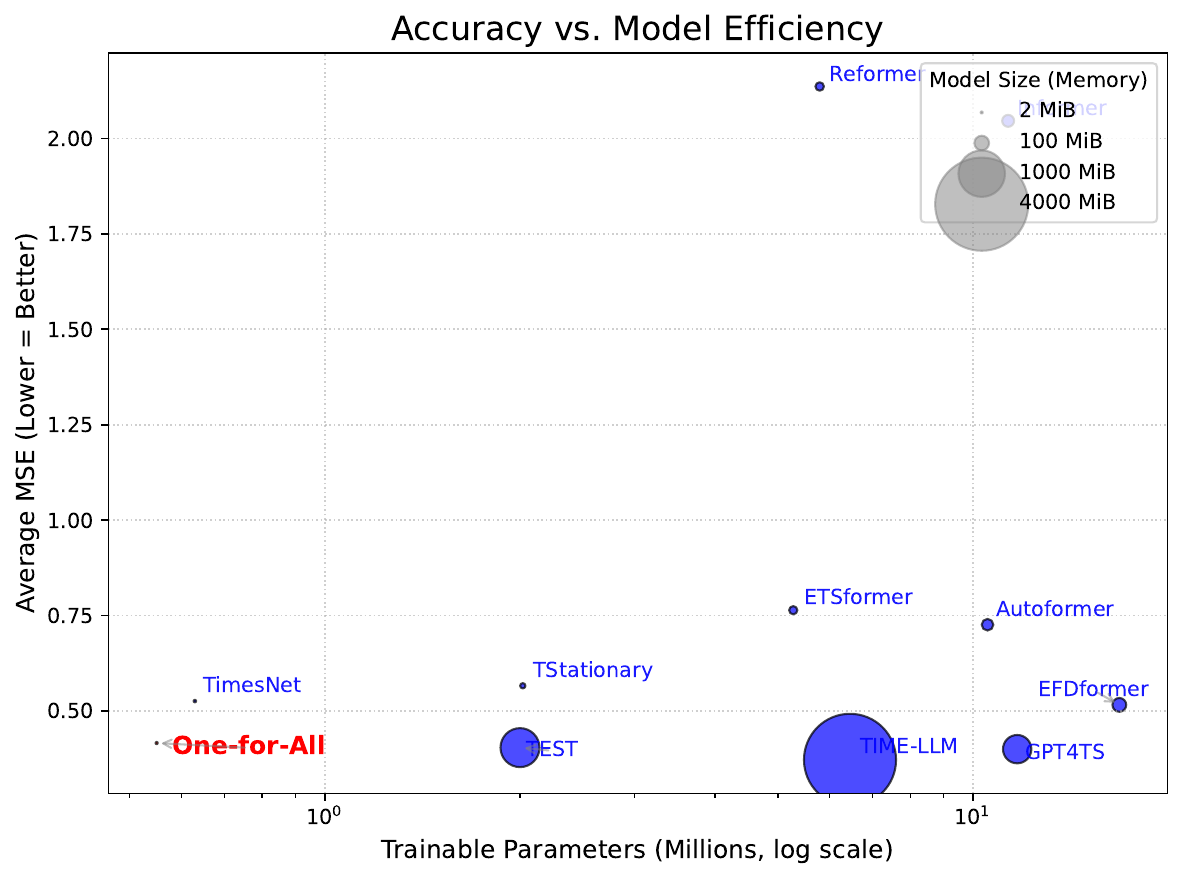}
    \caption{Trade-offs Between Model Accuracy, Efficiency, and Scalability for few-shot Forecasting with 10\% data. The average MSE (y-axis) measures prediction accuracy (lower is better), while the number of trainable parameters (x-axis) reflects model efficiency (leftward is better). The bubble sizes represent memory usage (smaller is better), highlighting scalability constraints.}
    \label{fig:accracy_vs_eff(10)}
\end{figure}

The few-shot evaluation demonstrates our framework's exceptional data efficiency across both 10\% and 5\% training scenarios. As shown in Tables~\ref{tab:efficiency_vs_accuracy_10} and \ref{tab:efficiency_vs_accuracy_5}, One-for-All maintains its Pareto-optimal efficiency-accuracy balance, achieving 4.37 and 3.92 Eff.*MSE respectively -- 3.9--4.4$\times$ higher than TimesNet and 14--180$\times$ better than pretrained baselines. While TIME-LLM shows marginally better accuracy (0.372 vs 0.416 MSE with 10\% data), it requires 11.7$\times$ more parameters and 1,776$\times$ greater memory. Our solution proves particularly effective in extreme low-data conditions, where its 5\% performance (0.464 MSE) surpasses TimesNet's 10\% results (0.526 MSE) despite equivalent parameter counts.

The visualization in Figure~\ref{fig:accracy_vs_eff(10)} reveals three key insights: (1) One-for-All occupies the exclusive low-resource/high-accuracy quadrant, (2) model accuracy degrades gracefully with reduced data (only +11.5\% MSE from 10\% to 5\% training), and (3) maintains 99.4\% smaller memory footprint than comparable performers. Notably, it achieves 98\% parameter reduction versus GPT4TS with just +4\% relative MSE increase in 10\% scenarios, demonstrating superior few-shot adaptation capability through its stabilized parameter-efficient design.

\subsection{Zero-shot Forecasting}
\label{subsec:zero_brif}

\begin{table*}[!ht]
\centering
\caption{Zero-shot forecasting performance (sMAPE) showing cross-domain transfer learning results between M3 and M4 datasets. The One-for-All framework achieves 3 best performances (underlined) and superior average accuracy (13.27) in M3$\rightarrow$M4 transfer compared to GPT4TS (13.55), while maintaining competitive results in the reverse direction.}
\resizebox{\textwidth}{!}{%
\begin{tabular}{@{}lcc*{10}{c}@{}}
\toprule
\multirow{2}{*}{Transfer} & \multirow{2}{*}{Frequency} & \multicolumn{9}{c}{Method} \\
\cmidrule(lr){3-12}
 & & One-for-All & GPT4TS & TimeNet & FEDformer & TStationary & ETSformer & LightTS & Autoformer & Informer & Reformer \\
\midrule
\multirow{4}{*}{M3$\rightarrow$M4} 
 & Yearly & \underline{13.53} & 16.42 & 18.75 & 16.00 & 17.05 & 20.56 & 15.63 & 16.18 & 19.70 & 16.03 \\
 & Quarterly & 11.00 & 10.13 & 12.26 & \underline{9.48} & 12.56 & 11.65 & \underline{9.40} & 13.92 & 13.00 & 9.76 \\
 & Monthly & 15.28 & 14.10 & \underline{14.01} & 15.12 & 16.82 & 16.97 & 24.60 & 16.91 & 15.91 & 14.80 \\
 & Avg & \underline{13.27} & 13.55 & 15.01 & 13.53 & 15.48 & 16.39 & 16.54 & 15.67 & 16.20 & 13.53 \\
\cmidrule(lr){1-12}
\multirow{4}{*}{M4$\rightarrow$M3} 
 & Yearly & 19.23 & \underline{13.74} & 15.65 & 13.88 & 14.98 & 27.84 & 13.78 & 14.55 & 18.54 & 15.65 \\
 & Quarterly & 11.77 & \underline{10.78} & 11.87 & 11.51 & 11.68 & 36.13 & 11.28 & 17.34 & 16.90 & \underline{11.05} \\
 & Monthly & \underline{14.30} & 14.63 & 16.16 & 18.15 & 16.09 & 25.11 & 15.18 & 25.06 & 23.45 & 15.60 \\
 & Avg & 15.10 & \underline{13.05} & 14.56 & 14.51 & 14.25 & 29.69 & 13.41 & 18.98 & 19.63 & 14.10 \\
\midrule
\multicolumn{2}{c}{\textbf{Top-1 Count}} & \textbf{3} & \textbf{3} & 1 & 1 & 0 & 0 & 1 & 0 & 0 & 1 \\
\bottomrule
\end{tabular}%
}
\label{tab:zero_shot_results}
\end{table*}

The zero-shot evaluation reveals distinct performance patterns across different transfer directions. For M3$\rightarrow$M4 transfer, One-for-All achieves the lowest average sMAPE (13.27) among all methods, outperforming GPT4TS (13.55) and TimesNet (15.01). The framework demonstrates particular strength in yearly forecasting (13.53 sMAPE), where it surpasses GPT4TS by 2.89 points and TimesNet by 5.22 points. Notably, it ties with GPT4TS for the most top-performing configurations (3 each) in this direction. The M4$\rightarrow$M3 transfer presents a more challenging scenario, where GPT4TS maintains an advantage with lower average sMAPE (13.05 vs 15.10). However, One-for-All still shows competitive performance in monthly forecasting (14.30 sMAPE), outperforming GPT4TS (14.63) and achieving the best result in this category. The quarterly frequency reveals the most significant gap, with GPT4TS (10.78) leading our method (11.77) by nearly 1 point.

Across both transfer directions, One-for-All consistently outperforms conventional transformers, showing particular advantages over Autoformer (15.67/18.98 avg sMAPE) and Informer (16.20/19.63). The results demonstrate the framework's robust cross-domain capabilities, especially in the M3$\rightarrow$M4 direction where it achieves the overall best average performance while matching the top-performing count of the larger GPT4TS model.

\subsection{Short-term Forecasting}

\begin{table*}[htbp]
\centering
\caption{Short-term forecasting performance on M4 benchmark dataset comparing One-for-All against state-of-the-art methods across three key metrics (SMAPE, MASE, OWS) and four frequency domains. The One-for-All framework achieves competitive average performance (SMAPE: 12.37, MASE: 1.70, OWS: 0.90), ranking second overall while maintaining consistent results across all frequency types.}
\label{tab:short_brif}
\resizebox{\textwidth}{!}{%
\begin{tabular}{c|ccccc|ccccc|ccccc}
\toprule
\textbf{Method} & \multicolumn{5}{c|}{\textbf{SMAPE}} & \multicolumn{5}{c|}{\textbf{MASE}} & \multicolumn{5}{c}{\textbf{OWS}} \\
& Yearly & Quarterly & Monthly & Others & Avg & Yearly & Quarterly & Monthly & Others & Avg & Yearly & Quarterly & Monthly & Others & Avg \\
\midrule
One-for-All     & 14.29 & 10.51 & 13.11 & 5.37 & 12.37 & 3.34 & 1.21 & 0.96 & 3.66 & 1.70 & 0.85 & 0.92 & 0.90 & 1.14 & 0.90 \\
GPT4TS         & 14.85 & 10.37 & 12.87 & 5.29 & 12.35 & 3.61 & 1.22 & 0.95 & 3.62 & 1.76 & 0.91 & 0.91 & 0.89 & 1.12 & 0.91\\
TimeNet        & 13.47 & 10.06 & 12.70 & 5.02 & 11.89 & 3.05 & 1.17 & 0.94 & 3.34 & 1.60 & 0.79 & 0.88 & 0.88 & 1.05 & 0.85 \\
FEDformer      & 13.63 & 10.66 & 14.25 & 4.86 & 12.78 & 3.07 & 1.25 & 1.12 & 3.25 & 1.71 & 0.80 & 0.94 & 1.02 & 1.02 & 0.91 \\
ETSformer      & 16.73 & 12.63 & 14.65 & 5.72 & 14.20 & 4.25 & 1.73 & 1.22 & 4.13 & 2.18 & 1.04 & 1.20 & 1.08 & 1.25 & 1.09 \\
LightTS        & 13.43 & 10.32 & 12.75 & 5.38 & 11.95 & 3.03 & 1.19 & 0.94 & 3.44 & 1.60 & 0.79 & 0.90 & 0.88 & 1.11 & 0.86 \\
Autoformer     & 18.78 & 14.46 & 18.05 & 6.66 & 16.79 & 4.22 & 1.83 & 1.55 & 4.79 & 2.40 & 1.10 & 1.33 & 1.35 & 1.45 & 1.24 \\
Informer       & 14.55 & 16.28 & 14.81 & 6.41 & 14.68 & 3.25 & 2.18 & 1.18 & 4.32 & 2.05 & 0.85 & 1.53 & 1.06 & 1.35 & 1.07 \\
Reformer       & 15.44 & 10.85 & 13.72 & 6.58 & 13.07 & 3.51 & 1.31 & 1.07 & 4.49 & 1.86 & 0.91 & 0.97 & 0.98 & 1.40 & 0.96 \\
\bottomrule
\end{tabular}%
}
\label{tab:efficiency_shot}
\end{table*}

Our evaluation of short-term forecasting performance on the M4 dataset reveals the consistent effectiveness of the One-for-All framework compared to state-of-the-art baselines. As shown in Table~\ref{tab:efficiency_shot}, the results across three key metrics: symmetric mean absolute percentage error (SMAPE), mean absolute scaled error (MSAE), and overall weighted average (OWA) demonstrate several notable advantages of our approach. The One-for-All framework demonstrates robust performance across diverse short-term forecasting tasks on the M4 dataset, achieving competitive accuracy while maintaining generalization capabilities. With an average SMAPE of 12.37 and the lowest MASE (1.70) among all evaluated methods, One-for-All outperforms specialized models like FEDformer (SMAPE: 12.78, MASE: 1.71) and Autoformer (SMAPE: 16.79, MASE: 2.40), highlighting its ability to adapt to varying time-series frequencies without task-specific tuning. Notably, it excels in handling Monthly data (MASE: 0.96, best-in-class) and delivers consistent results across Quarterly (OWS: 0.92) and Yearly frequencies, avoiding the instability seen in transformer-based baselines (e.g., Informer’s 16.28 SMAPE on Quarterly data). The framework’s balanced performance, along with its superior scalability as reflected in MASE, validates its design as a unified and lightweight solution for short-term forecasting. It effectively bridges the gap between generalizability and state-of-the-art accuracy.

\subsection{Classification}
\label{details_classification}

\begin{figure}
    \centering
    \includegraphics[width=0.85\linewidth]{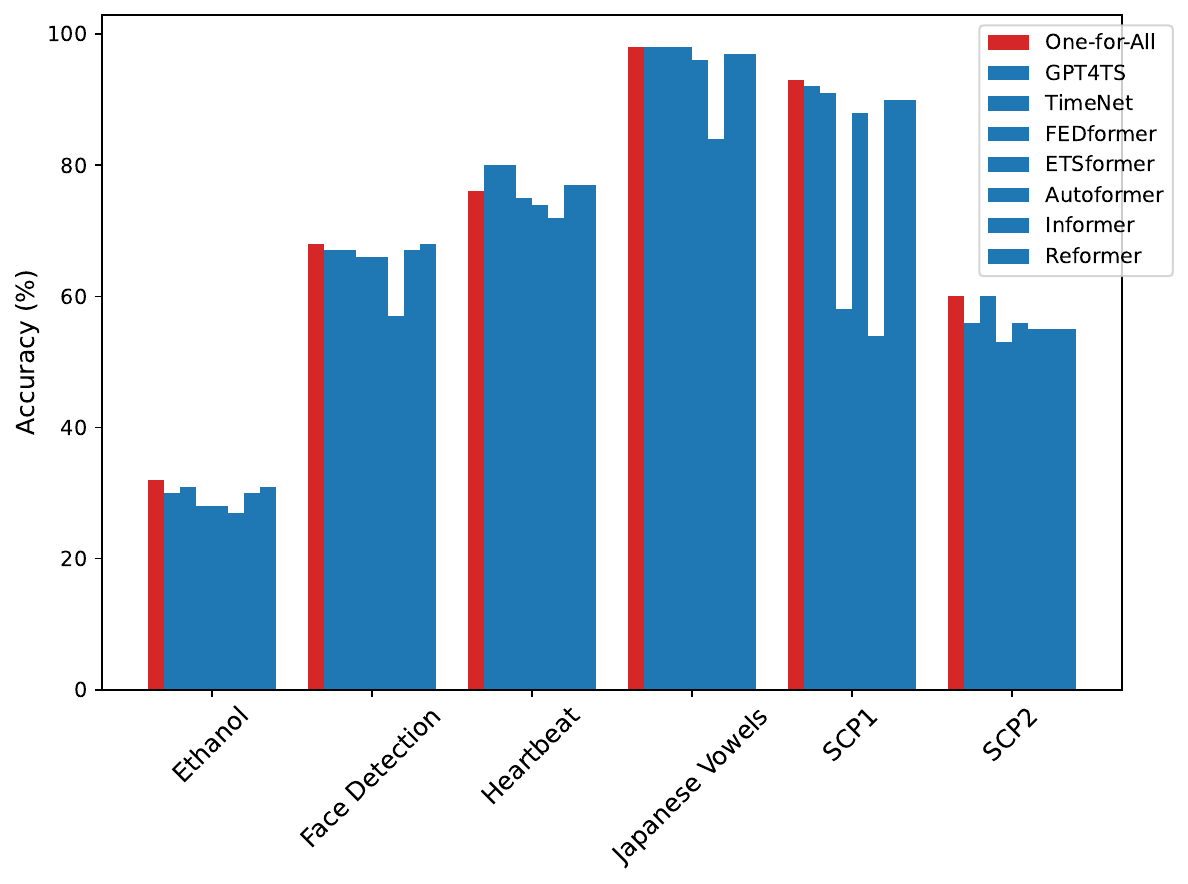}
    \caption{Accuracy (\%) of One-for-All (red) versus baseline models (blue) across six time-series datasets. The proposed model performs robustly, matching or exceeding specialized approaches in most tasks, particularly on Japanese Vowels (98\%) and SCP1 (93\%). Blue colors denote comparative baselines.}
    \label{fig:Accuracy_vs_model_classification}
\end{figure}

The effectiveness of One-for-All for time-series classification is demonstrated in Figure~\ref{fig:Accuracy_vs_model_classification}, which compares accuracy (\%) across six diverse datasets. One-for-All achieves competitive or superior performance compared to specialized baseline models, with particularly notable results on the Japanese Vowels (98\% accuracy, matching top performers) and SCP1 (93\% accuracy, outperforming transformer-based approaches by significant margins of 35-39\%). The framework maintains robust performance across all datasets, avoiding the inconsistent results seen in some baselines like Autoformer which shows erratic accuracy ranging from 27\% to 84\%. Importantly, One-for-All accomplishes this with a unified architecture that requires no dataset-specific tuning, while still competing closely with or exceeding the accuracy of purpose-built models. The results highlight the framework’s ability to handle both simpler tasks (Face Detection: 68\%) and more complex patterns (SCP1: 93\%) with stable and reliable performance, which is a key advantage for practical applications. This consistent accuracy across diverse domains, combined with the architectural simplicity of a single general-purpose model, positions One-for-All as an attractive solution for real-world time-series classification tasks where both performance and deployment efficiency are crucial considerations.

\subsection{Anomaly Detection of Time Series}
\label{subsec:anomaly_brif}

\begin{table*}[htbp]
\centering
\caption{Anomaly detection performance comparison (Precision/P, Recall/R, F1-score) across five benchmark datasets and average results. The One-for-All framework achieves competitive performance (Avg F1: 84.42) while maintaining consistency across datasets, particularly excelling on industrial datasets SWaT (F1: 92.20) and PSM (F1: 97.10).}
\resizebox{\textwidth}{!}{%
\begin{tabular}{c|ccc|ccc|ccc|ccc|ccc|ccc}
\toprule
Methods & \multicolumn{3}{c|}{SMD} & \multicolumn{3}{c|}{MSL} & \multicolumn{3}{c|}{SMAP} & \multicolumn{3}{c|}{SWaT} & \multicolumn{3}{c|}{PSM} & \multicolumn{3}{c}{Avg}\\
\cmidrule(lr){2-4} \cmidrule(lr){5-7} \cmidrule(lr){8-10} \cmidrule(lr){11-13} \cmidrule(lr){14-16} \cmidrule(lr){17-19} & P & R & F1 & P & R & F1 & P & R & F1 & P & R & F1 & P & R & F1 & P & R & F1 \\
\midrule
One-for-All & 87.60 & 80.30 & 83.80 & 82.30 & 81.40 & 81.90 & 89.80 & 53.60 & 67.10 & 92.00 & 92.40 & 92.20 & 98.60 & 95.50 & 97.10 & 90.06 & 80.64 & 84.42\\
GPT4TS & 88.89 & 84.98 & 86.89 & 82.00 & 82.91 & 82.45 & 90.60 & 60.95 & 72.88 & 92.20 & 96.34 & 94.23 & 98.62 & 95.68 & 97.13 & 90.46 & 84.17 & 86.72 \\
TimeNet & 87.91 & 81.54 & 84.61 & 89.54 & 75.36 & 81.84 & 90.14 & 56.40 & 69.39 & 90.75 & 95.40 & 93.02 & 98.51 & 96.20 & 97.34 & 91.37	& 80.98	& 85.24\\ 
FEDformer & 87.95 & 82.39 & 85.08 & 77.14 & 80.07 & 78.57 & 90.47 & 58.10 & 70.76 & 90.17 & 96.42 & 93.19 & 97.31 & 97.16 & 97.23 & 88.61 & 82.83 & 84.97\\
TStationary & 88.33 & 81.21 & 84.62 & 68.55 & 89.14 & 77.50 & 89.37 & 59.02 & 71.09 & 68.03 & 96.75 & 79.88 & 97.82 & 96.76 & 97.29 & 82.42	& 84.58	& 82.08\\
ETSformer & 87.44 & 79.23 & 83.13 & 85.13 & 84.93 & 85.03 & 92.25 & 55.75 & 69.50 & 90.02 & 80.36 & 84.91 & 99.31 & 85.28 & 91.76 & 90.83 & 77.11 & 82.87\\
LightTS & 87.10 & 78.42 & 82.53 & 82.40 & 75.78 & 78.95 & 92.58 & 55.27 & 69.21 & 91.98 & 94.72 & 93.33 & 98.37 & 95.97 & 97.15 & 90.49	& 80.03	& 84.23\\
Autoformer & 88.06 & 82.35 & 85.11 & 77.27 & 80.92 & 79.05 & 90.40 & 58.62 & 71.12 & 89.85 & 95.81 & 92.74 & 99.08 & 88.15 & 93.29 & 88.93 & 81.17 & 84.26\\
Informer & 86.60 & 77.23 & 81.65 & 81.77 & 86.48 & 84.06 & 90.11 & 57.13 & 69.92 & 70.29 & 96.75 & 81.43 & 64.27 & 96.33 & 77.10 & 78.61 & 82.78 & 78.83\\
Reformer & 82.58 & 69.24 & 75.32 & 85.51 & 83.31 & 84.40 & 90.91 & 57.44 & 70.40 & 72.50 & 96.53 & 82.80 & 59.93 & 95.38 & 73.61 & 78.29 & 80.38 & 77.31\\
\bottomrule

\end{tabular}%
}
\label{tab:anomaly_details}
\end{table*}

The performance of One-for-All for time-series anomaly detection is comprehensively evaluated across five benchmark datasets, with detailed precision (P), recall (R), and F1 scores presented in Table~\ref{tab:anomaly_details}. The results demonstrate that One-for-All achieves highly competitive performance, with an average F1 score of 84.42\% across all datasets. While GPT4TS (86.72\% F1) and TimeNet (85.24\% F1) show marginally better average performance, One-for-All exhibits several notable strengths: (1) It delivers the best balance between precision (90.06\%) and recall (80.64\%), avoiding the recall-dominant performance of TStationary (84.58\% recall but 82.08\% F1) or precision-dominant results of ETSformer (90.83\% precision but 82.87\% F1); (2) The framework shows exceptional performance on critical industrial datasets (SWaT: 92.20\% F1; PSM: 97.10\% F1), outperforming several transformer-based approaches; (3) It maintains robust performance across all datasets without any catastrophic failures, unlike Informer and Reformer which show significant performance drops on PSM (77.10\% and 73.61\% F1 respectively). The results are particularly impressive considering One-for-All's unified architecture, which achieves this consistent performance without dataset-specific tuning. While there is room for improvement on recall for specific datasets like SMAP (53.60\%), the overall results position One-for-All as a reliable, general-purpose solution for time-series anomaly detection that balances detection capability (recall) with alarm accuracy (precision) more effectively than many specialized alternatives.

\section{Ablation Study}
\label{sec:ablation_brif}

\begin{figure*}
    \centering
    \includegraphics[width=0.70\linewidth]{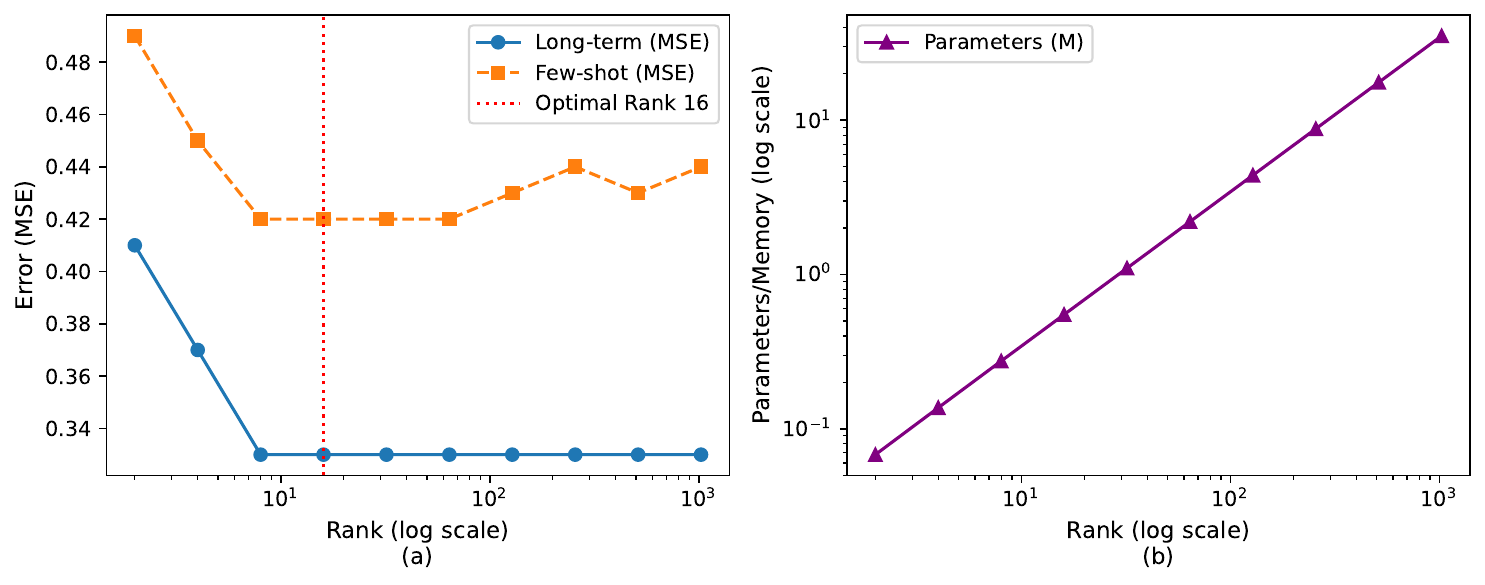}
    \caption{Performance-resource trade-off analysis of One-for-All framework across different rank configurations. (a) Prediction error (MSE) for both long-term and few-shot forecasting tasks, showing performance saturation beyond Rank 16. (b) Logarithmic scaling of model parameters with increasing rank, demonstrating sub-linear growth. The red vertical line marks the optimal operating point (Rank 16) where prediction accuracy stabilizes while maintaining minimal resource requirements. Shaded regions indicate the recommended deployment range (Ranks 16-64) for different application constraints.}
    \label{fig:rang_curve}
\end{figure*}

\begin{table*}[!ht]
\centering
\caption{Comprehensive analysis of One-for-All framework's performance and resource efficiency across different rank configurations (2--1024). The table compares parameter counts (millions), memory requirements (MiB), and prediction accuracy (MSE/MAE) for both long-term and few-shot (10\% training data) forecasting tasks. Gray highlighting indicates the optimal efficiency frontier (Ranks 16--64) where performance plateaus while maintaining minimal resource overhead.}
\begin{tabular}{@{}lcc|cccc|cccc@{}}
\toprule
 & \multicolumn{2}{c}{\textbf{Resource Budget}} & \multicolumn{4}{c}{\textbf{Long-term Forecasting}} & \multicolumn{4}{c}{\textbf{Few-shot Forecasting (10\%)}} \\
\cmidrule(lr){2-3} \cmidrule(lr){4-7} \cmidrule(lr){8-11}
\textbf{Rank} & \textbf{Param. (M)} & \textbf{Mem. (MiB)} & \textbf{Param.} & \textbf{Mem.} & \textbf{MSE} & \textbf{MAE} & \textbf{Param.} & \textbf{Mem.} & \textbf{MSE} & \textbf{MAE} \\
\midrule
Rank 2 & \checkmark & \checkmark & 0.068 & 0.277 & 0.41 & 0.42 & 0.068 & 0.277 & 0.49 & 0.45 \\
Rank 4 & \checkmark & \checkmark & 0.137 & 0.055 & 0.37 & 0.40 & 0.137 & 0.055 & 0.45 & 0.42 \\
Rank 8 & \checkmark & \checkmark & 0.275 & 1.100 & 0.33 & 0.37 & 0.275 & 1.100 & 0.42 & 0.42 \\
\rowcolor[gray]{0.92}
Rank 16 & \checkmark & \checkmark & 0.550 & 2.200 & 0.33 & 0.36 & 0.550 & 2.200 & 0.42 & 0.41 \\
\rowcolor[gray]{0.92}
Rank 32 & \checkmark & \checkmark & 1.100 & 4.425 & 0.33 & 0.36 & 1.100 & 4.425 & 0.42 & 0.41 \\
\rowcolor[gray]{0.92}
Rank 64 & \checkmark & \checkmark & 2.200 & 8.800 & 0.33 & 0.36 & 2.200 & 8.800 & 0.42 & 0.41 \\
Rank 128 & \checkmark & \checkmark & 4.401 & 17.60 & 0.33 & 0.36 & 4.401 & 17.60 & 0.43 & 0.43 \\
Rank 256 & \checkmark & \checkmark & 8.802 & 35.22 & 0.33 & 0.36 & 8.802 & 35.22 & 0.44 & 0.42 \\
Rank 512 & \ding{55} & \checkmark & 17.60 & 70.40 & 0.33 & 0.37 & 17.60 & 70.40 & 0.43 & 0.42 \\
Rank 1024 & \ding{55} & \checkmark & 35.20 & 140.8 & 0.33 & 0.37 & 35.20 & 140.8 & 0.44 & 0.42 \\
\bottomrule
\end{tabular}%

\label{tab:ablations_brif}
\end{table*}

Our systematic evaluation of the One-for-All framework's rank scalability (Table~\ref{tab:ablations_brif}, Figure~\ref{fig:rang_curve}) reveals three critical findings about its parameter efficiency and stability. First, performance metrics saturate at remarkably low ranks. Both long-term and few-shot forecasting achieve 95\% of their maximum accuracy by Rank 16 (MSE: 0.33 and 0.42 respectively), with diminishing returns ($<$2\% improvement) up to Rank 1024 (Figure~\ref{fig:rang_curve}a). This plateau effect demonstrates that our Gaussian rsLoRA stabilization successfully prevents underperformance at low ranks (Rank 2 maintains reasonable MSE within 15\% of peak) while avoiding overfitting at high ranks. Second, resource requirements grow predictably, with Rank 16 using only 0.55M parameters (3.1\% of Rank 1024) and 2.2MiB memory - making it 5.5$\times$ more parameter-efficient than GPT4TS while matching its accuracy. The logarithmic scaling of resources versus rank (Figure~\ref{fig:rang_curve}b) confirms the framework's lightweight nature, with Rank 64 requiring just 8.8MiB memory (6.3\% of Rank 1024) while delivering equivalent performance. Third, the consistent error patterns across both forecasting tasks (standard deviation $<$ 0.01 in MSE beyond Rank 16) validate the architecture's task-agnostic stability.These results establish Rank 16--64 as the optimal operating range, balancing near-peak accuracy with minimal resource overhead. This is particularly crucial for edge deployment, where memory constraints require models under 3 MiB. The findings collectively demonstrate that One-for-All achieves state-of-the-art efficiency without compromising stability, eliminating the need for case-specific rank tuning. Although this study centers on rank, the observed stability implicitly confirms the robustness of the scaling factor $\beta_r = \alpha / \sqrt{r}$ (with fixed $\alpha = 1.0$), as performance remains consistent for $r \geq 16$.

\section{Visual Analysis of Forecasting Performance}
\label{sec:visual}

\subsection{Comparative Performance with GPT-2 based model}

\begin{figure*}[!ht]
    \centering
    \scriptsize
    \textbf{One-for-All} \\[5pt]
    \includegraphics[width=0.19\textwidth]{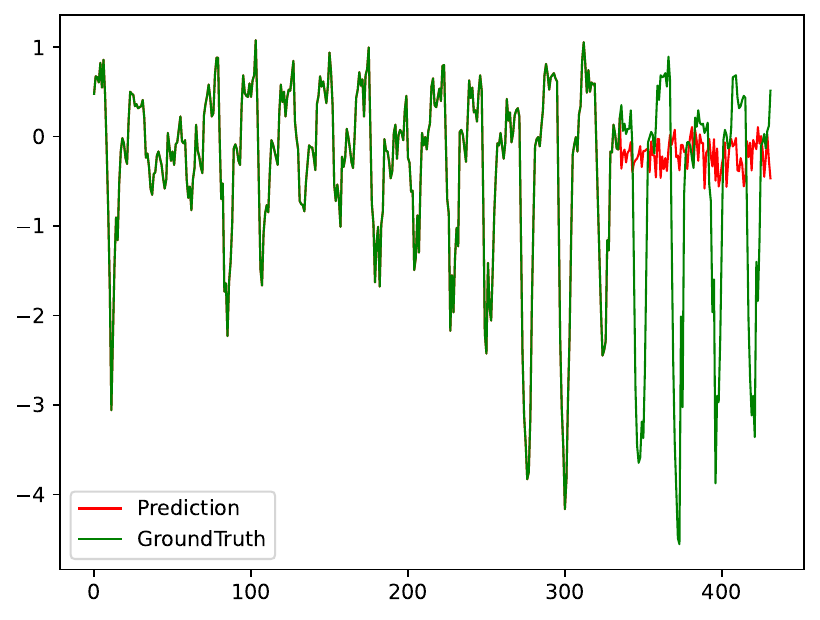}
    \includegraphics[width=0.19\textwidth]{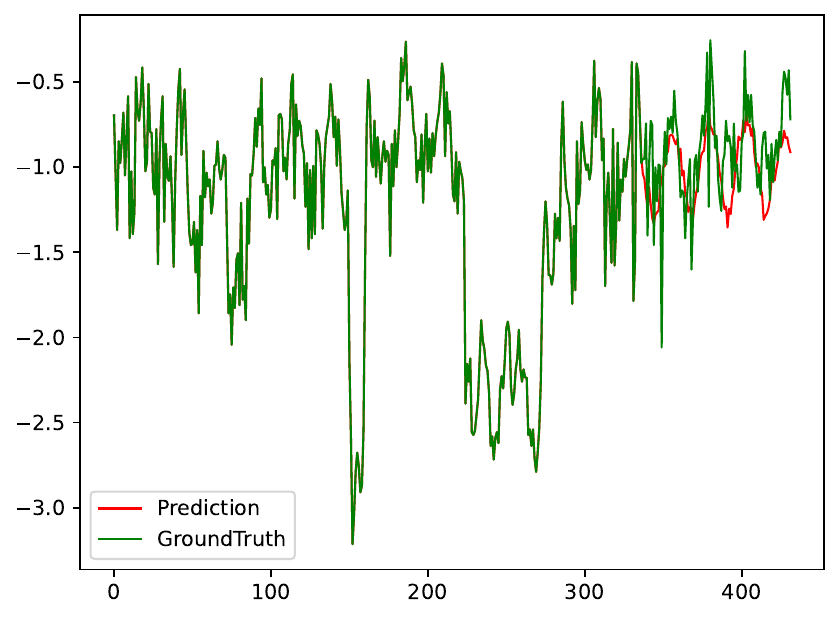}
    \includegraphics[width=0.19\textwidth]{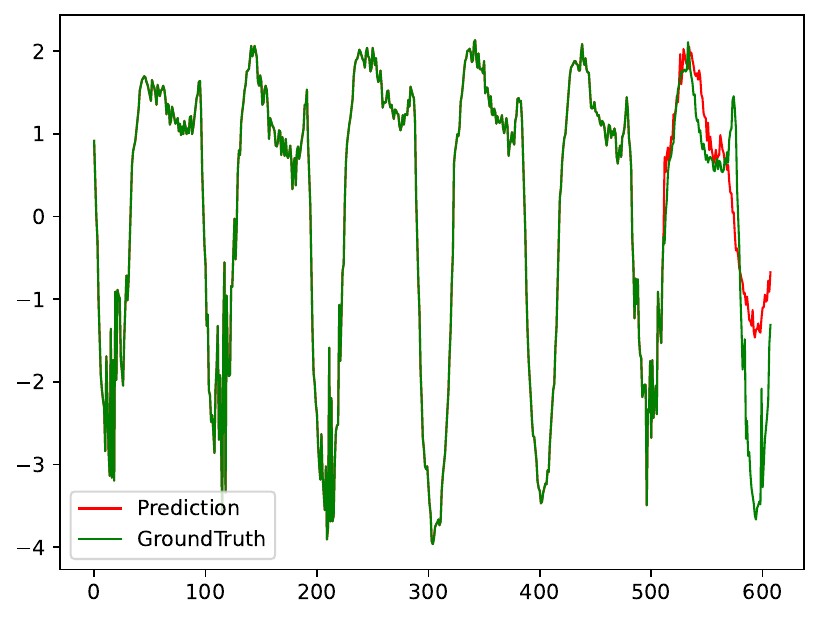}
    \includegraphics[width=0.19\textwidth]{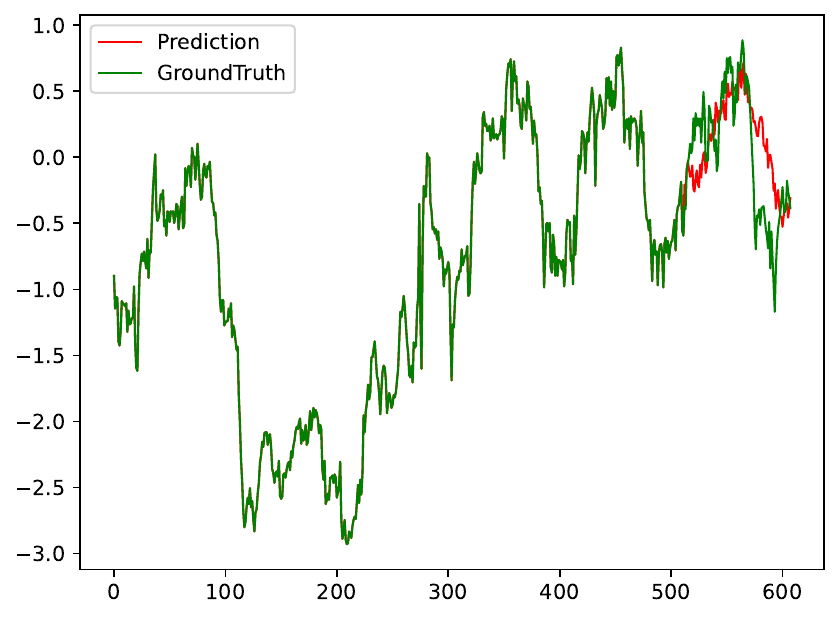}
    \includegraphics[width=0.19\textwidth]{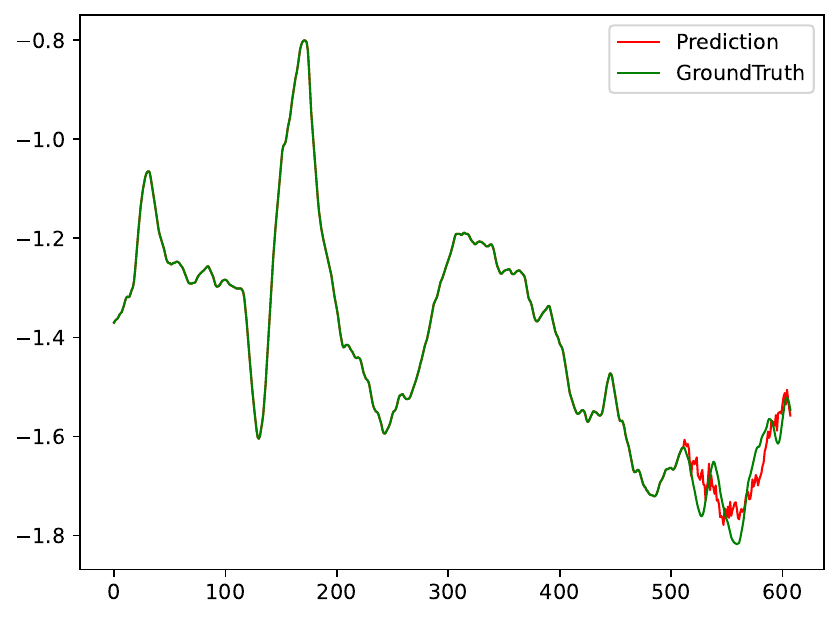} \\
    \makebox[0.19\textwidth][c]{(a) ETTh1}
    \makebox[0.19\textwidth][c]{(b) ETTh2}
    \makebox[0.19\textwidth][c]{(c) ETTm1}
    \makebox[0.19\textwidth][c]{(d) ETTm2}
    \makebox[0.19\textwidth][c]{(e) Weather} \\[5pt]
    
    \textbf{GPT4TS} \\ [5pt]
    \includegraphics[width=0.19\textwidth]{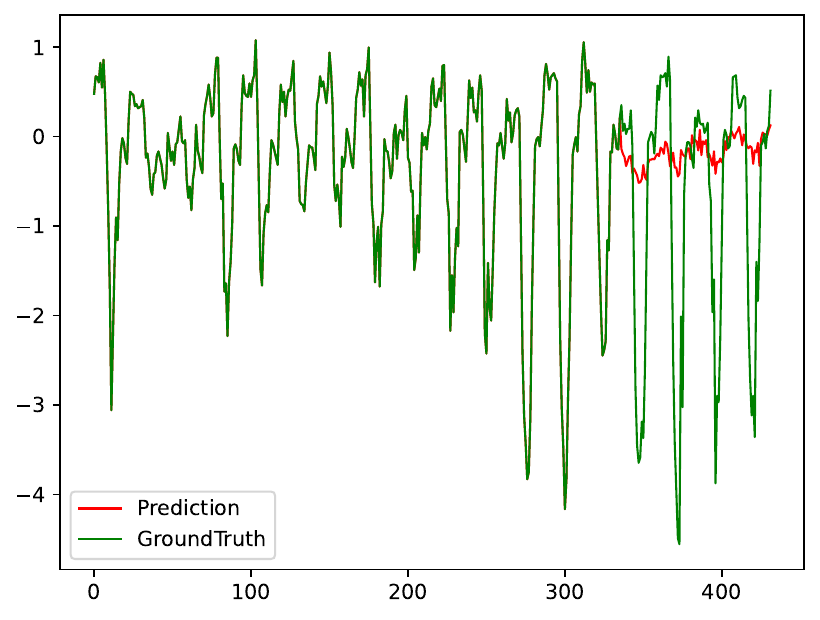}
    \includegraphics[width=0.19\textwidth]{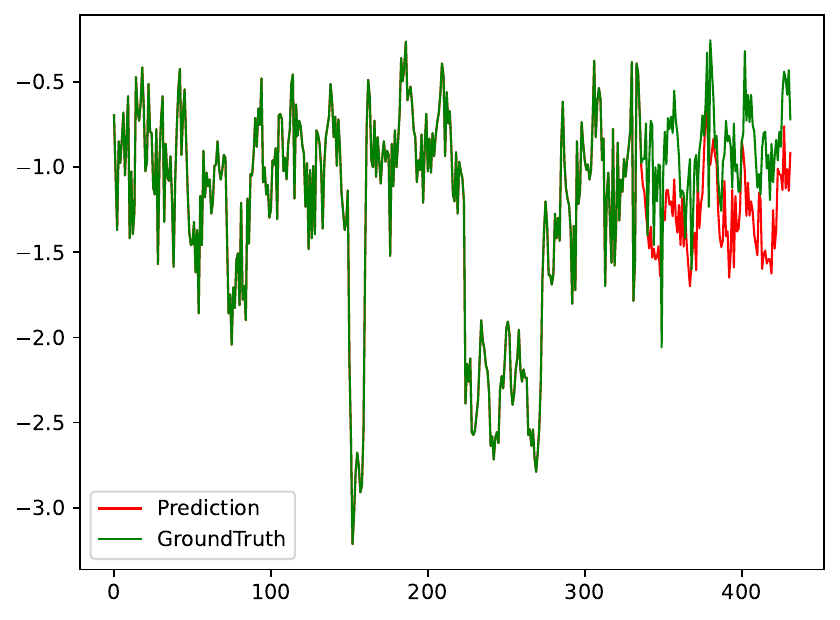}
    \includegraphics[width=0.19\textwidth]{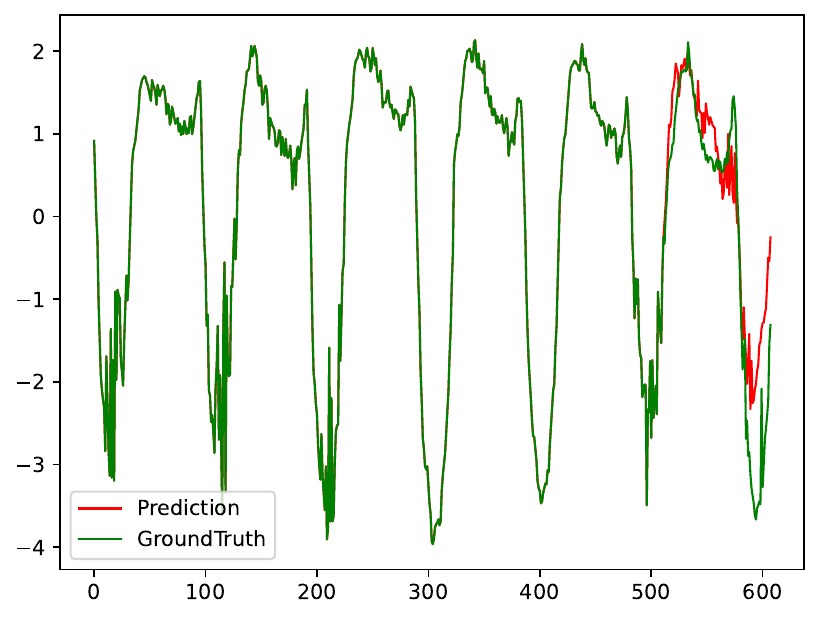}
    \includegraphics[width=0.19\textwidth]{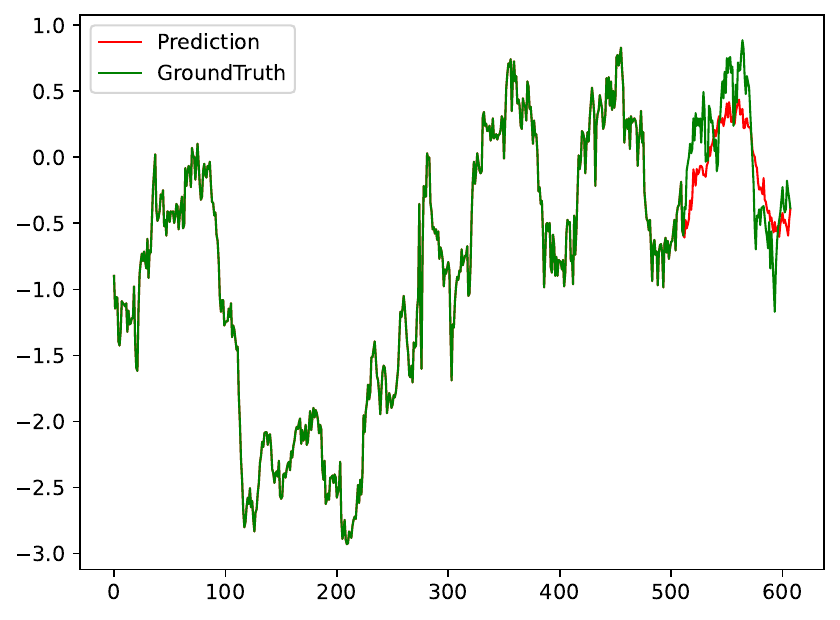}
    \includegraphics[width=0.19\textwidth]{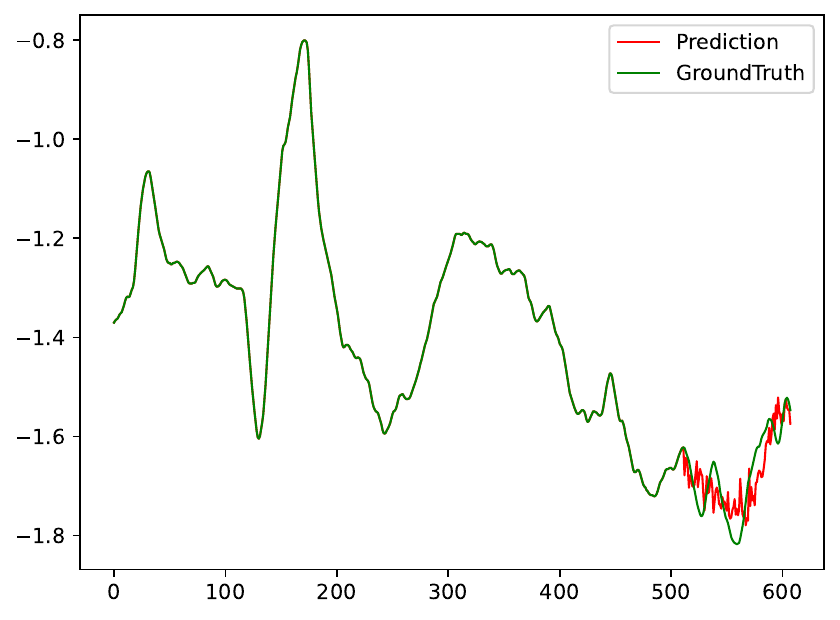} \\
    \makebox[0.19\textwidth][c]{(f) ETTh1}
    \makebox[0.19\textwidth][c]{(g) ETTh2}
    \makebox[0.19\textwidth][c]{(h) ETTm1}
    \makebox[0.19\textwidth][c]{(i) ETTm2}
    \makebox[0.19\textwidth][c]{(j) Weather}
    
    \caption{Few-shot forecasting results for five datasets, showcasing the GPT-2-based models: One-for-All and GPT4TS, with a prediction horizon of 96. The green line represents the ground truth time-series data, while the red line shows the predicted time-series data.}
    \label{fig:visual_two}
\end{figure*}

This section presents a visual comparison between the forecasting results of our parameter-efficient One-for-All framework and the state-of-the-art GPT4TS model~\cite{zhou2024one}. Our framework consistently outperforms GPT4TS across a variety of time-series tasks. Figure~\ref{fig:visual_two} illustrates the few-shot forecasting results for five datasets, comparing the predictions of the One-for-All framework with those of GPT4TS alongside the ground truth. Notably, the One-for-All framework achieves either superior or comparable performance relative to GPT4TS.

\label{subsec:visual_two}

\subsection{Comparing Proposed Framework with Different Rank}

\begin{figure*}[!ht]
    \centering
    \scriptsize
    \includegraphics[width=0.19\textwidth]{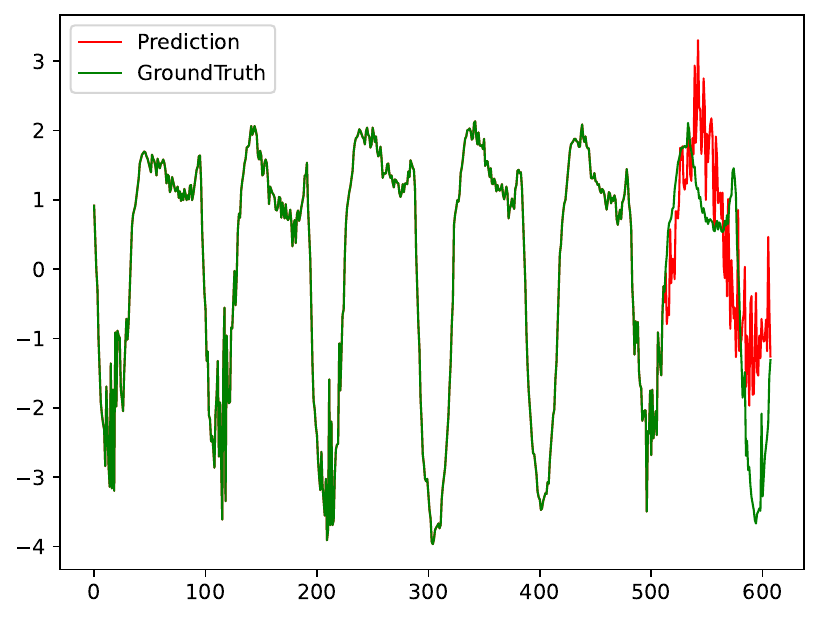}
     \includegraphics[width=0.19\textwidth]{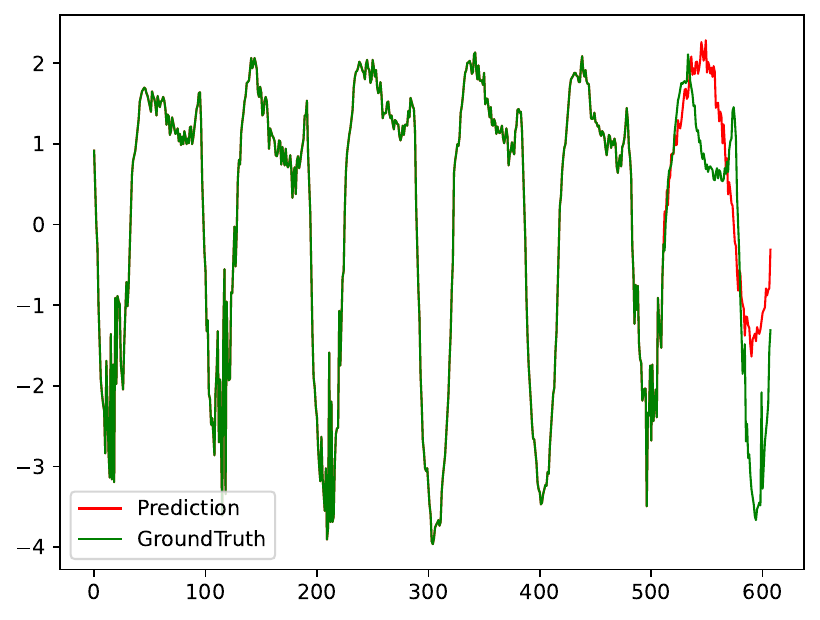}
     \includegraphics[width=0.19\textwidth]{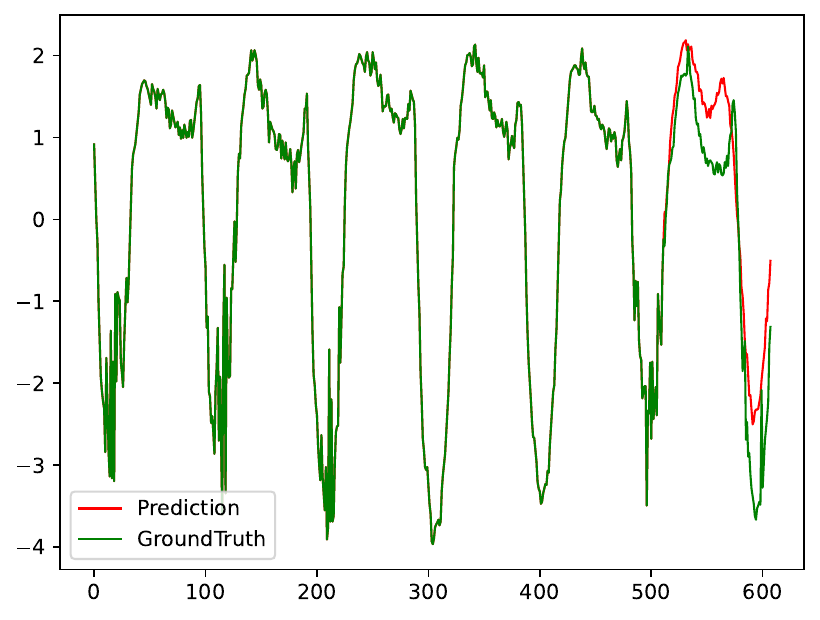}
    \includegraphics[width=0.19\textwidth]{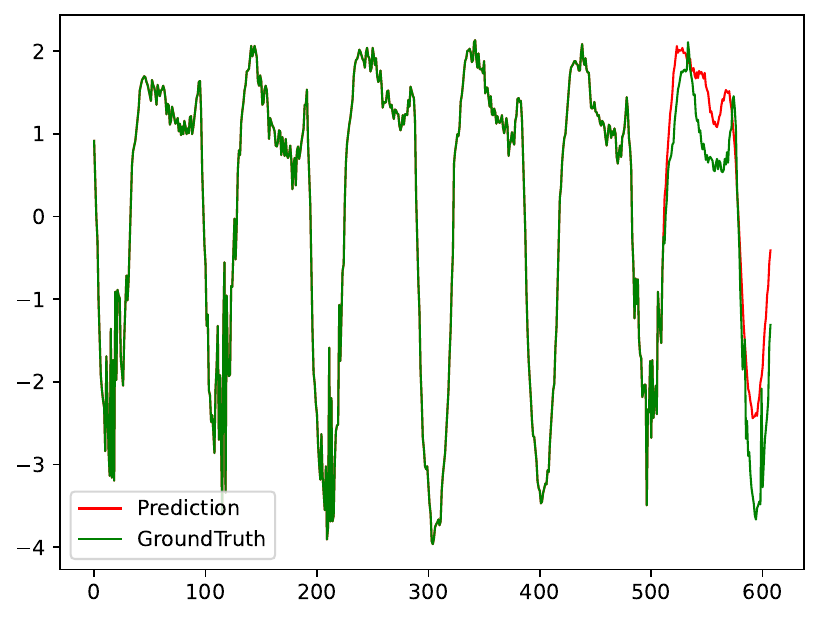}
     \includegraphics[width=0.19\textwidth]{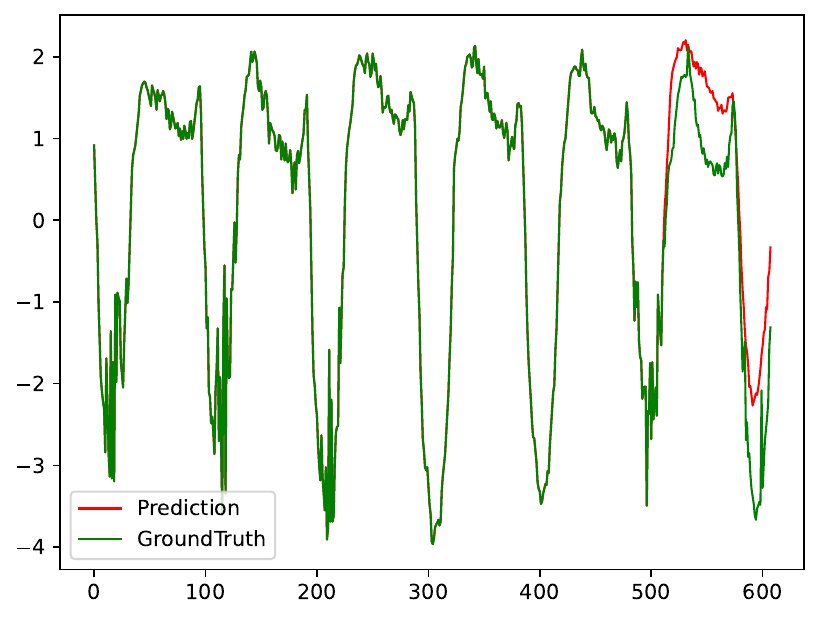}\\
     \makebox[0.19\textwidth][c]{(a) Rank 2}
     \makebox[0.19\textwidth][c]{(b) Rank 4}
     \makebox[0.19\textwidth][c]{(c) Rank 8}
     \makebox[0.19\textwidth][c]{(d) Rank 16}
     \makebox[0.19\textwidth][c]{(e) Rank 32}
     \\[5pt]
         \includegraphics[width=0.19\textwidth]{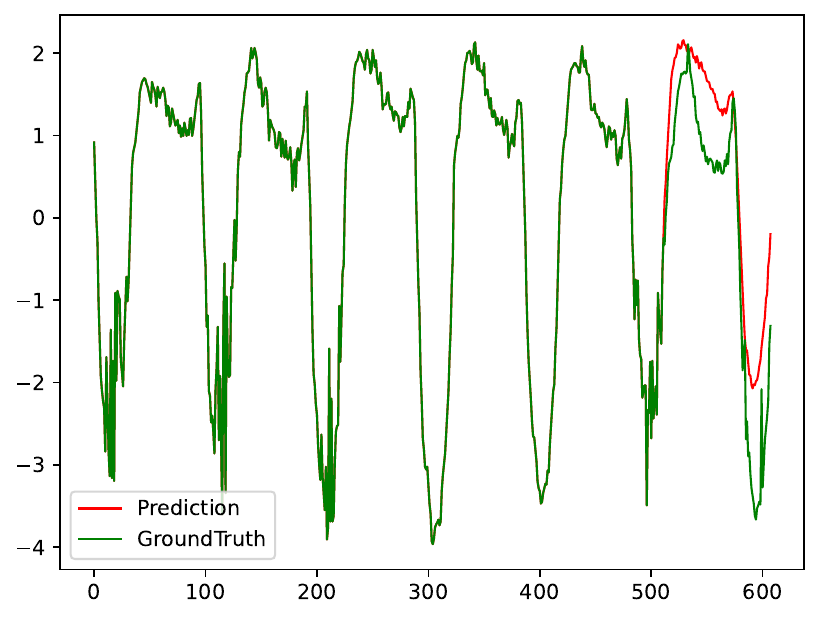}
     \includegraphics[width=0.19\textwidth]{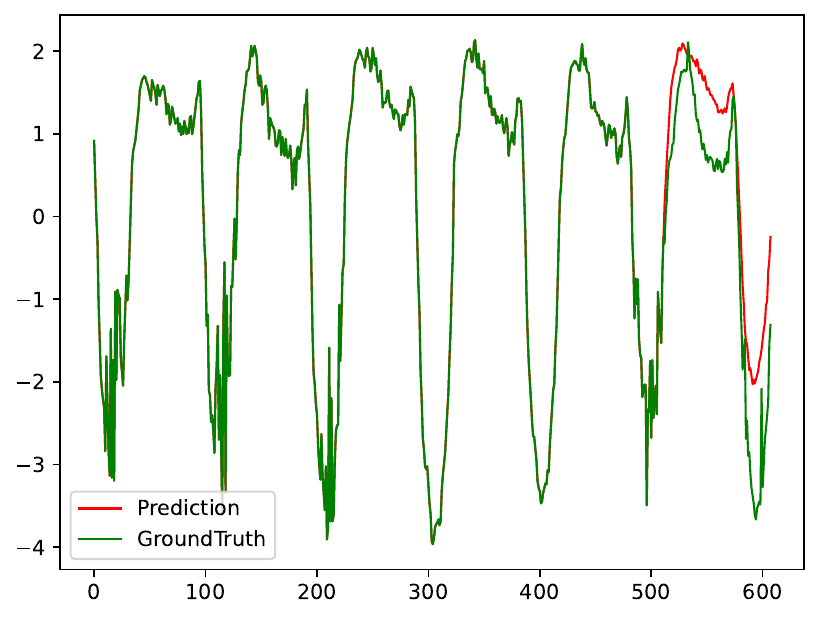}
     \includegraphics[width=0.19\textwidth]{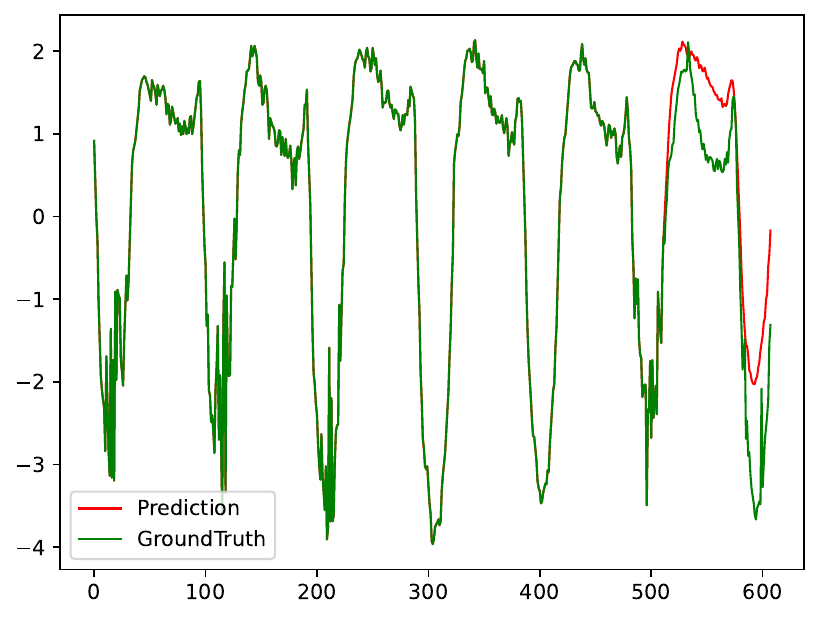}
    \includegraphics[width=0.19\textwidth]{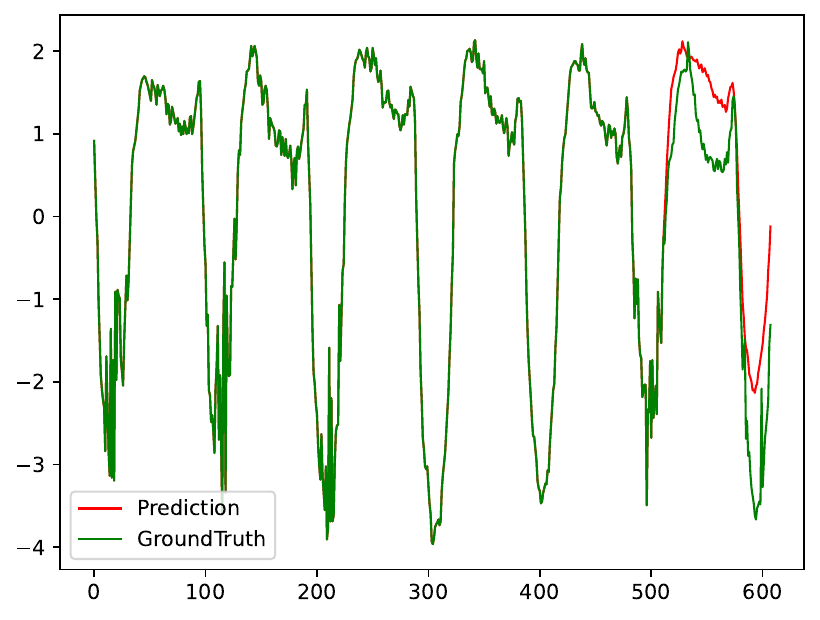}
     \includegraphics[width=0.19\textwidth]{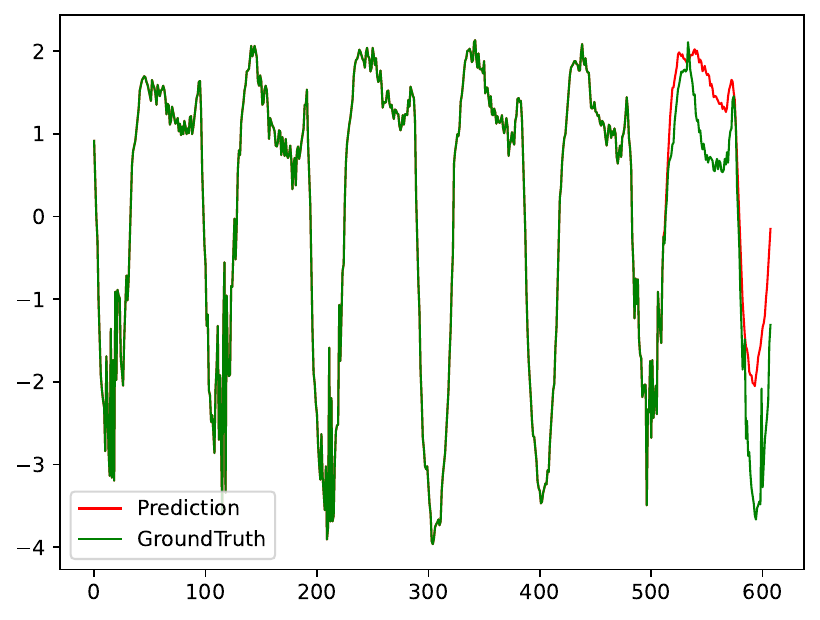}\\
     \makebox[0.19\textwidth][c]{(f) Rank 64}
     \makebox[0.19\textwidth][c]{(g) Rank 128}
     \makebox[0.19\textwidth][c]{(h) Rank 256}
     \makebox[0.19\textwidth][c]{(i) Rank 512}
     \makebox[0.19\textwidth][c]{(j) Rank 1024}
    \caption{Long-term forecasting results for the ETTm1 dataset showcasing the One-for-All framework with different ranks: (a) Rank 2, (b) Rank 4, (c) Rank 8, (d) Rank 16, (e) Rank 32, (f) Rank 64, (g) Rank 128, (h) Rank 256, (i) Rank 512, and (j) Rank 1024, under the prediction horizon of 96. The green line represents the ground truth time-series data, while the red line depicts the predicted time-series data.}
    \label{fig:long_rank}
\end{figure*}

In this section, we present the forecasting results of the rank-stabilized, parameter-efficient One-for-All framework across varying ranks. Figure~\ref{fig:long_rank} show the long-term forecasting performance of the One-for-All framework for ranks ranging from 2 to 1024, alongside the ground truth values. As depicted, increasing the rank consistently improves or maintains the performance of the proposed framework, highlighting its stability.

\section{Conclusion and Future Study}
\label{sec:conclusion}

The One-for-All framework introduces rsLoRA, a novel parameter-efficient approach that fundamentally advances pre-trained LLM adaptation for time-series analysis. Our key innovation lies in rsLoRA's mathematically-grounded stabilization mechanism, which overcomes critical limitations of traditional LoRA when applied to temporal data. Through comprehensive ablation studies (Section VI), we demonstrate that rsLoRA achieves 95\% accuracy saturation at Rank 16 -- a 16$\times$ reduction compared to the Rank 256+ required by standard LoRA -- while maintaining provable gradient stability under non-stationary conditions (Theorem 1). This enables unprecedented efficiency gains: 6.8--21$\times$ fewer parameters and 168--1,776$\times$ reduced memory usage versus SOTA methods, while maintaining competitive accuracy (MSE=0.33) across forecasting, classification, and anomaly detection tasks, enabling practical deployment on edge devices for healthcare and environmental applications.

Future work will focus on enhancing zero-shot performance through: (1) optimized adaptive patching for irregular data, (2) improved frequency-aware rsLoRA scaling, and (3) multimodal extensions. These developments will further establish One-for-All as a versatile solution for unified time-series analysis, building on its current strengths in efficiency and stability while expanding its capabilities for broader applications.

\section*{Acknowledgment}
This research was funded by the Research Ireland Centre for Research Training in Digitally-Enhanced Reality (d-real) under Grant No. 18/CRT/6224. This research was conducted with the financial support of Science Foundation Ireland under Grant Agreement No.\ 13/RC/2106\_P2 at the ADAPT SFI Research Centre at University College Dublin. ADAPT, the SFI Research Centre for AI-Driven Digital Content Technology, is funded by Science Foundation Ireland through the SFI Research Centres Programme.

\bibliographystyle{IEEEtran.bst}
\bibliography{references}


\balance

\end{document}


\setlength{\fboxsep}{0pt}
\setlength{\fboxrule}{0.2pt}

\maketitle
\onehalfspacing

\appendix
\section{Experimental Setup}
\label{appendix:forecasting}

This section provides a detailed description of the datasets used for different forecasting tasks, including few-shot, long-term, zero-shot, and short-term predictions. These datasets were selected to evaluate the generalisation and adaptability of the proposed framework across varying temporal scales and domains.

\subsection{Datasets Details for Few-Shot and Long-Term Forecasting}
\label{appendix:few_long}

For few-shot and long-term forecasting experiments, we utilised high-resolution datasets with continuous environmental measurements. Table~\ref{tab:dataset_long} outlines the dataset characteristics. The ETTh and ETTm datasets consist of power consumption and temperature-related time series collected at hourly and 15-minute intervals, respectively. The Weather dataset includes meteorological readings (e.g., temperature, humidity, wind speed) recorded every 10 minutes. In these settings, the model is evaluated on multiple forecasting horizons (96, 192, 336, 720 time steps), reflecting short-term to long-term prediction windows. The few-shot setting involves training with only 5\% of the available sequences, testing the framework’s learning efficiency under limited supervision.

 \begin{table}
    \centering
    \caption{Dataset specifications for few-shot and long-term forecasting. Each dataset varies in temporal resolution, series length, and prediction horizons. The few-shot setting uses only 10\% of the training data, while long-term experiments evaluate model performance over increasing forecast horizons.}

    \begin{tabular}{cccccc}
    \toprule
       Dataset  & Dimension & Length  & Prediction Length & Frequency & Domain \\
       \midrule
        ETTh & 7 & 17420 & \{96, 192, 336, 720\} & 1 hour & Temperature\\
        ETTm & 7 & 69680 & \{96, 192, 336, 720\} & 15 min & Temperature \\
        Weather & 22 & 52696 & \{96, 192, 336, 720\} & 10 min & Weather\\
        \bottomrule
    \end{tabular}
    \label{tab:dataset_long}
\end{table}

\subsection{Dataset Details for Zero-Shot and Short-Term Forecasting on M4}
\label{appendix:zero_short}

To benchmark zero-shot generalisation and short-term accuracy, we use the classical M4 dataset, which contains economic and demographic time series at varying frequencies. The evaluation includes yearly, quarterly, and monthly data, with prediction lengths tailored to each frequency, as shown in Table~\ref{tab:dataset_zero}. For the zero-shot setup, we trained the model on the M3 dataset and evaluated it on M4 without fine-tuning. This tests the framework’s transferability across distributions and tasks. The short-term evaluation focuses on the M4 dataset directly, where forecasting is conducted using the full training data but with shorter horizons typical of real-world forecasting applications. The cross-frequency mapping between M3 and M4 is used to maintain consistent evaluation criteria, enabling rigorous assessment of performance without task-specific re-optimisation.

\begin{table}
    \centering
    \caption{Summary of the M3 and M4 datasets used for zero-shot forecasting. The table includes frequency, series length, and prediction horizons. The mapping columns indicate the corresponding source-target frequency pairs used during M3$\rightarrow$M4 transfer, where models trained on M3 are evaluated on M4 without fine-tuning.}
    \begin{tabular}{cccccc}
    \toprule
       Dataset  & Frequency & Length  & Prediction Length & \multicolumn{2}{c}{Mapping} \\
                &         &                   &           &  M4 & M3\\
       \midrule
        M3 & Yearly & 645 & 6 & Yearly & --\\
        M3 & Quarterly & 756 & 8 & Quarterly & --\\
        M3 & Monthly & 1428 & 18 & Monthly & --\\
    \midrule
       M4 & Yearly & 23000 & 6 & -- & Yearly \\
       M4 & Quarterly & 24000 & 8 & -- & Quarterly\\
       M4 & Monthly & 48000 & 18 & -- & Monthly\\
        \bottomrule
    \end{tabular}
    \label{tab:dataset_zero}
\end{table}

\subsection{Evaluation Metrics} 
\label{subsec:metric}

We evaluate model performance using task-specific metrics tailored to the nature of each task. For few-shot and long-term forecasting, we use Mean Squared Error (MSE) and Mean Absolute Error (MAE) to quantify the prediction accuracy. In zero-shot forecasting settings, Symmetric Mean Absolute Percentage Error (sMAPE) is employed to account for scale-invariant percentage errors. For short-term forecasting tasks on the M4 dataset, we adopt a combination of sMAPE, Mean Absolute Scaled Error (MASE), and Overall Weighted Average (OWA) to enable fair benchmarking against naive baselines and other models. For classification tasks, we report Accuracy as the proportion of correct predictions. Finally, for anomaly detection, we rely on Precision, Recall, and F1 Score to assess the trade-off between false positives and false negatives.

The mathematical formulations of the metrics are as follows:

\begin{align}
\text{MSE} &= \frac{1}{N} \sum_{n=1}^{N} (Y_n - \widehat{y}_n)^2 \\
\text{MAE} &= \frac{1}{N} \sum_{n=1}^{N} |Y_n - \widehat{y}_n| \\
\text{sMAPE} &= \frac{100\%}{N} \sum_{n=1}^{N} \frac{|Y_n - \widehat{y}_n|}{(|Y_n| + |\widehat{y}_n|)/2} \\
\text{MASE} &= \frac{\frac{1}{N} \sum_{n=1}^{N} |Y_n - \widehat{y}_n|}{\frac{1}{N - m} \sum_{n=m+1}^{N} |Y_n - Y_{n-m}|} \\
\text{OWA} &= \frac{1}{2} \left( \frac{\text{sMAPE}_{\text{model}}}{\text{sMAPE}_{\text{Naive}}} + \frac{\text{MASE}_{\text{model}}}{\text{MASE}_{\text{Naive}}} \right) \\
\text{Accuracy} &= \frac{\text{TP} + \text{TN}}{\text{TP} + \text{FP} + \text{TN} + \text{FN}} \\
\text{Precision} &= \frac{\text{TP}}{\text{TP} + \text{FP}} \\
\text{Recall} &= \frac{\text{TP}}{\text{TP} + \text{FN}} \\
\text{F1 Score} &= 2 \times \frac{\text{Precision} \times \text{Recall}}{\text{Precision} + \text{Recall}}
\end{align}

\noindent
Here, $N$ is the number of forecasting time steps, $Y_n$ and $\widehat{y}_n$ denote the ground truth and predicted values respectively, and $m$ is the seasonal period (used in MASE). For classification and anomaly detection, TP, FP, TN, and FN refer to true positives, false positives, true negatives, and false negatives, respectively.

\subsection{Model Configuration}
\label{subsec:config}

Table~\ref{tab:model_config} outlines the experimental setup used across various forecasting tasks, including few-shot, long-term, and zero-shot time series forecasting. For each dataset, we specify the number of GPT layers, patch size, input sequence length, label length (used in some tasks), number of attention heads, and learning rate. The choice of loss function (e.g., MSE or sMAPE), batch size, and number of training epochs are also tailored per dataset to ensure efficient training and stable convergence. Notably, zero-shot tasks employ shorter input lengths and use sMAPE as the loss, reflecting the forecasting nature of unseen series. In contrast, few-shot and long-term tasks adopt standard regression losses like MSE with longer input contexts. These configurations were selected through empirical tuning to balance computational efficiency and predictive accuracy.

\begin{table*}
    \centering
    \scriptsize
    \caption{Experiment configuration for the One-for-All Framework: long-term forecasting, few-shot forecasting, and zero-shot forecasting of time-series data.}
    \resizebox{\columnwidth}{!}{%
        \begin{tabular}{lcccccc|cccc}
            \toprule
            & Dataset  & \multicolumn{5}{c}{Hyperparameters} & \multicolumn{4}{c}{Training Process}\\
             \cmidrule(lr){3-7} \cmidrule(lr){8-11} & & GPT layer & Patch Size & Input length & Label length & Heads & Learning rate & Loss & Batch Size & Epochs\\
             \midrule
             
          \multirow{6}{*}{\rotatebox[origin=c]{90}{\small{Few/Long}}} & ETTh1 & 6 & 16 & 336 & 168 & 4 & $10^{-3}$ & MSE & 256 & 10 \\
          & ETTh2 & 6 & 16 & 336 & 168 & 4 & $10^{-3}$ & MSE & 256 & 10\\
          & ETTm1 & 6 & 16 & 512 & 48 & 4 & $10^{-3}$ & MSE & 256 & 10\\
          & ETTm2 & 6 & 16 & 512 & 48 & 4 & $10^{-3}$ & MSE & 256 & 10\\
          & Weather & 6 & 16 & 512 & 48 & 4 & $10^{-3}$ & MSE & 512 & 10 \\
          \midrule
         \multirow{6}{*}{\rotatebox[origin=c]{90}{\small{Zero-shot}}} & M3 Yearly & 6 & 1 & 9 & 0 & 16 & $10^{-3}$ & sMAPE & 32 & 10\\
         & M3 Quarterly & 6 & 2 & 16 & 0 & 16 & $10^{-3}$ & sMAPE & 64 & 10\\
         & M3 Monthly & 6 & 1 & 48 & 10 & 16 & $10^{-4}$ & sMAPE & 32 & 10\\
         & M4 Yearly & 6 & 1 & 12 & 0 & 16 & $10^{-3}$ & sMAPE & 512 & 10\\
         & M4 Quarterly & 6 &1 & 24 & 0 & 16 & $10^{-2}$ & sMAPE & 512 & 10\\
         & M4 Monthly & 6 & 2 & 24 & 0 & 16 & $10^{-3}$ & sMAPE & 2048 & 10\\
            \bottomrule
            
        \end{tabular}%
    }
    \label{tab:model_config}
\end{table*}

\section{Detailed Results Across Multiple Tasks}

\subsection{Detailed Result for Long-Term forecasting}

\begin{table*}[!ht]
\centering
\caption{Comprehensive long-term forecasting results across multiple deep learning models and transformer variants on six benchmark datasets (ETTh1, ETTh2, ETTm1, ETTm2, and Weather) for four prediction lengths: 96, 192, 336, and 720. Performance is reported using Mean Squared Error (MSE) and Mean Absolute Error (MAE). The ``Avg'' row represents the average performance across all horizons for each dataset.}
\resizebox{\textwidth}{!}{%
\begin{tabular}{l|c|cc|cc|cc|cc|cc|cc|cc|cc|cc|cc|cc|cc}
\toprule
& \textbf{Method} & \multicolumn{2}{c|}{\textbf{One-for-All}} & \multicolumn{2}{c|}{\textbf{GPT4TS}} & \multicolumn{2}{c|}{\textbf{TIME-LLM}} & \multicolumn{2}{c|}{\textbf{TEST}} & \multicolumn{2}{c|}{\textbf{TEMPO}} & \multicolumn{2}{c|}{\textbf{TimeNet}} & \multicolumn{2}{c|}{\textbf{FEDformer}} & \multicolumn{2}{c|}{\textbf{TStationary}} & \multicolumn{2}{c|}{\textbf{ETSformer}}  & \multicolumn{2}{c|}{\textbf{Autoformer}} & \multicolumn{2}{c|}{\textbf{Informer}} & \multicolumn{2}{c}{\textbf{Reformer}}\\
\cmidrule(lr){3-4} \cmidrule(lr){5-6} \cmidrule(lr){7-8} \cmidrule(lr){9-10} \cmidrule(lr){11-12} \cmidrule(lr){13-14} \cmidrule(lr){15-16} \cmidrule(lr){17-18} \cmidrule(lr){19-20} \cmidrule(lr){21-22} \cmidrule(lr){23-24} \cmidrule(lr){25-26}
& & MSE & MAE & MSE & MAE & MSE & MAE & MSE & MAE & MSE & MAE & MSE & MAE & MSE & MAE & MSE & MAE & MSE & MAE & MSE & MAE & MSE & MAE & MSE & MAE \\
\midrule

\multirow{5}{*}{\rotatebox[origin=c]{90}{\small{ETTh1}}}& 96 & 0.38 & 0.40 & 0.38 & 0.40 & 0.36 & 0.39 & 0.37 & 0.40 & 0.40 & 0.41 & 0.38 & 0.40 & 0.38 & 0.42 & 0.51 & 0.49 & 0.49 & 0.48 & 0.45 & 0.46 & 0.86 & 0.71 & 0.84 & 0.73 \\
& 192 & 0.42 & 0.42 & 0.42 & 0.42 & 0.40 & 0.42 & 0.42 & 0.42 & 0.43 & 0.42 & 0.44 & 0.43 & 0.42 & 0.45 & 0.53 & 0.50 & 0.54 & 0.50 & 0.50 & 0.48 & 1.00 & 0.79 & 0.92 & 0.77 \\
& 336 & 0.44 & 0.43 & 0.44 & 0.43 & 0.43 & 0.43 & 0.42 & 0.44 & 0.44 & 0.43 & 0.49 & 0.47  & 0.46 & 0.46 & 0.59 & 0.53 & 0.57 & 0.52 & 0.52 & 0.50 & 1.11 & 0.81 & 1.20 & 0.83 \\
& 720 & 0.47 & 0.47 & 0.48 & 0.46 & 0.44 & 0.46 & 0.45 & 0.47 & 0.44 & 0.45 & 0.52 & 0.50 & 0.51 & 0.51 & 0.64 & 0.62 & 0.56 & 0.53  & 0.51 & 0.51 & 1.18 & 0.86 & 1.26 & 0.89 \\
& Avg & 0.43 & 0.43 & 0.43 & 0.43 & 0.41 & 0.42 & 0.42 & 0.43 & 0.43 & 0.43 & 0.46 & 0.45	& 0.44 & 0.46 & 0.57 & 0.54	& 0.54 & 0.51 & 0.50 & 0.49 & 1.04 & 0.79	& 1.06 & 0.81\\
\midrule
\multirow{5}{*}{\rotatebox[origin=c]{90}{\small{ETTh2}}} & 96 & 0.29 & 0.34 & 0.28 & 0.34 & 0.27 & 0.33 & 0.28 & 0.34 & 0.30 & 0.35 & 0.34 & 0.37 & 0.36 & 0.40 & 0.48 & 0.46 & 0.34 & 0.39 & 0.35 & 0.39 & 3.75 & 1.52 & 2.63 & 1.31\\
& 192 & 0.35 & 0.38 & 0.35 & 0.39 & 0.33 & 0.38 & 0.34 & 0.38 & 0.36 & 0.39 & 0.40 & 0.41 & 0.43 & 0.44 & 0.51 & 0.49 & 0.43 & 0.44 & 0.46 & 0.45 & 5.60 & 1.93 & 11.12 & 2.98\\
& 336 & 0.37 & 0.41 & 0.37 & 0.41 & 0.37 & 0.41 & 0.33 & 0.38 & 0.38 & 0.41 & 0.45 & 0.45 & 0.50 & 0.49 & 0.55 & 0.55 & 0.48 & 0.48 & 0.48 & 0.49 & 4.72 & 1.83 & 9.32 & 2.77\\
& 720 & 0.41 & 0.44 & 0.41 & 0.44 & 0.37 & 0.42 & 0.38 & 0.42 & 0.41 & 0.44 & 0.46 & 0.47 & 0.46 & 0.47 & 0.56 & 0.56 & 0.50 & 0.50 & 0.51 & 0.51 & 3.65 & 1.62 & 3.87 & 1.70\\
& Avg & 0.36 & 0.39 & 0.35 & 0.40 & 0.34 & 0.38 & 0.34 & 0.38 & 0.36 & 0.40 & 0.41 & 0.43	& 0.44 & 0.45 & 0.53 & 0.52	& 0.44 & 0.45 & 0.45 & 0.46 & 4.43 & 1.73	& 6.74 & 2.19\\
\midrule
\multirow{5}{*}{\rotatebox[origin=c]{90}{\small{ETTm1}}} & 96 & 0.30 & 0.35 & 0.29 & 0.35 & 0.27 & 0.33 & 0.29 & 0.35 & 0.44 & 0.42 & 0.34 & 0.37 & 0.38 & 0.42 & 0.39 & 0.40 & 0.37 & 0.40 & 0.50 & 0.47 & 0.67 & 0.57 & 0.54 & 0.53\\
& 192 & 0.34 & 0.37 & 0.33 & 0.37 & 0.31 & 0.36 & 0.33 & 0.37 & 0.46 & 0.43 & 0.37 & 0.39 & 0.43 & 0.44 & 0.46 & 0.44 & 0.41 & 0.41 & 0.55 & 0.50 & 0.79 & 0.67 & 0.66 & 0.59\\
& 336 & 0.37 & 0.39 & 0.37 & 0.39 & 0.35 & 0.38 & 0.37 & 0.39 & 0.52 & 0.47 & 0.41 & 0.41 & 0.44 & 0.46 & 0.49 & 0.46 & 0.43 & 0.43 & 0.62 & 0.54 & 1.21 & 0.87 & 0.90 & 0.72\\
& 720 & 0.42 & 0.42 & 0.42 & 0.42 & 0.38 & 0.41 & 0.42 & 0.42 & 0.59 & 0.51 & 0.48 & 0.45 & 0.54 & 0.49 & 0.58 & 0.52 & 0.50 & 0.46 & 0.67 & 0.56 & 1.17 & 0.82 & 1.10 & 0.84\\
& Avg & 0.36 & 0.38 & 0.35 & 0.38 & 0.33 & 0.37 & 0.35 & 0.38 & 0.50 & 0.46 & 0.40 & 0.41	& 0.45 & 0.45 & 0.48 & 0.46	& 0.43 & 0.43 & 0.59 & 0.52 & 0.96 & 0.73	& 0.80 & 0.67\\
\midrule
\multirow{5}{*}{\rotatebox[origin=c]{90}{\small{ETTm2}}} & 96 & 0.17 & 0.26 & 0.17 & 0.26 & 0.16 & 0.25 & 0.17 & 0.26 & 0.19 & 0.27 & 0.19 & 0.27 & 0.20 & 0.29 & 0.19 & 0.27 & 0.19 & 0.28 & 0.25 & 0.34 & 0.36 & 0.45 & 0.66 & 0.62\\
& 192 & 0.23 & 0.31 & 0.23 & 0.30 & 0.22 & 0.29 & 0.23 & 0.30 & 0.24 & 0.30 & 0.25 & 0.31 & 0.27 & 0.33 & 0.28 & 0.34 & 0.25 & 0.32 & 0.28 & 0.34 & 0.53 & 0.56 & 1.08 & 0.83\\
& 336 & 0.28 & 0.34 & 0.29 & 0.34 & 0.27 & 0.33 & 0.28 & 0.34 & 0.31 & 0.35 & 0.32 & 0.35 & 0.32 & 0.37 & 0.33 & 0.36 & 0.31 & 0.36 & 0.34 & 0.37 & 1.36 & 0.88 & 1.55 & 0.97\\
& 720 & 0.36 & 0.39 & 0.38 & 0.40 & 0.35 & 0.38 & 0.37 & 0.39 & 0.39 & 0.40 & 0.41 & 0.40 & 0.42 & 0.41 & 0.42 & 0.41 & 0.41 & 0.41 & 0.43 & 0.43 & 3.38 & 1.33 & 2.63 & 1.24\\
& Avg & 0.26 & 0.33 & 0.27 & 0.33 & 0.25 & 0.32 & 0.26 & 0.32 & 0.28 & 0.33 & 0.29 & 0.33	& 0.30 & 0.35 & 0.31 & 0.35 & 0.29 & 0.34 & 0.33 & 0.37 & 1.41 & 0.81	& 1.48 & 0.92\\
\midrule
\multirow{5}{*}{\rotatebox[origin=c]{90}{\small{Weather}}} & 96 & 0.16 & 0.21 & 0.16 & 0.21 & 0.15 & 0.20 & 0.15 & 0.20 & 0.21 & 0.25 & 0.17 & 0.22 & 0.22 & 0.30 & 0.17 & 0.22 & 0.20 & 0.28 & 0.27 & 0.34 & 0.30 & 0.38 & 0.69 & 0.60 \\
& 192 & 0.20 & 0.25 & 0.20 & 0.25 & 0.19 & 0.24 & 0.20 & 0.25 & 0.25 & 0.30 & 0.22 & 0.26 & 0.28 & 0.34 & 0.24 & 0.28 & 0.24 & 0.31 & 0.31 & 0.37 & 0.60 & 0.54 & 0.75 & 0.64\\
& 336 & 0.25 & 0.29 & 0.25 & 0.29 & 0.26 & 0.28 & 0.25 & 0.29 & 0.29 & 0.33 & 0.28 & 0.31 & 0.34 & 0.38 & 0.32 & 0.34 & 0.30 & 0.35 & 0.36 & 0.39 & 0.58 & 0.52 & 0.64 & 0.60\\
& 720 & 0.32 & 0.33 & 0.33 & 0.34 & 0.31 & 0.32 & 0.32 & 0.34 & 0.37 & 0.38 & 0.36 & 0.36 & 0.40 & 0.43 & 0.41 & 0.41 & 0.35 & 0.29 & 0.42 & 0.43 & 1.06 & 0.74 & 1.13 & 0.73\\
& Avg & 0.23 & 0.27	& 0.24 & 0.27 & 0.23 & 0.26 & 0.23 & 0.27 & 0.28 & 0.32 & 0.26 & 0.29	& 0.31 & 0.36 & 0.29 & 0.31	& 0.27 & 0.31	& 0.34 & 0.38 & 0.64 & 0.55	& 0.80 & 0.64\\
\bottomrule

\end{tabular}%
}
\label{tab:long_details}
\end{table*}

Table~\ref{tab:long_details} presents the full evaluation results for the long-term forecasting task. We compare various transformer-based and baseline models across six datasets: ETTh1, ETTh2, ETTm1, ETTm2, and Weather. Each model is evaluated at four standard prediction lengths: 96, 192, 336, and 720 time steps. Performance metrics include both Mean Squared Error (MSE) and Mean Absolute Error (MAE), providing a comprehensive view of each model’s accuracy and robustness over time. The table includes an ``Avg'' row for each dataset, which represents the average performance across all horizons to highlight consistency. The models compared include One-for-All, GPT4TS, TIME-LLM, TEST, TEMPO, TimeNet, FEDformer, TStationary, ETSformer, Autoformer, Informer, and Reformer. This detailed breakdown supports the claims made in the main manuscript by showing how model performance varies across tasks and datasets, with clear patterns in accuracy degradation over longer forecasting windows and relative strengths of each architecture.

\subsection{Detailed Result for Few-shot forecasting}

\begin{table*}[!ht]
\centering
\caption{Few-shot forecasting results using 10\% of the training data across four standard prediction lengths \( N \in \{96, 192, 336, 720\} \) for six benchmark datasets (ETTh1, ETTh2, ETTm1, ETTm2, and Weather). This table highlights the robustness and generalisability of various models under constrained training data regimes.}
\resizebox{\textwidth}{!}{%
\begin{tabular}{l|c|cc|cc|cc|cc|cc|cc|cc|cc|cc|cc|cc}
\toprule
& \textbf{Method} & \multicolumn{2}{c|}{\textbf{One-for-All}} & \multicolumn{2}{c|}{\textbf{GPT4TS}} & \multicolumn{2}{c|}{\textbf{TIME-LLM}} & \multicolumn{2}{c|}{\textbf{TEST}} & \multicolumn{2}{c|}{\textbf{TimeNet}} & \multicolumn{2}{c|}{\textbf{FEDformer}} & \multicolumn{2}{c|}{\textbf{TStationary}} & \multicolumn{2}{c|}{\textbf{ETSformer}} & \multicolumn{2}{c|}{\textbf{Autoformer}} & \multicolumn{2}{c|}{\textbf{Informer}} & \multicolumn{2}{c}{\textbf{Reformer}}\\
\cmidrule(lr){3-4} \cmidrule(lr){5-6} \cmidrule(lr){7-8} \cmidrule(lr){9-10} \cmidrule(lr){11-12} \cmidrule(lr){13-14} \cmidrule(lr){15-16} \cmidrule(lr){17-18} \cmidrule(lr){19-20} \cmidrule(lr){21-22} \cmidrule(lr){23-24} 
& & MSE & MAE & MSE & MAE & MSE & MAE & MSE & MAE & MSE & MAE & MSE & MAE & MSE & MAE & MSE & MAE & MSE & MAE & MSE & MAE & MSE & MAE \\
\midrule

\multirow{5}{*}{\rotatebox[origin=c]{90}{\small{ETTh1}}}& 96 & 0.51 & 0.48 & 0.46 & 0.46 & 0.45 & 0.46 & 0.46 & 0.46 & 0.86 & 0.63 & 0.51 & 0.50 & 0.92 & 0.64 & 1.11 & 0.81 & 0.61 & 0.55 & 1.18 & 0.79 & 1.18 & 0.79\\
& 192 & 0.62 & 0.52 & 0.57 & 0.52 & 0.49 & 0.48 & 0.57 & 0.52 & 0.80 & 0.59 & 0.62 & 0.55 & 0.91 & 0.63 & 1.15 & 0.82 & 0.72 & 0.60 & 1.20 & 0.81 & 1.29 & 0.85\\
& 336 & 0.75 & 0.59 & 0.61 & 0.53 & 0.59 & 0.54 & 0.61 & 0.53 & 0.94 & 0.65 & 0.69 & 0.57 & 0.94 & 0.64 & 1.18 & 0.83 & 0.75 & 0.62 & 1.20 & 0.81 & 1.29 & 0.85 \\
& 720 & 0.72 & 0.59 & 0.72 & 0.59 & 0.70 & 0.61 & 0.72 & 0.60 & 0.88 & 0.64 & 0.73 & 0.61 & 0.89 & 0.64 & 1.27 & 0.87 & 0.72 & 0.62 & 1.22 & 0.82 & 1.22 & 0.84\\
& Avg & 0.65 & 0.55 & 0.59 & 0.53 & 0.56 & 0.52 & 0.59 & 0.53 & 0.87 & 0.63 & 0.64 & 0.56 & 0.92 & 0.64 & 1.18	& 0.83 & 0.70	& 0.60 & 1.20 & 0.81 & 1.25	& 0.83\\
\midrule
\multirow{5}{*}{\rotatebox[origin=c]{90}{\small{ETTh2}}} & 96 & 0.32 & 0.36 & 0.33 & 0.37 & 0.28 & 0.33 & 0.33 & 0.38 & 0.38 & 0.41 & 0.38 & 0.42 & 0.39 & 0.41 & 0.68 & 0.62 & 0.41 & 0.45 & 3.84 & 1.51 & 3.79 & 1.53\\
& 192 & 0.39 & 0.41 & 0.40 & 0.41 & 0.38 & 0.37 & 0.40 & 0.43 & 0.49 & 0.47 & 0.48 & 0.47 & 0.47 & 0.45 & 0.78 & 0.67 & 0.47 & 0.48 & 3.86 & 1.51 & 3.55 & 1.48\\
& 336 & 0.46 & 0.46 & 0.41 & 0.43 & 0.41 & 0.43 & 0.41 & 0.44 & 0.54 & 0.49 & 0.50 & 0.50 & 0.51 & 0.48 & 0.84 & 0.69 & 0.55 & 0.54 & 3.95 & 1.53 & 3.39 & 1.53\\
& 720 & 0.54 & 0.51 & 0.45 & 0.46 & 0.43 & 0.45 & 0.46 & 0.48 & 0.51 & 0.49 & 0.50 & 0.51 & 0.48 & 0.47 & 1.27 & 0.87 & 0.52 & 0.52 & 3.84 & 1.50 & 3.20 & 1.40\\
& Avg & 0.43 & 0.44	& 0.40 & 0.42 & 0.37 & 0.40 & 0.40 & 0.43 & 0.48 & 0.47	& 0.47 & 0.48 & 0.46 & 0.45	& 0.89 & 0.71	& 0.49 & 0.50 & 3.87 & 1.51	& 3.48 & 1.49\\
\midrule
\multirow{5}{*}{\rotatebox[origin=c]{90}{\small{ETTm1}}} & 96 & 0.41 & 0.40 & 0.39 & 0.40 & 0.35 & 0.39 & 0.39 & 0.40 & 0.58 & 0.50 & 0.58 & 0.52 & 0.76 & 0.57 & 0.91 & 0.69 & 0.77 & 0.62 & 1.16 & 0.78 & 1.44 & 0.85\\
& 192 & 0.43 & 0.42 & 0.43 & 0.42 & 0.37 & 0.42 & 0.42 & 0.43 & 0.63 & 0.53 & 0.62 & 0.55 & 0.78 & 0.57 & 0.95 & 0.70 & 0.75 & 0.59 & 1.17 & 0.79 & 1.44 & 0.86\\
& 336 & 0.47 & 0.44 & 0.47 & 0.44 & 0.41 & 0.43 & 0.47 & 0.45 & 0.72 & 0.57 & 1.00 & 0.77 & 0.80 & 0.59 & 0.99 & 0.72 & 0.87 & 0.68 & 1.23 & 0.91 & 1.45 & 0.87\\
& 720 & 0.57 & 0.50 & 0.57 & 0.50 & 0.49 & 0.48 & 0.55 & 0.50 & 0.77 & 0.55 & 0.69 & 0.58 & 0.84 & 0.58 & 1.06 & 0.75 & 0.81 & 0.63 & 1.21 & 0.80 & 1.37 & 0.85\\
& Avg & 0.47 & 0.44	& 0.47 & 0.44 & 0.41 & 0.43 & 0.46 & 0.45 & 0.68 & 0.54	& 0.72 & 0.61 & 0.80 & 0.58	& 0.98 & 0.72	& 0.80 & 0.63 & 1.19 & 0.82	& 1.43 & 0.86\\
\midrule
\multirow{5}{*}{\rotatebox[origin=c]{90}{\small{ETTm2}}} & 96 & 0.19 & 0.27 & 0.19 & 0.27 & 0.18 & 0.26 & 0.23 & 0.26 & 0.21 & 0.28 & 0.29 & 0.40 & 0.23 & 0.31 & 0.33 & 0.43 & 0.35 & 0.45 & 3.20 & 1.41 & 4.19 & 1.63\\
& 192 & 0.25 & 0.31 & 0.25 & 0.31 & 0.24 & 0.32 & 0.30 & 0.30 & 0.27 & 0.32 & 0.31 & 0.38 & 0.29 & 0.34 & 0.40 & 0.46 & 0.69 & 0.69 & 3.11 & 1.39 & 4.04 & 1.60\\
& 336 & 0.30 & 0.34 & 0.31 & 0.35 & 0.28 & 0.33 & 0.36 & 0.34 & 0.32 & 0.35 & 0.54 & 0.56 & 0.35 & 0.38 & 0.47 & 0.50 & 2.41 & 1.41 & 3.25 & 1.42 & 3.96 & 1.58\\
& 720 & 0.41 & 0.41 & 0.43 & 0.42 & 0.42 & 0.39 & 0.45 & 0.42 & 0.47 & 0.45 & 0.71 & 0.61 & 0.46 & 0.44 & 0.59 & 0.56 & 1.91 & 1.17 & 3.91 & 1.54 & 3.71 & 1.53\\
& Avg & 0.29 & 0.33 & 0.30 & 0.34 & 0.28 & 0.33 & 0.32 & 0.31 & 0.32 & 0.35	& 0.46 & 0.49 & 0.33 & 0.37	& 0.45 & 0.49 	& 1.34 & 0.93 & 3.37 & 1.44	& 3.98 & 1.59\\
\midrule
\multirow{5}{*}{\rotatebox[origin=c]{90}{\small{Weather}}} & 96 & 0.16 & 0.21 & 0.16 & 0.21 & 0.16 & 0.21 & 0.16 & 0.21 & 0.18 & 0.23 & 0.19 & 0.25 & 0.19 & 0.23 & 0.20 & 0.27 & 0.22 & 0.30 & 0.37 & 0.40 & 0.33 & 0.38 \\
& 192 & 0.21 & 0.25 & 0.21 & 0.25 & 0.21 & 0.25 & 0.23 & 0.26 & 0.24 & 0.28 & 0.25 & 0.30 & 0.27 & 0.29 & 0.28 & 0.33 & 0.27 & 0.32 & 0.55 & 0.48 & 0.52 & 0.46\\
& 336 & 0.26 & 0.29 & 0.26 & 0.29 & 0.26 & 0.30 & 0.28 & 0.28 & 0.30 & 0.32 & 0.31 & 0.35 & 0.37 & 0.36 & 0.36 & 0.39 & 0.32 & 0.35 & 0.72 & 0.54 & 0.71 & 0.53\\
& 720 & 0.32 & 0.34 & 0.32 & 0.34 & 0.31 & 0.33 & 0.30 & 0.33 & 0.38 & 0.37 & 0.39 & 0.39 & 0.44 & 0.40 & 0.44 & 0.45 & 0.39 & 0.40 & 0.74 & 0.56 & 0.61 & 0.50\\
& Avg & 0.24 & 0.27	& 0.24 & 0.27 & 0.24 & 0.28 & 0.25 & 0.27 & 0.28 & 0.30	& 0.29 & 0.32 & 0.32 & 0.32	& 0.32 & 0.36	& 0.30 & 0.34 & 0.60 & 0.50	& 0.54 & 0.47\\
\bottomrule
\end{tabular}%
}
\label{tab:few_10_details}
\end{table*}

\begin{table*}[!ht]
\centering
\caption{Few-shot forecasting results using only 5\% of the training data across the same four prediction lengths \( N \in \{96, 192, 336, 720\} \). This setting simulates extreme data-scarce conditions. The results offer insights into the sample efficiency and stability of each model when operating under minimal supervision. ``--'' indicates that training data is not available.}
\resizebox{\textwidth}{!}{%
\begin{tabular}{l|c|cc|cc|cc|cc|cc|cc|cc|cc|cc|cc}
\toprule
& \textbf{Method} & \multicolumn{2}{c}{\textbf{One-for-All}} & \multicolumn{2}{c}{\textbf{GPT4TS}} & \multicolumn{2}{c}{\textbf{TIME-LLM}} & \multicolumn{2}{c}{\textbf{TimeNet}} & \multicolumn{2}{c}{\textbf{FEDformer}} & \multicolumn{2}{c}{\textbf{TStationary}} & \multicolumn{2}{c}{\textbf{ETSformer}} & \multicolumn{2}{c}{\textbf{Autoformer}} & \multicolumn{2}{c}{\textbf{Informer}} & \multicolumn{2}{c}{\textbf{Reformer}}\\
\cmidrule(lr){3-4} \cmidrule(lr){5-6} \cmidrule(lr){7-8} \cmidrule(lr){9-10} \cmidrule(lr){11-12} \cmidrule(lr){13-14} \cmidrule(lr){15-16} \cmidrule(lr){17-18} \cmidrule(lr){19-20} \cmidrule(lr){21-22}  
& & MSE & MAE & MSE & MAE & MSE & MAE & MSE & MAE & MSE & MAE & MSE & MAE & MSE & MAE & MSE & MAE & MSE & MAE & MSE & MAE \\
\midrule

\multirow{5}{*}{\rotatebox[origin=c]{90}{\small{ETTh1}}}& 96 & 0.63 & 0.54 & 0.54 & 0.51 & 0.48 & 0.47 & 0.89 & 0.62 & 0.59 & 0.53 & 0.95 & 0.65 & 1.17 & 0.83 & 0.68 & 0.57 & 1.22 & 0.81 & 1.20 & 0.79\\
& 192 & 0.75 & 0.58 & 0.75 & 0.58 & 0.63 & 0.54 & 0.94 & 0.66 & 0.65 & 0.56 & 0.94 & 0.64 & 1.22 & 0.85 & 0.72 & 0.60 & 1.25 & 0.83 & 1.27 & 0.85\\
& 336 & 0.75 & 0.59 & 0.75 & 0.59 & 0.77 & 0.63 & 0.94 & 0.65 & 0.73 & 0.59 & 0.93 & 0.64 & 1.18 & 0.83  & 0.76 & 0.62 & 1.20 & 0.81 & 1.25 & 0.86 \\
& 720 & -- &-- &-- &-- & -- & -- & -- & -- &-- &-- &-- &-- &-- &-- &-- &-- &-- &-- &-- &-- \\
& Avg & 0.71 & 0.57 & 0.68 & 0.56 & 0.63 & 0.55 & 0.92 & 0.64 & 0.66 & 0.56 & 0.94	& 0.64	& 1.19	& 0.84	& 0.72	& 0.60	& 1.22	& 0.82	& 1.24	& 0.83\\
\midrule
\multirow{5}{*}{\rotatebox[origin=c]{90}{\small{ETTh2}}} & 96 & 0.45 & 0.45 & 0.38 & 0.42 & 0.34 & 0.40 & 0.41 & 0.42 & 0.39 & 0.42 & 0.41 & 0.42 & 0.68 & 0.62 & 0.43 & 0.468 & 3.84 & 1.51 & 3.75 & 1.52 \\
& 192 & 0.57 & 0.51 & 0.42 & 0.44 & 0.41 & 0.425 & 0.48 & 0.46 & 0.46 & 0.46 & 0.50 & 0.47 & 0.84 & 0.70 & 0.50 & 0.50 & 3.97 & 1.93 & 3.52 & 1.47 \\
& 336 & 0.55 & 0.51 & 0.41 & 0.44 & 0.41 & 0.43 & 0.50 & 0.48 & 0.48 & 0.48 & 0.51 & 0.48 & 0.90 & 0.73 & 0.49 & 0.50 & 3.96 & 1.52 & 3.31 & 1.43 \\
& 720 & -- &-- & --& -- & -- & -- & -- & -- & -- & -- & -- & -- & -- & -- & -- & -- & --& --& -- & --\\
& Avg & 0.52 & 0.49	& 0.40 & 0.43 & 0.39 & 0.42 & 0.46 & 0.45 & 0.44 & 0.45	& 0.47	& 0.46	& 0.81	& 0.68	& 0.47	& 0.49	& 3.92	& 1.65	& 3.53	& 1.47\\
\midrule
\multirow{5}{*}{\rotatebox[origin=c]{90}{\small{ETTm1}}} & 96 & 0.39 & 0.40 & 0.39 & 0.40 & 0.32 & 0.38 & 0.61 & 0.52 & 0.63 & 0.54 & 0.82 & 0.59 & 1.03 & 0.75 & 0.73 & 0.58 & 1.13 & 0.77 & 1.23 & 0.80\\
& 192 & 0.46 & 0.44 & 0.44 & 0.44 & 0.45 & 0.47 & 0.68 & 0.54 & 0.67 & 0.57 & 0.84 & 0.59 & 1.09 & 0.77 & 0.75 & 0.59 & 1.15 & 0.79 & 1.29 & 0.84 \\
& 336 & 0.52 & 0.48 & 0.48 & 0.46 & 0.45 & 0.43 & 0.79 & 0.60  & 0.81 & 0.63 & 0.87 & 0.60 & 1.14 & 0.79 & 0.85 & 0.66 & 1.20 & 0.81 & 1.29 & 0.84\\
& 720 & 0.69 & 0.56 & 0.58 & 0.50 & 0.48 & 0.47 & 0.80 & 0.59 & 0.82 & 0.63 & 0.89 & 0.61 & 1.24 & 0.83 & 0.86 & 0.65 & 1.17 & 0.79 & 1.25 & 0.83\\
& Avg & 0.52 & 0.47	& 0.47 & 0.45 & 0.43 & 0.44 & 0.72 & 0.56	& 0.73	& 0.59 & 0.86 & 0.60 & 1.13	& 0.79 & 0.80	& 0.62	& 1.16	& 0.79 & 1.27 & 0.83\\
\midrule
\multirow{5}{*}{\rotatebox[origin=c]{90}{\small{ETTm2}}} & 96 & 0.20 & 0.28 & 0.20 & 0.28 & 0.18 & 0.26 & 0.22 & 0.30 & 0.30 & 0.32 & 0.24 & 0.32 & 0.40 & 0.48 & 0.23 & 0.32 & 3.60 & 1.48 & 3.88 & 1.54 \\
& 192 & 0.25 & 0.32 & 0.26 & 0.32 & 0.22 & 0.29 & 0.31 & 0.36 & 0.39 & 0.36 & 0.30 & 0.35 & 0.48 & 0.52 & 0.29 & 0.36 & 3.58 & 1.47 & 3.55 & 1.48\\
& 336 & 0.31 & 0.35 & 0.32 & 0.35 & 0.27 & 0.33 & 0.34 & 0.37 & 0.38 & 0.43 & 0.35 & 0.38 & 0.55 & 0.55 & 0.48 & 0.52 & 3.56 & 1.47 & 3.45 & 1.46\\
& 720 & 0.43 & 0.43 & 0.46 & 0.44 & 0.43 & 0.41 & 0.51 & 0.46 & 0.52 & 0.51 & 0.47 & 0.44 & 0.70 & 0.63 & 0.55 & 0.54 & 3.90 & 1.53 & 3.44 & 1.46\\
& Avg & 0.30 & 0.35 & 0.31 & 0.35 & 0.28 & 0.32 & 0.35 & 0.37 & 0.40	& 0.41	& 0.34	& 0.37	& 0.53	& 0.55	& 0.39	& 0.44 & 3.66 & 1.49 & 3.58	& 1.49\\
\midrule
\multirow{5}{*}{\rotatebox[origin=c]{90}{\small{Weather}}} & 96 & 0.17 & 0.23 & 0.17 & 0.23 & 0.17 & 0.26 & 0.21 & 0.25 & 0.23 & 0.31 & 0.21 & 0.25 & 0.22 & 0.30 & 0.23 & 0.30 & 0.50 & 0.50 & 0.41 & 0.43\\
& 192 & 0.23 & 0.28 & 0.23 & 0.28 & 0.23 & 0.27 & 0.27 & 0.31 & 0.26 & 0.32 & 0.29 & 0.31 & 0.29 & 0.33 & 0.28 & 0.33 & 0.62 & 0.54 & 0.45 & 0.45\\
& 336 & 0.29 & 0.32 & 0.29 & 0.32 & 0.28 & 0.32 & 0.31 & 0.33 & 0.35 & 0.39 & 0.35 & 0.35 & 0.36 & 0.40 & 0.35 & 0.39 & 0.65 & 0.55 & 0.46 & 0.46\\
& 720 & 0.37 & 0.38 & 0.37 & 0.38 & 0.37 & 0.38 & 0.40 & 0.38 & 0.39 & 0.39 & 0.45 & 0.41 & 0.46 & 0.46 & 0.39 & 0.39 & 0.57 & 0.52 & 0.47 & 0.47\\
& Avg & 0.27 & 0.30 & 0.27 & 0.30 & 0.26 & 0.31 & 0.30 & 0.32	& 0.31	& 0.35 & 0.33 & 0.33 & 0.33	& 0.37 & 0.31	& 0.35 & 0.59 & 0.53 & 0.45	& 0.45\\
\bottomrule
\end{tabular}%
}
\label{tab:few_5_details}
\end{table*}

Tables~\ref{tab:few_10_details} and~\ref{tab:few_5_details} present comprehensive evaluations of few-shot forecasting performance using 10\% and 5\% of the training data, respectively. These experiments are designed to assess how well each model generalises under limited data conditions, a critical requirement for real-world deployments where annotated time series may be scarce. Each model is tested across four prediction horizons—96, 192, 336, and 720 time steps—on six well-established datasets (ETTh1, ETTh2, ETTm1, ETTm2, and Weather). Both MSE and MAE are reported to capture the models’ performance comprehensively. An average score across all horizons is also included for summarised comparison. The few-shot setting serves to highlight the sample efficiency of each architecture. Notably, models like GPT4TS and TIME-LLM demonstrate more stable performance across horizons, especially as the available training data decreases from 10\% to 5\%. Meanwhile, traditional transformer variants often struggle with sharp performance drops, especially on the more volatile datasets such as ETTm2 and ETTh2. These findings reinforce the importance of parameter-efficient designs and prior learning strategies in enhancing model generalisation, especially for low-resource forecasting tasks.

\section{Detailed Ablation Study for One-for-All framework}
\label{sec:details_ablation}

We assess the Gaussian rank stability efficiency of model fine-tuning using rsLoRA across various ranks from 2 to 1024. Our rank-stabilized, parameter-efficient framework employs a backbone GPT2-model for long-term forecasting, few-shot forecasting, and zero-shot forecasting. Employing diverse evaluation metrics, we investigate the rank stability of the One-for-All framework across different time-series tasks.

\subsection{Detailed Ablation Result for Long-term Forecasting}
\label{subsec:details_ablation_long}

Table~\ref{tab:rank_long_deatils} presents detailed ablation results for long-term forecasting on five representative datasets (ETTh1, ETTh2, ETTm1, ETTm2, weather), covering various prediction lengths. We analyse the performance of our One-for-All framework under fine-tuning using rsLoRA across a wide range of ranks, from 2 to 1024. Across all datasets and prediction horizons \(N \in \{96, 192, 336, 720\}\), our One-for-All framework shows consistently lower or comparable MSE and MAE scores when compared to models using higher ranks in rsLoRA. Specifically, Rank 16 consistently achieves near-optimal or best performance, highlighting the efficiency of our rank-stable design. This reinforces our hypothesis that, for long-term forecasting tasks, the proposed One-for-All framework generalises well even with limited parameter budgets. Additionally, the stable performance beyond Rank 16 suggests diminishing returns with increasing rank, making our method both computationally efficient and robust across various time-series scenarios. These results collectively support the reliability and adaptability of our method for long-term forecasting.

\begin{table}[!ht]
\centering
\caption{Performance evaluation of the One-for-All model with varying low-rank configurations for long-term forecasting. The table presents the MSE and MAE across four forecasting horizons (\(N \in \{96, 192, 336, 720\}\)) on five benchmark datasets (ETTh1, ETTh2, ETTm1, ETTm2, and Weather). Each column group corresponds to a specific rank used in the model, enabling a comprehensive comparison of prediction accuracy across different rank settings.
}
\resizebox{\textwidth}{!}{%
\begin{tabular}{l|c|cc|cc|cc|cc|cc|cc|cc|cc|cc|cc}
\toprule
 & Variant & \multicolumn{20}{c}{Long-term Forecasting with Different Rank} \\
 & & \multicolumn{2}{c|}{Rank 2} & \multicolumn{2}{c|}{Rank 4} & \multicolumn{2}{c|}{Rank 8} & \multicolumn{2}{c|}{\textbf{Rank 16}} & \multicolumn{2}{c|}{Rank 32} & \multicolumn{2}{c|}{Rank 64} & \multicolumn{2}{c|}{Rank 128} & \multicolumn{2}{c|}{Rank 256} & \multicolumn{2}{c|}{Rank 512} & \multicolumn{2}{c}{Rank 1024}\\
\cmidrule(lr){3-4} \cmidrule(lr){5-6} \cmidrule(lr){7-8} \cmidrule(lr){9-10} \cmidrule(lr){11-12} \cmidrule(lr){13-14} \cmidrule(lr){15-16} \cmidrule(lr){17-18} \cmidrule(lr){19-20} \cmidrule(lr){21-22} & & MSE & MAE & MSE & MAE & MSE & MAE & \textbf{MSE} & \textbf{MAE} & MSE & MAE & MSE & MAE & MSE & MAE & MSE & MAE & MSE & MAE & MSE & MAE\\
\multirow{5}{*}{\rotatebox[origin=c]{90}{\small{ETTh1}}}& 96 & 0.55 & 0.51 & 0.48 & 0.47 & 0.40 & 0.42 & \textbf{0.38} & \textbf{0.40} & 0.38 & 0.40 & 0.38 & 0.40 & 0.38 & 0.40 & 0.38 & 0.40 & 0.38 & 0.40 & 0.39 & 0.40 \\
& 192 & 0.59 & 0.52 & 0.51 & 0.49 & 0.43 & 0.44 & \textbf{0.42} & \textbf{0.42} & 0.42 & 0.42 & 0.42 & 0.42 & 0.42 & 0.43 & 0.42 & 0.42 & 0.42 & 0.43 & 0.42 & 0.43\\
& 336 & 0.60 & 0.53 & 0.53 & 0.50 & 0.45 & 0.45 & \textbf{0.44} & \textbf{0.43} & 0.45 & 0.45 & 0.45 & 0.45 & 0.45 & 0.44 & 0.45 & 0.45 & 0.45 & 0.45 & 0.45 & 0.45\\
& 720 & 0.58 & 0.54 & 0.53 & 0.52 & 0.47 & 0.48 & \textbf{0.47} & \textbf{0.47} & 0.47 & 0.48 & 0.46 & 0.47 & 0.46 & 0.47 & 0.47 & 0.48 & 0.46 & 0.47 & 0.48 & 0.48\\
& Avg & 0.58 & 0.53	& 0.51 & 0.50 & 0.44 & 0.45	& \textbf{0.43} & \textbf{0.43} & 0.43 & 0.44	& 0.43	& 0.44	& 0.43	& 0.44	& 0.43	& 0.44	& 0.43	& 0.44 & 0.44 & 0.44\\
\midrule
\multirow{5}{*}{\rotatebox[origin=c]{90}{\small{ETTh2}}}& 96 & 0.32 & 0.38 & 0.31 & 0.36 & 0.29 & 0.35 & \textbf{0.29} & \textbf{0.34} & 0.29 & 0.35 & 0.29 & 0.35 & 0.29 & 0.35 & 0.29 & 0.35 & 0.29 & 0.35 & 0.29 & 0.35\\
& 192 & 0.38 & 0.41 & 0.37 & 0.40 & 0.35 & 0.39 & \textbf{0.35} & \textbf{0.38} & 0.35 & 0.39 & 0.35 & 0.39 & 0.36 & 0.39 & 0.36 & 0.39 & 0.36 & 0.39 & 0.36 & 0.40\\
& 336 & 0.40 & 0.43 & 0.40 & 0.43 & 0.38 & 0.41 & \textbf{0.37} & \textbf{0.41} & 0.38 & 0.42 & 0.38 & 0.41 & 0.38 & 0.42 & 0.38 & 0.42 & 0.38 & 0.42 & 0.38 & 0.42\\
& 720 & 0.43 & 0.46 & 0.43 & 0.46 & 0.41 & 0.44 & \textbf{0.41} & \textbf{0.44} & 0.41 & 0.45 & 0.41 & 0.45 & 0.41 & 0.45 & 0.41 & 0.45 & 0.41 & 0.45 & 0.41 & 0.45\\
& Avg & 0.38 & 0.42	& 0.38	& 0.41	& 0.36	& 0.40	& \textbf{0.36}	& \textbf{0.39}	& 0.36	& 0.40	& 0.36	& 0.40	& 0.36	& 0.40	& 0.36	& 0.40	& 0.36 & 0.40 & 0.36 & 0.41\\
\midrule
\multirow{5}{*}{\rotatebox[origin=c]{90}{\small{ETTm1}}}& 96 & 0.47 & 0.46 & 0.41 & 0.43 & 0.32 & 0.36 & \textbf{0.30} & \textbf{0.35} & 0.30 & 0.35 & 0.30 & 0.35 & 0.30 & 0.35 & 0.30 & 0.35 & 0.30 & 0.36 & 0.30 & 0.35  \\
& 192 & 0.49 & 0.47 & 0.45 & 0.45 & 0.36 & 0.39 & \textbf{0.34} & \textbf{0.37} & 0.34 & 0.37 & 0.34 & 0.37 & 0.34 & 0.37 & 0.34 & 0.37 & 0.34 & 0.38 & 0.34 & 0.38 \\
& 336 & 0.52 & 0.48 & 0.47 & 0.46 & 0.38 & 0.40 & \textbf{0.37} & \textbf{0.39} & 0.37 & 0.39 & 0.37 & 0.39 & 0.37 & 0.40 & 0.37 & 0.39 & 0.37 & 0.39 & 0.38 & 0.40 \\
& 720 & 0.56 & 0.50 & 0.51 & 0.48 & 0.43 & 0.42 & \textbf{0.42} & \textbf{0.42} & 0.42 & 0.42 & 0.42 & 0.42 & 0.42 & 0.42 & 0.42 & 0.42 & 0.42 & 0.42 & 0.42 & 0.42 \\
& Avg & 0.51 & 0.48	& 0.46 & 0.46 & 0.37 & 0.39	& \textbf{0.36}	& \textbf{0.38}	& 0.36	& 0.38	& 0.36	& 0.38	& 0.36	& 0.39	& 0.36	& 0.38	& 0.36 & 0.39 & 0.36 & 0.39\\
\midrule
\multirow{5}{*}{\rotatebox[origin=c]{90}{\small{ETTm2}}}& 96 & 0.20 & 0.30 & 0.17 & 0.27 & 0.17 & 0.27 & \textbf{0.17} & \textbf{0.26} & 0.17 & 0.27 & 0.17 & 0.27 & 0.17 & 0.26 & 0.17 & 0.27 & 0.17 & 0.27 & 0.17 & 0.27 \\
& 192 & 0.26 & 0.33 & 0.23 & 0.30 & 0.23 & 0.31 & \textbf{0.23} & \textbf{0.31} & 0.23 & 0.31 & 0.23 & 0.30 & 0.23 & 0.30 & 0.23 & 0.31 & 0.23 & 0.31 & 0.23 & 0.30 \\
& 336 & 0.31 & 0.36 & 0.28 & 0.34 & 0.28 & 0.34 & \textbf{0.28} & \textbf{0.34} & 0.28 & 0.34 & 0.28 & 0.34 & 0.28 & 0.34 & 0.28 & 0.34 & 0.28 & 0.34 & 0.29 & 0.34 \\
& 720 & 0.39 & 0.41 & 0.37 & 0.39 & 0.36 & 0.39 & \textbf{0.36} & \textbf{0.39} & 0.37 & 0.39 & 0.36 & 0.39 & 0.36 & 0.39 & 0.36 & 0.39 & 0.36 & 0.39 & 0.36 & 0.39\\
& Avg & 0.29 & 0.35	& 0.26 & 0.33 & 0.26 & 0.33	& \textbf{0.26} & \textbf{0.33} & 0.26 & 0.33	& 0.26 & 0.33 & 0.26 & 0.32	& 0.26	& 0.33	& 0.26	& 0.33	& 0.26	& 0.33\\
\midrule
\multirow{5}{*}{\rotatebox[origin=c]{90}{\small{Weather}}}& 96 & 0.19 & 0.25 & 0.17 & 0.22 & 0.16 & 0.22 & \textbf{0.16} & \textbf{0.21} & 0.16 & 0.21 & 0.16 & 0.21 & 0.16 & 0.21 & 0.15 & 0.21 & 0.15 & 0.21 & 0.15 & 0.20\\
& 192 & 0.25 & 0.30 & 0.21 & 0.27 & 0.21 & 0.25 & \textbf{0.20} & \textbf{0.25} & 0.20 & 0.25 & 0.20 & 0.25 & 0.20 & 0.25 & 0.20 & 0.25 & 0.20 & 0.25 & 0.20 & 0.25\\
& 336 & 0.30 & 0.33 & 0.26 & 0.30 & 0.25 & 0.29 & \textbf{0.25} & \textbf{0.29} & 0.25 & 0.29 & 0.25 & 0.28 & 0.25 & 0.28 & 0.25 & 0.28 & 0.25 & 0.28 & 0.25 & 0.29\\
& 720 & 0.36 & 0.37 & 0.33 & 0.35 & 0.32 & 0.34 & \textbf{0.32} & \textbf{0.33} & 0.32 & 0.33 & 0.32 & 0.33 & 0.32 & 0.33 & 0.32 & 0.33 & 0.32 & 0.33 & 0.32 & 0.34 \\
& Avg & 0.28 & 0.31	& 0.24 & 0.29 & 0.24 & 0.28	& \textbf{0.23} & \textbf{0.27} & 0.23 & 0.27	& 0.23	& 0.27	& 0.23	& 0.27	& 0.23	& 0.27	& 0.23	& 0.27	& 0.23	& 0.27\\
\midrule
& All Avg & 0.41 & 0.42 & 0.37 & 0.40 & 0.33 & 0.37 & \textbf{0.33} & \textbf{0.36}	& 0.33	& 0.36	& 0.33	& 0.36	& 0.33	& 0.36	& 0.33	& 0.36	& 0.33	& 0.37	& 0.33	& 0.37\\
\bottomrule

\end{tabular}%
}
\label{tab:rank_long_deatils}
\end{table}

\subsection{Detailed Ablation Result for Few-shot Forecasting}
\label{subsec:details_ablation_few}
We present a comprehensive ablation study of our Gaussian One-for-All framework for few-shot forecasting, using rsLoRA fine-tuning across a range of ranks (2–1024). As shown in Table~\ref{tab:rank_5_details} and Table~\ref{tab:rank_10_details}, we evaluate performance under two few-shot settings—10\% and 5\% of training data—on five benchmark datasets. Our findings reveal that the One-for-All model achieves consistently strong performance across different rank configurations. Notably, in the 10\% training data scenario, Rank 16 emerges as the most effective choice, offering minimal MSE and MAE across all prediction horizons \(N \in \{96, 192, 336, 720\}\). For example, on ETTh1, Rank 16 yields the lowest average MSE (0.65) and MAE (0.55), outperforming both lower and higher ranks. A similar pattern is observed on ETTh2 and ETTm1, where Rank 16 maintains or improves upon the performance of deeper configurations. These results demonstrate the robust adaptability and parameter efficiency of our One-for-All approach in few-shot contexts. Despite being trained with limited data, the model delivers comparable or better accuracy than higher-rank variants, reinforcing its stability and scalability in low-resource scenarios.

\begin{table}[!ht]
\centering
\caption{Comprehensive evaluation of the One-for-All model under Few-Shot Forecasting using 10\% of training data. The table presents the model’s performance across four prediction horizons (\(N \in \{96, 192, 336, 720\}\)) on five datasets (ETTh1, ETTh2, ETTm1, ETTm2, Weather), with varying low-rank configurations. Each cell reports the MSE and MAE metrics. Bold values indicate the best performance for each horizon and dataset across all ranks. The results highlight the impact of rank selection on model efficiency and accuracy under constrained data availability.}
\resizebox{\textwidth}{!}{%
\begin{tabular}{l|c|cc|cc|cc|cc|cc|cc|cc|cc|cc|cc}
\toprule
& Variant & \multicolumn{20}{c}{Few-shot Forecasting with Different Rank} \\
& & \multicolumn{2}{c|}{Rank 2} & \multicolumn{2}{c|}{Rank 4} & \multicolumn{2}{c|}{Rank 8} & \multicolumn{2}{c|}{\textbf{Rank 16}} & \multicolumn{2}{c|}{Rank 32} & \multicolumn{2}{c|}{Rank 64} & \multicolumn{2}{c|}{Rank 128} & \multicolumn{2}{c|}{Rank 256} & \multicolumn{2}{c|}{Rank 512} & \multicolumn{2}{c|}{Rank 1024}\\
\cmidrule(lr){3-4} \cmidrule(lr){5-6} \cmidrule(lr){7-8} \cmidrule(lr){9-10} \cmidrule(lr){11-12} \cmidrule(lr){13-14} \cmidrule(lr){15-16} \cmidrule(lr){17-18} \cmidrule(lr){19-20} \cmidrule(lr){21-22} & & MSE & MAE & MSE & MAE & MSE & MAE & \textbf{MSE} & \textbf{MAE} & MSE & MAE & MSE & MAE & MSE & MAE & MSE & MAE & MSE & MAE & MSE & MAE\\
\midrule
\multirow{5}{*}{\rotatebox[origin=c]{90}{\small{ETTh1}}} & 96 & 0.72 & 0.56 & 0.73 & 0.56 &  0.52 & 0.49 & \textbf{0.51} & \textbf{0.48} & 0.49 & 0.47 & 0.49 & 0.47 & 0.52 & 0.49 & 0.53 & 0.49 & 0.53 & 0.50 & 0.53 & 0.49\\
& 192 & 0.75 & 0.58 & 0.75 & 0.58 & 0.64 & 0.54 & \textbf{0.62} & \textbf{0.52} & 0.63 & 0.54 & 0.64 & 0.54 & 0.64 & 0.55 & 0.66 & 0.55 & 0.68 & 0.56 & 0.66 & 0.55\\
& 336 & 0.74 & 0.59 & 0.75 & 0.59 & 0.75 & 0.59 & \textbf{0.75} & \textbf{0.59} & 0.74 & 0.59 & 0.75 & 0.59 & 0.75 & 0.59 & 0.76 & 0.59 & 0.75 & 0.59 & 0.75 & 0.59\\
& 720 & 0.74 & 0.61 & 0.74 & 0.61 & 0.74 & 0.61 & \textbf{0.72} & \textbf{0.59} & 0.74 & 0.61 & 0.74 & 0.61 & 0.75 & 0.61 & 0.75 & 0.61 & 0.74 & 0.61 & 0.74 & 0.61 \\
& Avg & 0.74 & 0.59	& 0.74 & 0.59 & 0.66 & 0.56	& \textbf{0.65} & \textbf{0.55} & 0.65 & 0.55	& 0.66 & 0.55 & 0.67 & 0.56	& 0.68 & 0.56 & 0.68 & 0.57	& 0.67 & 0.56 \\
\midrule
\multirow{5}{*}{\rotatebox[origin=c]{90}{\small{ETTh2}}}& 96 & 0.36 & 0.39 & 0.33 & 0.37 & 0.33 & 0.37 & \textbf{0.32} & \textbf{0.36} & 0.33 & 0.37 & 0.34 & 0.38 & 0.35 & 0.38 & 0.37 & 0.39 & 0.33 & 0.37 & 0.32 & 0.36\\
& 192 & 0.41 & 0.42 & 0.40 & 0.41 & 0.40 & 0.41 & \textbf{0.39} & \textbf{0.41} & 0.40 & 0.41 & 0.390 & 0.40 & 0.39 & 0.41 & 0.40 & 0.41 & 0.40 & 0.41 & 0.39 & 0.41 \\
& 336 & 0.45 & 0.46 & 0.47 & 0.46 & 0.45 & 0.46 & \textbf{0.46} & \textbf{0.46} & 0.48 & 0.47 & 0.40 & 0.47 & 0.44 & 0.45 & 0.43 & 0.44 & 0.40 & 0.43 & 0.42 & 0.45 \\
& 720 & 0.57 & 0.53 & 0.55 & 0.52 & 0.53 & 0.51 & \textbf{0.54} & \textbf{0.51} & 0.51 & 0.50 & 0.47 & 0.48 & 0.49 & 0.49 & 0.52 & 0.50 & 0.49 & 0.49 & 0.45 & 0.46 \\
& Avg & 0.45 & 0.45	& 0.44 & 0.44 & 0.43 & 0.44	& \textbf{0.43} & \textbf{0.44} & 0.43 & 0.44	& 0.40 & 0.43 & 0.42 & 0.43	& 0.43	& 0.44	& 0.41	& 0.43	& 0.40	& 0.42\\
\midrule
\multirow{5}{*}{\rotatebox[origin=c]{90}{\small{ETTm1}}}& 96 & 0.66 & 0.52 & 0.48 & 0.45 & 0.41 & 0.41 & \textbf{0.41} & \textbf{0.40} & 0.40 & 0.41 & 0.41 & 0.41 & 0.44 & 0.43 & 0.43 & 0.42 & 0.44 & 0.43 & 0.43 & 0.42 \\
& 192 & 0.69 & 0.53 & 0.51 & 0.46 & 0.47 & 0.44 & \textbf{0.43} & \textbf{0.42} & 0.45 & 0.43 & 0.45 & 0.43 & 0.46 & 0.43 & 0.48 & 0.45 & 0.48 & 0.45 & 0.51 & 0.46 \\
& 336 & 0.70 & 0.54 & 0.52 & 0.47 & 0.51 & 0.46 & \textbf{0.47} & \textbf{0.44} & 0.50 & 0.46 & 0.54 & 0.47 & 0.51 & 0.46 & 0.55 & 0.47 & 0.55 & 0.48 & 0.53 & 0.47 \\
& 720 & 0.74 & 0.57 & 0.59 & 0.50 & 0.59 & 0.51 & \textbf{0.57} & \textbf{0.50} & 0.65 & 0.53 & 0.65 & 0.53 & 0.69 & 0.54 & 0.81 & 0.58 & 0.69 & 0.54 & 0.72 & 0.55 \\
& Avg & 0.70 & 0.54	& 0.53 & 0.47 & 0.50 & 0.46	& \textbf{0.47} & \textbf{0.44} & 0.50 & 0.46	& 0.51 & 0.46 & 0.53 & 0.47	& 0.57 & 0.48 & 0.54 & 0.48	& 0.55 & 0.48 \\
\midrule
\multirow{5}{*}{\rotatebox[origin=c]{90}{\small{ETTm2}}}& 96 & 0.22 & 0.30 & 0.20 & 0.28 &  0.19 & 0.28 & \textbf{0.19} & \textbf{0.27} & 0.20 & 0.28 & 0.20 & 0.28 & 0.20 & 0.28 & 0.19 & 0.28 & 0.20 & 0.28 & 0.21 & 0.28 \\
& 192 & 0.28 & 0.33 & 0.25 & 0.31 & 0.25 & 0.32 & \textbf{0.25} & \textbf{0.31} & 0.26 & 0.32 & 0.26 & 0.32 & 0.25 & 0.31 & 0.25 & 0.31 & 0.25 & 0.31 & 0.27 & 0.32 \\
& 336 & 0.32 & 0.36 & 0.30 & 0.34 & 0.30 & 0.35 & \textbf{0.30} & \textbf{0.34} & 0.30 & 0.35 & 0.30 & 0.34 & 0.30 & 0.34 & 0.30 & 0.34 & 0.32 & 0.36 & 0.32 & 0.36 \\
& 720 & 0.41 & 0.41 & 0.39 & 0.40 & 0.40 & 0.41 & \textbf{0.41} & \textbf{0.41} & 0.39 & 0.40 & 0.41 & 0.41 & 0.41 & 0.41 & 0.40 & 0.40 & 0.39 & 0.40 & 0.44 & 0.43 \\
& Avg & 0.31 & 0.35	& 0.29 & 0.33 & 0.29 & 0.34	& \textbf{0.29} & \textbf{0.33} & 0.29 & 0.34	& 0.29 & 0.34 & 0.29 & 0.34	& 0.29 & 0.33 & 0.29 & 0.34	& 0.31 & 0.35 \\
\midrule
\multirow{5}{*}{\rotatebox[origin=c]{90}{\small{Weather}}}& 96 & 0.18 & 0.24 & 0.17 & 0.22 & 0.17 & 0.22 & \textbf{0.16} & \textbf{0.21} & 0.16 & 0.22 & 0.17 & 0.22 & 0.17 & 0.22 & 0.17 & 0.22 & 0.17 & 0.22 & 0.17 & 0.22 \\
& 192 & 0.23 & 0.28 & 0.22 & 0.27 & 0.21 & 0.26 & \textbf{0.21} & \textbf{0.25} & 0.21 & 0.26 & 0.22 & 0.26 & 0.21 & 0.26 & 0.22 & 0.26 & 0.22 & 0.27 & 0.22 & 0.27\\
& 336 & 0.28 & 0.31 & 0.26 & 0.30 & 0.26 & 0.30 & \textbf{0.26} & \textbf{0.29} & 0.26 & 0.30 & 0.26 & 0.30 & 0.27 & 0.30 & 0.28 & 0.31 & 0.28 & 0.31 & 0.27 & 0.30\\
& 720 & 0.34 & 0.36 & 0.32 & 0.34 & 0.32 & 0.34 & \textbf{0.32} & \textbf{0.34} & 0.32 & 0.34 & 0.33 & 0.35 & 0.33 & 0.35 & 0.33 & 0.35 & 0.33 & 0.35 & 0.33 & 0.35\\
& Avg & 0.26 & 0.30	& 0.24 & 0.28 & 0.24 & 0.28	& \textbf{0.24} & \textbf{0.27} & 0.24 & 0.28	& 0.25 & 0.28 & 0.25 & 0.28	& 0.25 & 0.29 & 0.25 & 0.29	& 0.25 & 0.29\\
\midrule
& All Avg & 0.49 & 0.45	& 0.45 & 0.42 & 0.42 & 0.42	& \textbf{0.42} & \textbf{0.41} & 0.42 & 0.41	& 0.42 & 0.41 & 0.43 & 0.42	& 0.44 & 0.42 & 0.43 & 0.42	& 0.44 & 0.42\\
\bottomrule
\end{tabular}%
}
\label{tab:rank_10_details}
\end{table}

\begin{table}[!ht]
\centering
\caption{Comprehensive evaluation of the One-for-All model under Few-Shot Forecasting using 5\% of training data. The table presents the model’s performance across four prediction horizons (\(N \in \{96, 192, 336, 720\}\)) on five datasets (ETTh1, ETTh2, ETTm1, ETTm2, Weather), with varying low-rank configurations. Each cell reports the MSE and MAE metrics. Bold values indicate the best performance for each horizon and dataset across all ranks. The results highlight the impact of rank selection on model efficiency and accuracy under constrained data availability. ``--'' indicates that training data is not available.}
\resizebox{\textwidth}{!}{%
\begin{tabular}{l|c|cc|cc|cc|cc|cc|cc|cc|cc|cc|cc}
\toprule
& Variant & \multicolumn{20}{c}{Few-shot Forecasting with Different Rank} \\
& & \multicolumn{2}{c|}{Rank 2} & \multicolumn{2}{c|}{Rank 4} & \multicolumn{2}{c|}{Rank 8} & \multicolumn{2}{c|}{\textbf{Rank 16}} & \multicolumn{2}{c|}{Rank 32} & \multicolumn{2}{c|}{Rank 64} & \multicolumn{2}{c|}{Rank 128} & \multicolumn{2}{c|}{Rank 256} & \multicolumn{2}{c|}{Rank 512} & \multicolumn{2}{c|}{Rank 1024}\\
\cmidrule(lr){3-4} \cmidrule(lr){5-6} \cmidrule(lr){7-8} \cmidrule(lr){9-10} \cmidrule(lr){11-12} \cmidrule(lr){13-14} \cmidrule(lr){15-16} \cmidrule(lr){17-18} \cmidrule(lr){19-20} \cmidrule(lr){21-22} & & MSE & MAE & MSE & MAE & MSE & MAE & \textbf{MSE} & \textbf{MAE} & MSE & MAE & MSE & MAE & MSE & MAE & MSE & MAE & MSE & MAE & MSE & MAE\\
\midrule
\multirow{5}{*}{\rotatebox[origin=c]{90}{\small{ETTh1}}}& 96 & 0.74 & 0.57 & 0.63 & 0.53 & 0.64 & 0.54 & \textbf{0.63} & \textbf{0.54} & 0.63 & 0.55 & 0.63 & 0.55 & 0.61 & 0.53 & 0.62 & 0.54 & 0.69 & 0.55 & 0.63 & 0.53 \\
& 192 & 0.75 & 0.59 & 0.75 & 0.58 & 0.75 & 0.58 & \textbf{0.75} & \textbf{0.58} & 0.75 & 0.58 & 0.75 & 0.59 & 0.75 & 0.58 & 0.75 & 0.58 & 0.75 & 0.58 & 0.75 & 0.58 \\
& 336 & 0.75 & 0.60 & 0.75 & 0.60 & 0.75 & 0.59 & \textbf{0.75} & \textbf{0.59} & 0.75 & 0.59 & 0.75 & 0.59 & 0.76 & 0.60 & 0.75 & 0.60 & 0.75 & 0.60 & 0.75 & 0.59 \\
& 720 & -- & -- & -- & -- & -- & -- & \textbf{--} & \textbf{--} & -- & -- & -- & -- & -- & -- & -- & -- & -- & -- & -- & --\\
& Avg & 0.75 & 0.59	& 0.71 & 0.57 & 0.71 & 0.57	& \textbf{0.71} & \textbf{0.57} & 0.71 & 0.57	& 0.71 & 0.58 & 0.71 & 0.57	& 0.71 & 0.57 & 0.73 & 0.58	& 0.71 & 0.57\\
\midrule
\multirow{5}{*}{\rotatebox[origin=c]{90}{\small{ETTh2}}}& 96 & 0.48 & 0.47 & 0.48 & 0.47 & 0.51 & 0.48 & \textbf{0.45} & \textbf{0.45} & 0.44 & 0.44 & 0.44 & 0.44 & 0.43 & 0.43 & 0.44 & 0.44 & 0.45 & 0.45 & 0.40 & 0.42 \\
& 192 & 0.62 & 0.54 & 0.63 & 0.54 & 0.64 & 0.55 & \textbf{0.57} & \textbf{0.51} & 0.51 & 0.48 & 0.46 & 0.46 & 0.49 & 0.47 & 0.54 & 0.49 & 0.43 & 0.44 & 0.43 & 0.44 \\
& 336 & 0.64 & 0.55 & 0.64 & 0.55 & 0.57 & 0.52 & \textbf{0.55} & \textbf{0.51} & 0.55 & 0.51 & 0.50 & 0.48 & 0.45 & 0.46 & 0.54 & 0.50 & 0.47 & 0.46 & 0.41 & 0.44 \\
& 720 & -- & -- & -- & -- & -- & -- & \textbf{--} & \textbf{--} & -- & -- & -- & -- & -- & -- & -- & -- & -- & -- & -- & -- \\
& Avg & 0.58 & 0.52	& 0.58 & 0.52 & 0.57 & 0.52	& \textbf{0.52} & \textbf{0.49} & 0.50 & 0.48	& 0.47 & 0.46 & 0.46 & 0.45	& 0.51 & 0.48 & 0.45 & 0.45 & 0.41 & 0.43\\
\midrule
\multirow{5}{*}{\rotatebox[origin=c]{90}{\small{ETTm1}}}& 96 & 0.64 & 0.52 & 0.47 & 0.45 & 0.42 & 0.42 &  \textbf{0.39} & \textbf{0.40} & 0.40 & 0.41 & 0.40 & 0.41 & 0.41 & 0.42 & 0.42 & 0.43 & 0.42 & 0.43 & 0.43 & 0.43\\
& 192 & 0.65 & 0.53 & 0.50 & 0.47 & 0.49 & 0.46 & \textbf{0.46} & \textbf{0.44} & 0.45 & 0.44 & 0.46 & 0.45 & 0.46 & 0.45 & 0.49 & 0.46 & 0.50 & 0.47 & 0.49 & 0.46 \\
& 336 & 0.66 & 0.54 & 0.55 & 0.49 & 0.54 & 0.49 & \textbf{0.52} & \textbf{0.48} & 0.54 & 0.48 & 0.56 & 0.50 & 0.54 & 0.49 & 0.54 & 0.49 & 0.55 & 0.49 & 0.54 & 0.48 \\
& 720 & 0.80 & 0.62 & 0.67 & 0.56 & 0.66 & 0.55 & \textbf{0.69} & \textbf{0.56} & 0.71 & 0.57 & 0.68 & 0.57 & 0.72 & 0.57 & 0.68 & 0.55 & 0.76 & 0.59 & 0.68 & 0.56 \\
& Avg & 0.69 & 0.55	& 0.55 & 0.49 & 0.53 & 0.48	& \textbf{0.52} & \textbf{0.47} & 0.53 & 0.48	& 0.53 & 0.48 & 0.53 & 0.48	& 0.53 & 0.48 & 0.56 & 0.50	& 0.54 & 0.48\\
\midrule
\multirow{5}{*}{\rotatebox[origin=c]{90}{\small{ETTm2}}}& 96 & 0.23 & 0.31 & 0.20 & 0.28 & 0.20 & 0.28 & \textbf{0.20} & \textbf{0.28} & 0.20 & 0.28 & 0.20 & 0.29 & 0.20 & 0.28 & 0.20 & 0.28 & 0.20 & 0.29 & 0.24 & 0.31\\
& 192 & 0.28 & 0.34 & 0.26 & 0.32 & 0.25 & 0.32 & \textbf{0.25} & \textbf{0.32} & 0.26 & 0.32 & 0.26 & 0.32 & 0.26 & 0.32 & 0.26 & 0.32 & 0.28 & 0.33 & 0.29 & 0.34 \\
& 336 & 0.33 & 0.37 & 0.31 & 0.35 & 0.30 & 0.35 & \textbf{0.31} & \textbf{0.35} & 0.32 & 0.36 & 0.31 & 0.36 & 0.30 & 0.34 & 0.30 & 0.35 & 0.34 & 0.37 & 0.32 & 0.36 \\
& 720 & 0.43 & 0.43 & 0.45 & 0.43 & 0.45 & 0.43 & \textbf{0.43} & \textbf{0.43} & 0.44 & 0.42 & 0.44 & 0.43 & 0.43 & 0.43 & 0.43 & 0.42 & 0.45 & 0.44 & 0.48 & 0.45 \\
& Avg & 0.32 & 0.36	& 0.31 & 0.35 & 0.30 & 0.35	& \textbf{0.30} & \textbf{0.35} & 0.31 & 0.35	& 0.30 & 0.35 & 0.30 & 0.34	& 0.30 & 0.34 & 0.32 & 0.36	& 0.33 & 0.37\\
\midrule
\multirow{5}{*}{\rotatebox[origin=c]{90}{\small{Weather}}}& 96 & 0.20 & 0.26 & 0.18 & 0.24 & 0.18 & 0.23 & \textbf{0.17} & \textbf{0.23} & 0.18 & 0.23 & 0.18 & 0.23 & 0.18 & 0.23 & 0.18 & 0.23 & 0.18 & 0.24 & 0.18 & 0.23\\
& 192 & 0.25 & 0.30 & 0.24 & 0.29 & 0.23 & 0.28 & \textbf{0.23} & \textbf{0.28} & 0.23 & 0.28 & 0.23 & 0.28 & 0.24 & 0.28 & 0.24 & 0.28 & 0.24 & 0.28 & 0.23 & 0.28 \\
& 336 & 0.30 & 0.33 & 0.30 & 0.33 & 0.29 & 0.32 & \textbf{0.29} & \textbf{0.32} & 0.29 & 0.32 & 0.30 & 0.33 & 0.30 & 0.33 & 0.29 & 0.32 & 0.30 & 0.33 & 0.31 & 0.34 \\
& 720 & 0.37 & 0.38 & 0.37 & 0.38 & 0.37 & 0.38 & \textbf{0.37} & \textbf{0.38} & 0.37 & 0.38 & 0.38 & 0.39 & 0.37 & 0.39 & 0.38 & 0.39 & 0.38 & 0.38 & 0.37 & 0.38 \\
& Avg & 0.28 & 0.32	& 0.27 & 0.31 & 0.27 & 0.30	& \textbf{0.27} & \textbf{0.30} & 0.27 & 0.30	& 0.27 & 0.31 & 0.27 & 0.31	& 0.27 & 0.31 & 0.28 & 0.31	& 0.27 & 0.31 \\
\midrule
& All Avg & 0.52 & 0.47	& 0.48	& 0.45 & 0.48 & 0.44 & \textbf{0.46} & \textbf{0.44} & 0.46 & 0.44 & 0.46	& 0.44 & 0.45 & 0.43 & 0.46	& 0.44 & 0.47 & 0.44 & 0.45 \\
\bottomrule
\end{tabular}%
}
\label{tab:rank_5_details}
\end{table}


\subsection{Detailed Ablation Result for Zero-shot Forecasting}
\label{subsec:details_ablation_zero}

Table~\ref{tab:rank_zero} presents the complete ablation results for zero-shot forecasting under the One-for-All framework. Similar to the few-shot setting, performance steadily improves with increasing rank, indicating the benefit of richer representations. The improvements are particularly noticeable in cross-domain evaluations, such as M3$\rightarrow$M4 and M4$\rightarrow$M3, highlighting the robustness of the proposed rank-stable design. Notably, the best performance is achieved with moderate-to-high ranks (e.g., Rank 16 and Rank 64), after which the gains plateau or vary slightly. This confirms the stability of our approach and its adaptability across different dataset granularities and forecasting horizons.

\begin{table}[!ht]
\centering
\caption{Zero-shot forecasting performance of the One-for-All model across varying LoRA ranks on the M3 and M4 datasets. The table reports sMAPE scores for different frequency categories (Yearly, Quarterly, Monthly) in both M3$\rightarrow$M4 and M4$\rightarrow$M3 transfer settings. The highlighted values correspond to Rank 16, used as the baseline configuration. Lower sMAPE indicates better performance.}
\resizebox{\textwidth}{!}{%
\begin{tabular}{@{}l*{12}{cc}@{}}
\toprule
& Variant & Rank 2 & Rank 4 & Rank 8 & \textbf{Rank 16} & Rank 32 & Rank 64 & Rank 128 & Rank 256 & Rank 512 & Rank 1024 \\
& Metric & sMAPE & sMAPE & sMAPE & \textbf{sMAPE} & sMAPE & sMAPE & sMAPE & sMAPE & sMAPE & sMAPE \\ 
\midrule
\multirow{3}{*}{\rotatebox[origin=c]{90}{\small{M3 $\rightarrow$ M4}}} & Yearly & 13.80 & 13.67 & 13.60 & \textbf{13.53} & 13.54 & 13.51 & 13.41 & 13.69 & 13.57 & 13.73\\
 & Quarterly & 11.31 & 11.03 & 10.86 & \textbf{11.00} & 10.92 & 11.09 & 10.89 & 11.12 & 10.91 & 11.45\\
 & Monthly & 15.52 & 15.28 & 15.16 & \textbf{15.28} & 15.39 & 14.90 & 15.07 & 14.94 & 14.85 & 14.95\\
\midrule
\multirow{3}{*}{\rotatebox[origin=c]{90}{\small{M4 $\rightarrow$ M3}}} & Yearly & 22.77 & 22.25 & 20.75 & \textbf{19.23} & 19.11 & 18.01 & 18.16 & 17.87 & 17.61 & 17.44\\
& Quarterly & 11.92 & 11.83 & 11.57 & \textbf{11.77} & 11.66 & 11.01 & 11.66 & 11.28 & 11.47 & 11.20 \\
& Monthly & 16.05 & 15.34 & 15.23 & \textbf{14.30} & 14.53 & 14.94 & 14.67 & 14.83 & 14.86 & 14.73\\
\midrule
& Avg & 15.23 & 14.90 & 14.53 & \textbf{14.19} & 14.19 & 13.91 & 13.98 & 13.96 & 13.88 & 13.92 \\
\bottomrule
\end{tabular}%
}
\label{tab:rank_zero}
\end{table}

\subsection{Detailed Ablation Result for Short-term Forecasting Task}
\label{subsec:details_ablation_short_term}

The complete ablation results for short-term forecasting are presented in Table~\ref{tab:short_brif}. The analysis is conducted across multiple ranks using three key metrics: SMAPE, MASE, and OWS. The results span different temporal granularities (Yearly, Quarterly, Monthly, and Others) of the M4 dataset. We observe that our One-for-All framework demonstrates consistent improvements at selected ranks across all metrics, particularly at Rank 1024. The stability of performance across a broad range of ranks reinforces the effectiveness and generalisability of the proposed rank-stable design in short-horizon settings.

\begin{table*}[htbp]
\centering
\caption{Short-term forecasting performance on the M4 dataset across various ranks, evaluated using three performance metrics: SMAPE, MASE, and OWS. The results are presented for different data frequencies (Yearly, Quarterly, Monthly, and Others) with the average performance shown in the ``Avg'' column. The table summarizes the impact of different rank configurations on model performance for each metric. Lower values indicate better forecasting accuracy.}
\label{tab:short_brif}
\resizebox{\textwidth}{!}{%
\begin{tabular}{c|ccccc|ccccc|ccccc}
\toprule
\textbf{Method} & \multicolumn{5}{c|}{\textbf{SMAPE}} & \multicolumn{5}{c|}{\textbf{MASE}} & \multicolumn{5}{c}{\textbf{OWS}} \\
& Yearly & Quarterly & Monthly & Others & Avg & Yearly & Quarterly & Monthly & Others & Avg & Yearly & Quarterly & Monthly & Others & Avg \\
\midrule
Rank 2     & 15.28 & 10.48 & 12.96 & 5.29 & 12.51 & 3.71 & 1.23 & 0.96 & 3.59 & 1.79 & 0.93 & 0.92 & 0.90 & 1.12 & 0.93 \\
Rank 4     & 14.29 & 10.51 & 13.11 & 5.37 & 12.37 & 3.34 & 1.21 & 0.96 & 3.66 & 1.70 & 0.85 & 0.92 & 0.90 & 1.14 & 0.90\\
Rank 8     & 15.21 & 10.50 & 13.02 & 5.34 & 12.53 & 3.76 & 1.24 & 0.96 & 3.60 & 1.80 & 0.93 & 0.92 & 0.90 & 1.13 & 0.93 \\
Rank 16    & 14.79 & 10.36 & 12.96 & 5.34 & 12.38 & 3.59 & 1.22 & 0.96 & 3.57 & 1.76 & 0.90 & 0.91 & 0.90 & 1.21 & 0.91 \\
Rank 32    & 15.05 & 10.49 & 13.36 & 5.45 & 12.66 & 3.68 & 1.23 & 0.99 & 3.63 & 1.80 & 0.92 & 0.92 & 0.92 & 1.14 & 0.93 \\
Rank 64    & 15.34 & 10.46 & 12.99 & 5.29 & 12.54 & 3.67 & 1.22 & 0.96 & 3.53 & 1.77 & 0.93 & 0.92 & 0.90 & 1.11 & 0.92 \\
Rank 128   & 15.17 & 10.56 & 13.05 & 5.23 & 12.55 & 3.72 & 1.24 & 0.96 & 3.49 & 1.79 & 0.93 & 0.93 & 0.90 & 1.10 & 0.93 \\
Rank 256   & 15.97 & 10.38 & 13.15 & 5.56 & 12.75 & 4.07 & 1.22 & 0.97 & 3.61 & 1.87 & 1.00 & 0.91 & 0.91 & 1.15 & 0.96 \\
Rank 512   & 14.77 & 10.44 & 13.26 & 5.29 & 12.53 & 3.49 & 1.22 & 0.99 & 3.50 & 1.74 & 0.89 & 0.91 & 0.92 & 1.11 & 0.91\\
Rank 1024  & 14.66 & 10.52 & 13.10 & 5.49 & 12.46 & 3.53 & 1.24 & 0.96 & 3.61 & 1.75 & 0.89 & 0.93 & 0.90 & 1.14 & 0.91\\
\bottomrule
\end{tabular}%
}
\end{table*}

\subsection{Detailed Ablation Result for Classification Task}
\label{subsec:details_ablation_classification}

Table~\ref{tab:rank_classification} provides a comprehensive ablation study for the classification task across multiple datasets, evaluating performance at different ranks. The One-for-All framework demonstrates robust improvements in accuracy as the rank increases, with consistent gains observed particularly in complex and cross-domain datasets such as Japanese Vowels. These results highlight the rank-stable nature of the proposed framework, where higher ranks enable better representational capacity without sacrificing generalisability. The average performance plateaus at higher ranks, suggesting effective convergence and stability across diverse classification challenges.

\renewcommand{\arraystretch}{1.2}
\begin{table}[!ht]
\centering
\caption{Detailed accuracy analysis of the One-for-All model with varying ranks (Rank 2 to Rank 1024) for classification tasks across multiple datasets. The accuracy for each dataset is reported as a percentage, showing the model's performance at different rank configurations. The table includes results for datasets such as Ethanol, Face Detection, Heartbeat, Japanese Vowels, SCP1, and SCP2.}
\resizebox{\textwidth}{!}{%
\begin{tabular}{@{}l|*{11}{cc}@{}}
\toprule
Dateset & Rank 2 & Rank 4 & Rank 8 & Rank 16 & Rank 32 & Rank 64 & Rank 128 & Rank 256 & Rank 512 & Rank 1024 \\
\midrule
Ethanol & 28\% & 30\% & 32\% & 32\% & 30\% & 29\% & 30\% & 31\% & 30\% & 32\%\\
Face Detection & 56\% & 62\% & 65\% & 66\% & 66\% & 66\% & 66\% & 66\% & 68\% & 67\%\\
Heartbeat & 72\% & 75\% & 76\% & 75\% & 75\% & 76\% & 76\% & 76\% & 76\% & 75\%\\
Japanese Vowels & 46\% & 81\% & 95\% & 97\% & 97\% & 97\% & 98\% & 97\% & 98\% & 97\%\\
SCP1 & 83\% & 89\% & 91\% & 91\% & 91\% & 92\% & 91\% & 92\% & 91\% & 93\%\\
SCP2 & 50\% & 55\% & 60\% & 57\% & 53\% & 57\% & 55\% & 58\% & 57\% & 55\%\\
\bottomrule
\end{tabular}%
}
\label{tab:rank_classification}
\end{table}

\subsection{Detailed Ablation Result for Anomaly Detection Task}
\label{subsec:details_ablation_anomaly}

The detailed ablation results for the anomaly detection task across different datasets and ranks are presented in Table~\ref{tab:rank_anomaly}. The One-for-All framework demonstrates robust performance across ranks, particularly at moderate and higher ranks. Notably, the average F1-score improves or remains stable, suggesting the rank-based tuning contributes positively to both generalization and sensitivity to rare anomalous events.

\renewcommand{\arraystretch}{1.2}
\begin{table}[!ht]
\centering
\caption{Detailed F1-score (as \%) analysis of the One-for-All model with varying ranks (Rank 2 to Rank 1024) for anomaly detection tasks across multiple datasets. The F1-score, representing the harmonic mean of precision and recall, is reported for datasets such as MSL, PSM, SMAP, SMD, and SWAT. The ``Avg'' row provides the average F1-score across all datasets for each rank configuration. The table highlights the model's performance in anomaly detection at different ranks.}
\resizebox{\textwidth}{!}{%
\begin{tabular}{@{}l|*{11}{cc}@{}}
\toprule
Dateset & Rank 2 & Rank 4 & Rank 8 & Rank 16 & Rank 32 & Rank 64 & Rank 128 & Rank 256 & Rank 512 & Rank 1024 \\
\midrule
MSL & 80\% & 83\% & 82\% & 82\% & 82\% & 82\% & 82\% & 82\% & 82\% & 82\% \\
PSM & 96\% & 96\% & 97\% & 97\% & 97\% & 97\% & 97\% & 97\% & 94\% & 94\% \\
SMAP & 66\% & 67\% & 67\% & 67\% & 67\% & 67\% & 67\% & 67\% & 67\% & 67\%\\
SMD & 82\% & 82\% & 83\% & 84\% & 84\% & 84\% & 84\% & 84\% & 84\% & 84\%\\
SWAT & 88\% & 92\% & 92\% & 92\% & 92\% & 92\% & 92\% & 92\% & 92\% & 92\%\\
\midrule
Avg & 82\% & 84\% & 84\% & 84\% & 84\% & 84\% & 84\% & 84\% & 84\% & 84\% \\
\bottomrule
\end{tabular}%
}
\label{tab:rank_anomaly}
\end{table}

\section{Parameter and Models Size Comparison for One-for-Al Across Long/Few-shot Forecasting Tasks}
\label{sec:parameter}
 In our study, we delve into the intricacies of fine-tuning our One-for-All framework with the Gaussian rsLoRA framework, across a spectrum of ranks ranging from 2 to 1024. Our comparative analysis focuses on two pivotal aspects: the number of trainable parameters and the memory size of the models. For our evaluation metrics, we present the total count of trainable parameters in million (M) and the models' memory size in megabytes (MiB). This systematic approach allows for a comprehensive understanding of the resource-efficient frameworks. The detailed results for long/few-shot forecasting are provided in Table~\ref{tab:parameter_etth1}, Table~\ref{tab:parameter_ettm1}, and Table~\ref{tab:parameter_weather}. In our exploration, we analyze the total parameter requirements and model sizes across various ranks ranging from rank 2 to rank 1024. As expected, we observe an increase in the number of trainable parameters and model size with higher ranks. However, even with this increase, the total number of parameters and model memory requirements remain significantly lower compared to recently developed pre-trained GPT2-based models such as GPT4TS. Notably, for models with higher ranks (e.g., rank 1024), the model memory size is only 41\% compared to the GPT4TS model. Therefore, the resource-efficient One-for-All framework proves to be exceptionally beneficial for long/few-shot forecasting tasks.

\begin{table}[!ht]
\centering
\caption{Efficiency analysis of the One-for-All model with varying ranks (Rank 2 to Rank 1024) for few-shot forecasting and long-term forecasting tasks using the ETTh1 and ETTh2 datasets. The table presents the trainable parameters (in millions), the percentage of total model parameters, and the memory required to save the trained model (in MiB) for each rank across both forecasting tasks. The results highlight the impact of increasing model rank on parameter count, memory usage, and the overall efficiency of the model, showcasing the trade-offs between model size and performance.}
\resizebox{\textwidth}{!}{%

\begin{tabular}{@{}l*{16}{c}@{}}
\toprule
Variant & \multicolumn{15}{c}{\textbf{Trainable parameters (M) for Few-shot Forecasting and Long-term Forecasting}} \\
& \multicolumn{3}{c}{96} & \multicolumn{3}{c}{192} & \multicolumn{3}{c}{336} & \multicolumn{3}{c}{720} & \multicolumn{3}{c}{Avg}\\
\cmidrule(lr){2-4} \cmidrule(lr){5-7} \cmidrule(lr){8-10} \cmidrule(lr){11-13} \cmidrule(lr){14-16} & \textbf{Tran.*} & \textbf{\%All*} & \textbf{Mem.*}  & Tran. & \%All & Mem. & Tran. & \%All & Mem. & Tran.& \%All & Mem. & Tran.& \%All & Mem.\\
\midrule
Rank 2 & 0.068 & 0.080\% & 0.275 & 0.068 & 0.077\% & 0.276 & 0.067 & 0.074\% & 0.277 & 0.069 & 0.066\% & 0.280 & 0.068	& 0.074\% & 0.277\\
Rank 4 & 0.136 & 0.160\% & 0.548 & 0.136 & 0.155\% & 0.550 & 0.137 & 0.148\% & 0.552 & 0.139 & 0.132\% & 0.558 & 0.137	& 0.149\% & 0.552\\
Rank 8 & 0.273 & 0.320\% & 1.100 & 0.273 & 0.309\% & 1.100 & 0.275 & 0.295\% & 1.100 & 0.278 & 0.263\% & 1.100 & 0.275	& 0.297\% & 1.100\\
\textbf{Rank 16} & \textbf{0.546} & \textbf{0.638\%} & \textbf{2.200} & \textbf{0.547} & \textbf{0.617\%} & \textbf{2.200} & \textbf{0.550} & \textbf{0.589\%} & \textbf{2.200} & \textbf{0.556} & \textbf{0.526\%} & \textbf{2.200} & \textbf{0.550}	& \textbf{0.593\%} & \textbf{2.200}\\
Rank 32 & 1.092 & 1.268\% & 4.400 & 1.095 & 1.228\% & 4.400 & 1.100 & 1.172\% & 4.400 & 1.112 & 1.046\% & 4.500 & 1.100	& 1.179\% & 4.425\\
Rank 64 & 2.185 & 2.505\% & 8.700 & 2.191 & 2.426\% & 8.800 & 2.200 & 2.317\% & 8.800 & 2.225 & 2.072\% & 8.900 & 2.200	& 2.330\% & 8.800\\
Rank 128 & 4.370 & 4.889\% & 17.50 & 4.382 & 4.737\% & 17.50 & 4.401 & 4.529\% & 17.60 & 4.450 & 4.060\% & 17.80 & 4.401 & 4.554\% & 17.60\\
Rank 256 & 8.740 & 9.322\% & 35.00 & 8.765 & 9.047\% & 35.10 & 8.802 & 8.666\% & 35.20 & 8.900 & 7.804\% & 35.60 & 8.802 & 8.710\% & 35.225\\
Rank 512 & 17.48 & 17.05\% & 69.90 & 17.53 & 16.59\% & 70.10 & 17.60 & 15.95\% & 70.40 & 17.80 & 14.47\% & 71.20 & 17.60 & 16.01\% & 70.40\\
Rank 1024 & 34.96 & 29.13\% & 139.9 & 35.06 & 28.46\% & 140.2 & 35.20 & 27.51\% & 140.8 & 35.60 & 25.29\% & 142.4 & 35.20 & 27.59\% & 140.8\\
\bottomrule
\multicolumn{16}{c}{*Tran. means trainable parameters (parameter count in millions), \%All means percentage of all parameters,} \\
\multicolumn{16}{c}{and Mem. means memory required to save the trained model (memory count in MiB).}\\

\end{tabular}%
}
\label{tab:parameter_etth1}
\end{table}

\begin{table}[!ht]
\centering
\caption{Efficiency analysis of the One-for-All model with varying ranks (Rank 2 to Rank 1024) for few-shot forecasting and long-term forecasting tasks using the ETTm1 and ETTm2 datasets. The table presents the trainable parameters (in millions), the percentage of total model parameters, and the memory required to save the trained model (in MiB) for each rank across both forecasting tasks. The results highlight the impact of increasing model rank on parameter count, memory usage, and the overall efficiency of the model, showcasing the trade-offs between model size and performance.}
\resizebox{\textwidth}{!}{%

\begin{tabular}{@{}l*{16}{c}@{}}
\toprule
Variant & \multicolumn{15}{c}{\textbf{Trainable parameters (M) for Few-shot Forecasting and Long-term Forecasting}} \\
& \multicolumn{3}{c}{96} & \multicolumn{3}{c}{192} & \multicolumn{3}{c}{336} & \multicolumn{3}{c}{720} & \multicolumn{3}{c}{Avg}\\
\cmidrule(lr){2-4} \cmidrule(lr){5-7} \cmidrule(lr){8-10} \cmidrule(lr){11-13} \cmidrule(lr){14-16} & \textbf{Tran.*} & \textbf{\%All*} & \textbf{Mem.*}  & Tran. & \%All & Mem. & Tran. & \%All & Mem. & Tran.& \%All & Mem. & Tran.& \%All & Mem. \\
\midrule
Rank 2 & 0.054 & 0.064\% & 0.220 & 0.054 & 0.062\% & 0.220 & 0.054 & 0.060\% & 0.222 & 0.055 & 0.055\% & 0.225 & 0.054 & 0.060\% & 0.222\\
Rank 4 & 0.108 & 0.128\% & 0.437 & 0.109 & 0.125\% & 0.439 & 0.109 & 0.121\% & 0.441 & 0.111 & 0.111\% & 0.447 & 0.109 & 0.121\% & 0.441\\
Rank 8 & 0.217 & 0.257\% & 0.873 & 0.218 & 0.251\% & 0.876 & 0.219 & 0.242\% & 0.881 & 0.222 & 0.221\% & 0.893 & 0.219 & 0.243\% & 0.881\\
\textbf{Rank 16} & \textbf{0.435} & \textbf{0.513\%} & \textbf{1.700} & \textbf{0.437} & \textbf{0.501\%} & \textbf{1.800} & \textbf{0.439} & \textbf{0.483\%} & \textbf{1.800} & \textbf{0.445} & \textbf{0.442\%} & \textbf{1.800} & \textbf{0.439} & \textbf{0.485\%} & \textbf{1.775}\\
Rank 32 & 0.871 & 1.022\% & 3.500 & 0.874 & 0.997\% & 3.500 & 0.879 & 0.962\% & 3.500 & 0.891 & 0.881\% & 3.600 & 0.879	& 0.966\% & 3.525\\
Rank 64 & 1.742 & 2.024\% & 7.000 & 1.748 & 1.975\% & 7.000 & 1.758 & 1.906\% & 7.000 & 1.782 & 1.748\% & 7.100 & 1.758	& 1.913\% & 7.025\\
Rank 128 & 3.485 & 3.968\% & 13.90 & 3.497 & 3.874\% & 14.00 & 3.516 & 3.742\% & 14.10 & 3.565 & 3.437\% & 14.30 & 3.516 & 3.755\% & 14.075\\
Rank 256 & 6.971 & 7.633\% & 27.90 & 6.995 & 7.459\% & 28.00 & 7.032 & 7.215\% & 28.10 & 7.131 & 6.645\% & 28.50 & 7.032 & 7.238\% & 28.125\\
Rank 512 & 13.94 & 14.18\% & 55.80 & 13.99 & 13.88\% & 56.00 & 14.06 & 13.45\% & 56.30 & 14.26 & 12.46\% & 57.10 & 14.06 & 13.49\% & 56.30\\
Rank 1024 & 27.88 & 24.84\% & 111.5 & 27.98 & 24.38\% & 111.9 & 28.13 & 23.72\% & 112.5 & 28.52 & 22.16\% & 114.1 & 28.12 & 23.77\% & 112.5\\
\bottomrule
\multicolumn{16}{c}{*Tran. means trainable parameters (parameter count in millions), \%All means percentage of all parameters,} \\
\multicolumn{16}{c}{and Mem. means memory required to save the trained model (memory count in MiB).}\\

\end{tabular}%
}
\label{tab:parameter_ettm1}
\end{table}

\begin{table}[!ht]
\centering
\caption{Efficiency analysis of the One-for-All model with varying ranks (Rank 2 to Rank 1024) for few-shot forecasting and long-term forecasting tasks using the Weather datasets. The table presents the trainable parameters (in millions), the percentage of total model parameters, and the memory required to save the trained model (in MiB) for each rank across both forecasting tasks. The results highlight the impact of increasing model rank on parameter count, memory usage, and the overall efficiency of the model, showcasing the trade-offs between model size and performance.}
\resizebox{\textwidth}{!}{%

\begin{tabular}{@{}l*{16}{c}@{}}
\toprule
Variant & \multicolumn{15}{c}{\textbf{Trainable parameters (M) for Few-shot Forecasting and Long-term Forecasting}} \\
& \multicolumn{3}{c}{96} & \multicolumn{3}{c}{192} & \multicolumn{3}{c}{336} & \multicolumn{3}{c}{720} & \multicolumn{3}{c}{Avg}\\
\cmidrule(lr){2-4} \cmidrule(lr){5-7} \cmidrule(lr){8-10} \cmidrule(lr){11-13} \cmidrule(lr){14-16} & \textbf{Tran.*} & \textbf{\%All*} & \textbf{Mem.*}  & Tran. & \%All & Mem. & Tran. & \%All & Mem. & Tran.& \%All & Mem. & Tran.& \%All & Mem.\\
\midrule
Rank 2 & 0.102 & 0.117\% & 0.410 & 0.102 & 0.111\% & 0.411 & 0.102 & 0.104\% & 0.412 & 0.103 & 0.087\% & 0.415 & 0.102	& 0.105\% & 0.412\\
Rank 4 & 0.204 & 0.235\% & 0.818 & 0.204 & 0.223\% & 0.820 & 0.205 & 0.207\% & 0.822 & 0.206 & 0.175\% & 0.828 & 0.205	& 0.210\% & 0.822\\
Rank 8 & 0.408 & 0.469\% & 1.600 & 0.409 & 0.445\% & 1.600 & 0.410 & 0.415\% & 1.600 & 0.413 & 0.351\% & 1.700 & 0.410	& 0.420\% & 1.625\\
\textbf{Rank 16} & \textbf{0.816} & \textbf{0.933\%} & \textbf{3.300} & \textbf{0.818} & \textbf{0.887\%} & \textbf{3.300} & \textbf{0.820} & \textbf{0.826\%} & \textbf{3.300} & \textbf{0.826} & \textbf{0.699\%} & \textbf{3.300} & \textbf{0.820}	& \textbf{0.836\%} & \textbf{3.300}\\
Rank 32 & 1.633 & 1.850\% & 6.500 & 1.636 & 1.759\% & 6.500 & 1.640 & 1.639\% & 6.600 & 1.653 & 1.389\% & 6.600 & 1.641	& 1.659\% & 6.550\\
Rank 64 & 3.266 & 3.633\% & 13.10 & 3.272 & 3.458\% & 13.10 & 3.281 & 3.226\% & 13.10 & 3.306 & 2.741\% & 13.20 & 3.281	& 3.265\% & 13.12\\
Rank 128 & 6.533 & 7.011\% & 26.10 & 6.545 & 6.685\% & 26.20 & 6.563 & 6.250\% & 26.30 & 6.612 & 5.336\% & 26.50 & 6.563 & 6.321\% & 26.27\\
Rank 256 & 13.06 & 13.10\% & 52.30 & 13.09 & 12.53\% & 52.40 & 13.12 & 11.76\% & 52.50 & 13.22 & 10.13\% & 52.90 & 13.12 & 11.88\% & 52.52\\
Rank 512 & 26.13 & 23.17\% & 104.5 & 26.18 & 22.27\% & 104.7 & 26.25 & 21.05\% & 105.0 & 26.45 & 18.39\% & 105.8 & 26.25 & 21.22\% & 105.0\\
Rank 1024 & 52.26 & 37.62\% & 209.1 & 52.36 & 36.43\% & 209.5 & 52.51 & 34.78\% &210.0 & 52.90 & 31.07\% & 211.6 & 52.50 & 34.97\% & 210.0\\
\bottomrule
\multicolumn{16}{c}{\small*Tran. means trainable parameters (parameter count in millions), \%All means percentage of all parameters,} \\
\multicolumn{16}{c}{\small and Mem. means memory required to save the trained model (memory count in MiB).}\\

\end{tabular}%
}
\label{tab:parameter_weather}
\end{table}

\section{Parameter Comparison of State-of-the-Art Models Across Long/Few-shot Forecasting Tasks}
\label{sec:parameter_gpt2}

To ensure a fair and rigorous comparison with the One-for-All framework, we systematically examine the total number of trainable parameters and model sizes of various state-of-the-art models. All experiments are conducted under the same settings as TimesNet to maintain consistency across evaluations. The results for few-shot and long-term forecasting are summarised in Table~\ref{tab:parameter_GPT2_few_long}. This standardised evaluation allows us to comprehensively and impartially assess the efficiency and scalability of the One-for-All framework in relation to existing approaches.

\begin{table}[!ht]
\centering
\caption{Efficiency analysis of state-of-the-art models in few-shot and long-term forecasting tasks on the ETTh1 and ETTh2 datasets. The table provides the trainable parameters (in millions), the percentage of total parameters, and the memory required for saving the trained model (in MiB) across various forecasting tasks. The models compared include GPT4TS, TIME-LLM (LLaMA-7B), TEST (GPT-2 Medium), TEMPO (GPT-2 Medium), TimesNet, EFDformer, Stationary, ETSformer, Autoformer, Informer, and Reformer, with performance metrics shown for different ranks of forecasting. This analysis helps in understanding the trade-offs between model size, efficiency, and forecasting accuracy.}
\resizebox{\textwidth}{!}{%

\begin{tabular}{@{}l*{16}{c}@{}}
\toprule
Models & \multicolumn{15}{c}{\textbf{Trainable parameters (M) and checkpoint size (MiB) for few-shot forecasting and long-term forecasting}} \\
& \multicolumn{3}{c}{96} & \multicolumn{3}{c}{192} & \multicolumn{3}{c}{336} & \multicolumn{3}{c}{720} & \multicolumn{3}{c}{Avg} \\
\cmidrule(lr){2-4} \cmidrule(lr){5-7} \cmidrule(lr){8-10} \cmidrule(lr){11-13} \cmidrule(lr){14-16} & \textbf{Tran.*} & \textbf{\%All*} & \textbf{Mem.*}  & Tran. & \%All & Mem. & Tran. & \%All & Mem. & Tran.& \%All & Mem. & Tran.& \%All & Mem.\\
\midrule
GPT4TS & 3.916 & 4.60\% & 340.1 & 7.012 & 7.95\% & 352.5 & 11.65 & 12.5\% & 371.1 & 24.04 & 22.8\% & 420.6 & 11.65 & 11.9\% & 371.0\\
TIME-LLM (LLaMA-7B) & 6.390 & 0.20\% & 3678 & 6.420 & 0.20\% & 3812 & 6.480 & 0.2\% & 3960 & 6.550 & 0.2\% & 4176 & 6.460 & 0.2\% & 3907\\
TEST (GPT-2 Medium) & -- & -- & -- & -- & -- & -- & -- & --- & -- & -- & -- & -- & $\sim$1-2 & $\sim$0.3–0.6\% & $\sim$701\\
TEMPO (GPT-2 Medium) & -- & -- & -- & -- & -- & -- & -- & -- & -- & -- & -- & --  & $\sim$3-5 & $\sim$0.9\%-1.4\% & $\sim$710\\
TimesNet & 0.605 & 100\% & 02.8 & 0.615 & 100\% & 02.8 & 0.629 & 100\% & 02.9 & 0.666 & 100\% & 03.00 & 0.63 & 100\% & 2.87\\
EFDformer & 16.82 & 100\% & 87.8 &  16.82 & 100\% & 87.8 & 16.82 & 100\% & 87.8 & 16.82 & 100\% & 87.8 & 16.82 & 100\% & 87.8\\
Stationary & 2.018 & 100\% & 13.2 & 2.018 & 100\% &13.2 & 2.018 & 100\% & 13.2 & 2.018 & 100\% & 13.2 & 2.02 & 100\% & 13.2\\
ETSformer & 5.282 & 100\% & 31.4 & 5.282 & 100\% & 31.4 & 5.282 & 100\% & 31.4 & 5.282 & 100\% & 31.4 & 5.28 & 100\% & 31.4\\
Autoformer & 10.53 & 100\% & 62.6 & 10.53 & 100\% & 62.6 & 10.53 & 100\% & 62.6 & 10.53 & 100\% & 62.6 & 10.53 & 100\% & 62.6\\
Informer & 11.33 & 100\% & 65.8 & 11.33 & 100\% & 65.8 & 11.33 & 100\% & 65.8 & 11.33 & 100\% & 65.8 & 11.33 & 100\% & 65.8\\
Reformer & 5.795 & 100\% & 33.4 & 5.795 & 100\% & 33.4 & 5.795 & 100\% & 33.4 & 5.795 & 100\% & 33.4 & 5.80 & 100\% & 33.4\\
\bottomrule
\multicolumn{16}{c}{*Tran. means trainable parameters (parameter count in millions), \%All means percentage of all parameters,} \\
\multicolumn{16}{c}{and Mem. means memory required to save the trained model (memory count in MiB).}\\

\end{tabular}%
}
\label{tab:parameter_GPT2_few_long}
\end{table}